\documentclass{llncs}
\usepackage{graphicx}
\usepackage{amsmath,amssymb} 
\usepackage{color}

\usepackage{times}
\usepackage{epsfig}
\usepackage{dsfont}
\usepackage[ruled]{algorithm2e}
\usepackage{subfig,wrapfig}
\usepackage{rotating}

\usepackage[width=122mm,left=12mm,paperwidth=146mm,height=193mm,top=12mm,paperheight=217mm]{geometry}
\begin{document}
\mainmatter

\title{A Compact Linear Programming Relaxation for Binary Sub-modular MRF} 

\author{Junyan Wang \and Sai-Kit Yeung}
\institute{Singapore University of Technology and Design, Singapore}

\maketitle

\begin{abstract}
We propose a novel compact linear programming (LP) relaxation for binary sub-modular MRF in the context of object segmentation. Our model is obtained by linearizing an $l_1^+$-norm derived from the quadratic programming (QP) form of the MRF energy. The resultant LP model contains significantly fewer variables and constraints compared to the conventional LP relaxation of the MRF energy. In addition, unlike QP which can produce ambiguous labels, our model can be viewed as a quasi-total-variation minimization problem, and it can therefore preserve the discontinuities in the labels. We further establish a relaxation bound between our LP model and the conventional LP model. In the experiments, we demonstrate our method for the task of interactive object segmentation. Our LP model outperforms QP when converting the continuous labels to binary labels using different threshold values on the entire Oxford interactive segmentation dataset. The computational complexity of our LP is of the same order as that of the QP, and it is significantly lower than the conventional LP relaxation.
\end{abstract}

\section{Introduction}


Markov Random Field (MRF) is a fundamental model for various computer vision tasks. In an MRF model, an MRF energy is to be minimized in order to find an optimal solution to the task. Minimizing general MRF energies is NP-hard \cite{kolmogorov2007minimizing}. Nevertheless, it has been shown that a particular MRF energy can be minimized efficiently and exactly by using max-flow/min-cut algorithms, i.e. the graph cuts \cite{Greig89ExactMRF,Boykov01GraphCut} and such MRF energy is known as the binary sub-modular MRF energy. A typical problem modeled by binary sub-modular MRF is object segmentation \cite{BoykovJolly01GMM-MRF}. Accordingly, we also present our work in the context of object segmentation in this paper.

\begin{figure}
      \begin{center}
\begin{tabular}
{
@{\hspace{0mm}}c@{\hspace{0.1mm}}c@{\hspace{0.1mm}}c @{\hspace{0.1mm}}c
@{\hspace{0.1mm}}c@{\hspace{1mm}}c@{\hspace{1mm}}c @{\hspace{1mm}}c
@{\hspace{1mm}}c
}
\includegraphics[height = 0.8in,width=1.2in]{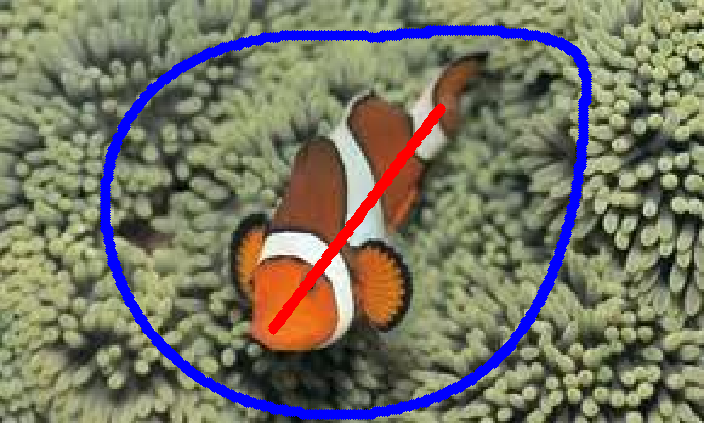}&
\includegraphics[height = 0.8in,width=1.2in]{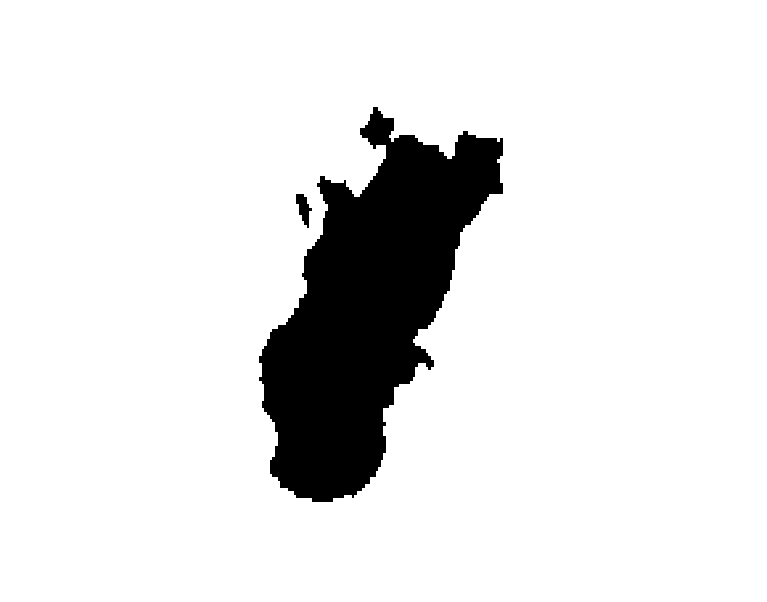}&
 \includegraphics[height = 0.8in,width=1.2in]{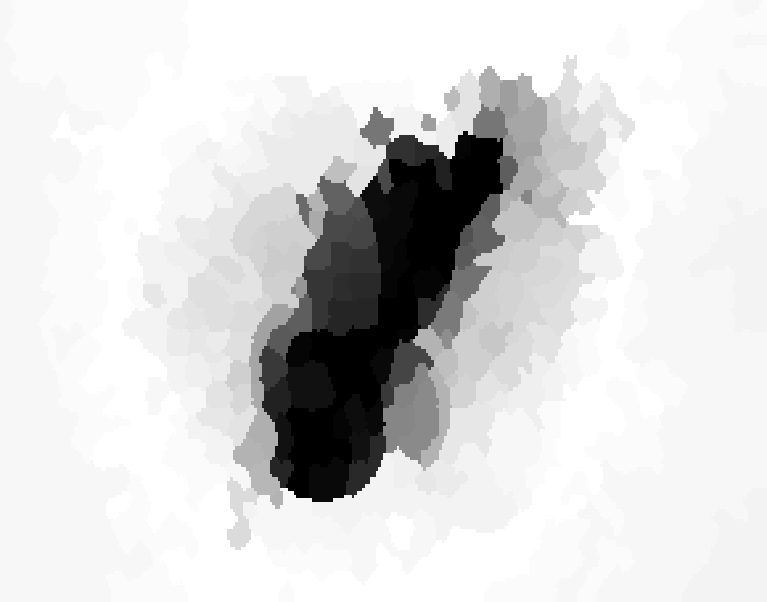}&
 \includegraphics[height = 0.8in,width=1.2in]{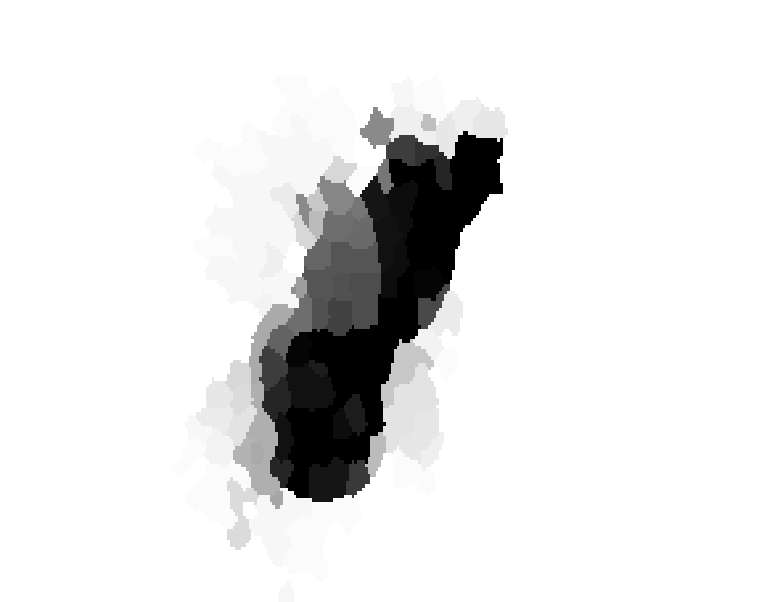}\\
  Input image with seeds & Conventional LP~\cite{LiHongDong10LPSeg} & QP~\cite{grady2006randomwalk,sinop2007seeded} & Our method \\
 \end{tabular}
 \end{center}\vspace{-0.6cm}
  \caption{Conventional Linear programming (LP)~\cite{LiHongDong10LPSeg} is computationally heavy but it preserves discontinuity of the labels at the boundary. Quadratic programming (QP)~\cite{grady2006randomwalk,sinop2007seeded} has much lower computational complexity but often produces over-smooth labels at the boundary. We propose a novel compact LP relaxation that is capable of preserving discontinuities in the labels but with computational complexity in the same order of QP.}\label{FIG:CMP_Label}
\end{figure}

Recently, graph cuts has been criticized for its rigidity in modeling \cite{LiHongDong10LPSeg,bhusnurmath2008graph} the inconveniences caused in its parallelization \cite{bhusnurmath2008graph}. The linear programming (LP) relaxation model has been promoted, since it offers a more flexible and parallelizable substitute to graph cuts for minimizing binary sub-modular MRF energies \cite{LiHongDong10LPSeg,bhusnurmath2008graph}. We will use LP relaxation model or LP model interchangeably if there is no risk of confusion. The conventional LP relaxation model \cite{LiHongDong10LPSeg,bhusnurmath2008graph} was obtained by relaxing the $l_1$-norm pairwise potential in the binary sub-modular MRF. Bhusnurmath and Taylor \cite{bhusnurmath2008graph} proved that the solution by the LP model is identical to the solution by graph cuts. They also argued that the most apparent benefit of LP model over graph cuts is that it allows for convenient parallelization, whereas the existing parallel implementations of graph cuts are based on heuristics, and their effectiveness can only be validated experimentally \cite{liu2010parallel,shekhovtsov2013distributed}. Therefore, the LP model can benefit more from the recent and future development on computing devices \cite{che2008performance}, and it can be more suitable for future applications.

The computational advantage of the LP relaxation model for the binary sub-modular MRF has not been fully revealed in the past. The existing LP relaxation model~\cite{LiHongDong10LPSeg,bhusnurmath2008graph} contains a large amount of auxiliary variables and constraints, which will correspond to very large computational complexity~\cite{ye1991On3L_LP}. It has been reported in~\cite{LiHongDong10LPSeg} that it takes 10 minutes to compute a solution to the LP problem for an image of $300 \times 300$ pixels on a machine with Intel P4(2.8Ghz) CPU and 1G RAM memory, and it has also been shown in~\cite{bhusnurmath2008graph} that the computation time for solving the LP problem on a standard GPU is the same as that required by graph cuts for the same task on a standard CPU. This motivates us to explore the possibilities of reducing the number of auxiliary variables and constraints, i.e. the size of the LP problem, in order to improve the efficiency.

In contrast to the LP model, the computational complexity of the quadratic programming (QP) relaxation for the binary sub-modular MRF energy proposed in \cite{grady2006randomwalk,sinop2007seeded} is much smaller than that required by the conventional LP model. However, the QP model may produce over-smooth ambiguous labels at the boundaries and this may cause incorrect segmentation. Example continuous labels produced by conventional LP and QP are shown in Fig. \ref{FIG:CMP_Label}, in which the solution by conventional LP is clean and more desirable than that by QP. This motivates us to leverage the benefits of both the QP and LP relaxations to achieve efficient and discontinuity-preserving labeling.

In this paper, we propose a novel compact LP relaxation for binary sub-modular MRF. Our LP relaxation is obtained by linearizing a novel $l_1^+$-norm minimization problem that is derived from the QP relaxation model \cite{grady2006randomwalk,sinop2007seeded}. The complexity of the algorithm for solving the proposed LP problem is of the same order of the corresponding QP model, and it is significantly smaller than that of the conventional LP. According to our theoretical analysis, the derived new $l_1^+$-norm minimization is strongly related to the conventional LP which is actually a total-variation minimization problem. Thus, it should be able to preserve discontinuities in labels, while the QP model can over-diffuse the labels. In the experiments, our method also produces segmentation results comparable to those of conventional LP while being more desirable than those of the QP. Our method outperforms QP for all most any threshold values used for binarizing continuous labels on the entire Oxford object segmentation dataset. To summarize, our LP relaxation can be more desirable than the conventional LP and QP relaxations, if the computational cost of the conventional LP is unacceptable and the quality of the solution to QP is not sufficiently satisfactory. At last, we would like to mention that we focus on the binary sub-modular MRF in this work. Hence, our method should not be confused with the recent developments on the approximate solutions to general MRF models \cite{kolmogorov2007minimizing,komodakis2011mrf,kappes2013comparative}.



\section{Background}
\subsection{The MRF model for object segmentation}
The general model of object segmentation considered in this paper is the Potts model that contains two terms: an object-background region model term (data term) $\mathbf{A}(\cdot)$, a.k.a. the unary potential, and a boundary model term $\mathbf{B}(\cdot)$, a.k.a. the pairwise potential. The most general form of this model can be written as follows:
\begin{equation}\label{EQ:SEG_MD}
\min_{\mathcal{L}} \mathbf{A}(\mathcal{L},\theta) + \lambda\mathbf{B}(\mathcal{L}),
\end{equation}
where $\mathcal{L}$ is a label vector corresponding to all pixels/superpixels, an element of $\mathcal{L}$ is $1$ if the pixel/superpixel is from the object, or $0$ otherwise. In this paper, we consider superpixel henceforth. $\theta$ is the set of parameters for the region model, and $\lambda$ is the penalty coefficient.

The object-background region model term is often defined by the probability models of the image values in the respective regions, as shown in the Boykov-Jolly model \cite{BoykovJolly01GMM-MRF}. Given such model, we are able to find the optimal segmentation that corresponds to the highest probability.

The region model alone is generally insufficient for locating the object boundary accurately. Hence, a boundary model term, i.e. the pairwise potential, is often used. The general form of the boundary model term $\mathbf{B}(\cdot)$ can be written as
\begin{equation}\label{EQ:B_Potential}
\mathbf{B}(\mathcal{L}) = \sum_{(i,j)\in \mathcal{E}}B_{ij}|\mathcal{L}_i-\mathcal{L}_{j}|^p,
\end{equation}
where $i$ and $j$ are the indices of pixels/superpixels, $B_{ij} = {1\over1+\{\|I_i-I_{j}\|^2\}}+c$, $I_i,I_{j}$ are the image values at the $i$- and $j$-th pixel/superpixel in the image. $\mathcal{E}$ is a neighborhood system and $p$ is either 1 or 2 in this paper. We will elaborate on the choice of the value of $p$ in this paper. Note that the weight $B_{ij}$ contains two parts. One is ${1\over1+\{\|I_i-I_{j}\|^2\}}$ which encourages a discontinuity at image edges, and the other is a constant $c$ that imposes smoothness to the resultant boundary. The rationale of this form of weights in pairwise potential can be found in the literature of Markov Random Field based segmentation models \cite{StanLi09MRFbook}. The constant weight in the latter part is related to the curve-shortening flow and is more frequently used in the active contour models \cite{kass88snakes,ChanVese01ActiveCon}.


\subsection{LP relaxation of $l_1$-norm minimization for segmentation}\label{SEC:RW}

The region model $\mathbf{A}$ is often formulated linear in the label vector, and it is relatively easy to handle. The complexity of the optimization for the MRF model is often determined by the boundary model term. It has been pointed out that the minimization of the boundary model term is equivalent to an $l_1$-norm minimization problem in \cite{LiHongDong10LPSeg} and \cite{bhusnurmath2008graph} separately and individually. It is further shown in \cite{LiHongDong10LPSeg} and \cite{bhusnurmath2008graph} that the $l_1$-norm minimization problem can be solved by LP, and it is proven in \cite{bhusnurmath2008graph} that the solution to the LP problem in \cite{bhusnurmath2008graph} converges to either 0 or 1 without any external prodding. Careful readers will find that the linear constraints in the LP relaxations in \cite{LiHongDong10LPSeg} and \cite{bhusnurmath2008graph} are slightly different. Without resorting to a rigorous proof, it is easy to see the equivalence between them by setting the RHS of the linear constraints in \cite{bhusnurmath2008graph} to be close to zeros.

In this paper, we shall adopt the LP in \cite{LiHongDong10LPSeg} because its derivation is straightforward. The LP model can be rewritten as follows:
\begin{equation}\label{EQ:LPSEG_POTTS}
\begin{split}
\min_{\mathcal{L}}& \sum_j A_i\mathcal{L}_i+ \sum_{(i,j)\in\mathcal{E}} B_{ij}X_{ij}\\
\hbox{s.t. }&\forall (ij)\in\mathcal{E}, -X_{ij}\leq\mathcal{L}_i-\mathcal{L}_{j}\leq X_{ij}\\
            &\forall i, 0\leq\mathcal{L}_i\leq1, \forall (i,j)\in\mathcal{E}, 0\leq X_{ij}.
\end{split}
\end{equation}
where $\mathbf{A}=\sum_j A_i\mathcal{L}_i$, and the variable $X_{ij}$ is an auxiliary variable induced by the linearization process. A drawback of this well-known LP formulation is that it requires a large number of auxiliary variables and constraints. Suppose there are $N$ elements to be labeled, then there can be \emph{as many as} $N+N\times N$ variables and $N+2N\times N$ linear constraints, which is the worst case. Here, we briefly review the result of complexity analysis for LP. The number of variables, often denoted as $n$, and the number of constraints, denoted as $m$, are the main characteristics of the computational complexity of LP. The computational complexity of LP is known as $O(n^3)$~\cite{ye1991On3L_LP} and when $n$ is fixed the complexity is $O(m)$~\cite{Megiddo1984ConstraintCompCost}. 
As a result, the computational complexity for solving the above LP problem is $O(N^6)$, and the computational cost can be high, which has been witnessed in \cite{LiHongDong10LPSeg}.

Another interesting LP framework of segmentation has been reported in \cite{Schoenemann12LPSegInp}, in which Schoenemann \emph{et al.} proposed an alternative linear formulation of smoothness and curvature priors in segmentation and inpainting.

In this paper, we are interested in the pairwise potential in MRF which is used in almost all the papers on segmentation based on MRF.

\section{MRF and norms minimization}
In this section, we revisit the connections between MRF and $l_1$- and $l_2$-norm minimization.

\subsection{From MRF to $l_1$-norm minimization.}
The boundary potential defined in Eq. (\ref{EQ:B_Potential}) can be reformulated as an $l_1$-norm minimization problem when $p=1$ in Eq. (\ref{EQ:B_Potential}), which is conventional in graph cuts~\cite{Boykov01GraphCut}. This is written as follows:
\begin{equation}\label{EQ:SEG_BW1}
\mathbf{B}_{l_1}(\mathcal{L}) = \sum_{(i,j)\in\mathcal{E}} E_{ij}{B}_{ij}|\mathcal{L}_i-\mathcal{L}_{j}| =\|\textrm{diag}(\vec{w})\mathbf{D}\mathcal{L}\|_{l_1},
\end{equation}
where $E_{ij} = 1$ if $i$ and $j$ are neighbors and $E_{ij} = 0$ otherwise. ${B}_{ij}$ is defined previously. $[\mathrm{diag}(\vec{w})]^{N^2\times N^2}$ is the diagonal matrix composed of $\vec{w}$ and $w_k=B_{ij}$, if $k=i+(j-1)\times N$. $\mathbf{D}$ is an incidence matrix defined as follow:
\begin{equation}
[D]^{N^2\times N}_{ij}=\left\{\begin{array}{ll}
           1  ,&\hbox{ if } j=(i~\textrm{mod}~N) \\
           -1 ,&\hbox{ if } (i~\textrm{mod}~N,j)\in\mathcal{E}
         \end{array}\right.
\end{equation}
We shall call $\mathbf{B}_{l_1}(\mathcal{L})$ the $l_1$-norm boundary term henceforth. This reformulation has been reported previously \cite{sinop2007seeded,LiHongDong10LPSeg}, and it was immediately considered as an LP problem in \cite{LiHongDong10LPSeg}.

\subsection{From MRF to $l_2$-norm minimization.}
In addition, the boundary potential can also be reformulated as an $l_2$-norm minimization problem when $p=2$. This is shown as follows:
\begin{equation}\label{EQ:SEG_BW2}
\begin{split}
\|\textrm{diag}(w)\mathbf{D}\mathcal{L}\|_{l_2}
&=\big\langle\textrm{diag}(w)\mathbf{D}\mathcal{L},\textrm{diag}(w)\mathbf{D}\mathcal{L}\big\rangle^{1\over2}\\
&=\left(\sum_{k} w_k^2(\vec{d}_k\mathcal{L})^2\right)^{1\over2}\\
&=\left(\sum_{(i,j)\in\mathcal{E}} B_{ij}^2(\mathcal{L}_i-\mathcal{L}_j)^2\right)^{1\over2}\\
&\Leftrightarrow\sum_{(i,j)\in\mathcal{E}} B_{ij}^2(\mathcal{L}_i-\mathcal{L}_j)^2
\end{split}
\end{equation}
where $\vec{d}_k$ is the $k^{th}$ row vector of $\mathbf{D}$, and we may define:
\begin{equation}
\mathbf{B}_{l_2}(\mathcal{L}) = \|\textrm{diag}(w)\mathbf{D}\mathcal{L}\|_{l_2},
\end{equation}
and we shall call $\mathbf{B}_{l_2}(\mathcal{L})$ the $l_2$-norm boundary term henceforth.

In the last line of Eq. (\ref{EQ:SEG_BW2}), we omit the square-root operation since it is a scaling of the value without changing the optimality of the solutions and hence we use ``$=$'' instead of ``$\Leftrightarrow$''. This quadratic form is exactly the $l_2$-norm minimization model in \cite{sinop2007seeded}, and it is known as the random walker model. The rationale of this model has been well justified for the model with continuous label values.

\paragraph{The relationship between $l_1$-norm boundary term and $l_2$-norm boundary term.} A canonical relationship between the two terms can be characterized by the following classic inequalities \cite{Trefethen1997numerical}:
\begin{equation}\label{EQ:NormEq1}
{1\over\sqrt{n}}\|\vec{x}\|_{l_1}\leq\|\vec{x}\|_{l_2}\leq\|\vec{x}\|_{l_1},
\end{equation}
The above holds for any vector $\vec{x}\in \mathds{R}^n$. In our case, $\vec{x}=\textrm{diag}(w)\mathbf{D}\mathcal{L}$ and $n=N\times N$. This relation between $l_1$-norm and $l_2$-norm is known as the \emph{equivalence of norms}.

\subsection{Comparing $l_1$-norm minimization with $l_2$-norm minimization}
The $l_2$-norm minimization problem is quadratic in the label variable, and it often results in smooth labels at the object boundary, which may cause ambiguity in the boundary location. In contrast, the original $l_1$-norm minimization will offer clearly distinct labels. See Fig. \ref{FIG:CMP_Label} for one example. A similar problem was identified about 20 years ago in image denoising. It was observed that the minimization of square of image gradients will result in blurry edges. This leads to the invention of the celebrated ROF total-variation minimization model for denoising \cite{ROF92TV}. It has already been pointed out that the $l_1$-norm minimization in our context corresponds to total variation minimization \cite{Chambolle2005BinaryTV}. Likewise, the $l_2$-norm minimization corresponds to the problem of minimization of square of gradients in the context of denoising.

A discontinuity will cause finite total variation, while it yields infinitely large square of gradient. Hence, the discontinuities are can be preserved during the minimization of total variation, while they are allowed by minimization of square of gradient. Hence, the total-variation minimization is more preferable to the minimization of square of gradient, as discontinuities are prevalent in images.

In segmentation, the solution from $l_2$-norm minimization may also become over-smooth and therefore ambiguous at the boundaries. This can affect the accuracy of boundary locating in the segmentation. Accordingly, we also expect the solution of our model to contain sharp discontinuities, which is often allowed by the $l_1$-norm minimization \cite{ROF92TV,Sun2003BPStereo,Brox04OpticalFlow}.

\section{A compact LP model for object segmentation}
\subsection{A compact LP relaxation of $l_1$-norm minimization by factorizing $l_2$-norm}
In the following, we will show that a new $l_1$-norm, which is induced by factorizing the $l_2$-norm boundary term in Eq.(\ref{EQ:SEG_BW2}), can lead to a more compact LP problem with significantly less computational complexity compared to the original LP problem.

First, we rewrite the $l_2$-norm in quadratic form:
\begin{equation}
\begin{split}\label{EQ:SEG_BW}
\mathbf{B}_{l_2}(\mathcal{L}) = \sum_{(i,j)\in\mathcal{E}} B_{ij}^2(\mathcal{L}_i-\mathcal{L}_j)^2 = \mathcal{L}^T\mathbf{\widetilde{W}}\mathcal{L}
\end{split}
\end{equation}
where $\mathbf{\widetilde{W}}=diag(\bar{w})+diag(\hat{w})-2\mathbf{W}$, $\bar{w}_j=\sum_{j'} w_{jj'}$, $\hat{w}_{j'}=\sum_j w_{jj'}$ and $\mathbf{W}=[w_{jj'}]=[E_{jj}{B}_{jj'}]$. The full derivation of the above is included in the supplementary material.

A quadratic optimization problem is NP-hard if the matrix in the quadratic term is non-definite, i.e. the optimization is non-convex. In fact, having even single negative eigenvalue leads to NP-hard problem \cite{Pardalos91QPnp-hard1}. Regarding the convexity of the formulation, we have the following proposition.
\begin{proposition}\label{PP:PSD}
The matrix $\mathbf{\widetilde{W}}$ in Eq.(\ref{EQ:SEG_BW}) is positive semi-definite.
\end{proposition}
The proof is included in the supplementary material. Since $\mathbf{\widetilde{W}}$ is positive semi-definite, the formulation is convex. It is also possible to ensure the matrix to be positive definite by adding a small positive value to the diagonals. 
In addition to the well-posedness of this formulation, we show that positive definiteness of the matrix $\mathbf{\widetilde{W}}$ allows the problem to be linearized.

Our linear relaxation is based on the following facts:
\begin{equation}\label{EQ:W2U}
\mathcal{L}^T\mathbf{\widetilde{W}}\mathcal{L} = \mathcal{L}^T\mathbf{U}^T\mathbf{U}\mathcal{L}=\|\mathbf{U}\mathcal{L}\|_{l_2}^2,
\end{equation}
where $\mathbf{U}$ is an upper triangular matrix of the same dimension of $\mathbf{\widetilde{W}}$ and $\mathbf{\widetilde{W}}=\mathbf{U}^T\mathbf{U}$ is known as the Cholesky factorization/decomposition. The matrix $\mathbf{U}$ uniquely exists for symmetric positive definite matrix $\mathbf{\widetilde{W}}$.

We observe that the matrix $[\mathrm{diag}(\vec{w})\mathbf{D}]$ operating on $\mathcal{L}$ in the $l_1$-norm can also be thought of as being factorized from the matrix $\widetilde{\mathbf{W}}$. To see this, we can rewrite Eq. (\ref{EQ:SEG_BW2}) as follows:
\begin{equation}\label{EQ:W2TV}
\begin{split}
\|\mathrm{diag}(\vec{w})\mathbf{D}\mathcal{L}\|_{l_2}^2&=\mathcal{L}^T[\mathrm{diag}(\vec{w})\mathbf{D}]^T[\mathrm{diag}(\vec{w})\mathbf{D}]\mathcal{L}\\
&=\mathcal{L}^T\mathbf{\widetilde{W}}\mathcal{L}.
\end{split}
\end{equation}

This motivates us to have the following new reformulation of the boundary term $\mathbf{B}$:
\begin{equation}\label{EQ:SEG_LBW}
{
\mathbf{B}_{l_1^+}(\mathcal{L}) = \|\mathbf{U}\mathcal{L}\|_{l_1}
}
\end{equation}
Here, we call the above norm to be minimized as the $l_1^+$-norm boundary term because it is related to the $l_2$-norm boundary term in the same fashion of the conventional $l_1$-norm boundary term, and it can be more desirable. The above resemblance between the $l_1^+$-norm and the $l_1$-norm, also motivates us to define our $l_1^+$-norm boundary term as the \emph{quasi-total-variation}, since $l_1$-norm boundary term is actually the total variation.

A major difference between the conventional $l_1$-norm and our $l_1^+$-norm boundary terms is that the linear operator $\mathbf{U}$ is more compact than $[\mathrm{diag}(\vec{w})D]$, giving rise to a more compact LP relaxation.
\begin{equation}\label{EQ:LPSEG}
\boxed{\begin{split}
\min_{\mathcal{L},\vec{\delta}^+} B(\mathcal{L},&\vec{\delta}^+) = \vec{1}^T \vec{\delta}^+\\
\hbox{s. t. :}  &\hspace{10pt} -\vec{\delta}^+\preceq\mathbf{U}\mathcal{L}\preceq\vec{\delta}^+\\
&\hspace{10pt}\forall i, 0\leq\mathcal{L}_i\leq 1,~\delta^+_i\geq0,
\end{split}}
\end{equation}
where $\vec{\delta}^+$ is an additional vector of auxiliary variables used for the linear relaxation and its dimension is $N$, as the same as $\mathcal{L}$. Essentially, Eq.(\ref{EQ:LPSEG}) reduces the bound of $\mathbf{U}\mathcal{L}$. The above LP is obtained by applying the equivalence between $l_1$-norm minimization and linear programming (Please refer the supplementary materials for full details).

Compared with the conventional LP model in Eq.(\ref{EQ:LPSEG_POTTS}), our model in Eq.(\ref{EQ:LPSEG}) has a significantly less number of variables and number of equality or inequality constraints. This is why we call our formulation a compact linear programming relaxation. Specifically, for the image containing $N$ superpixels, there are $N+N\times N$ variables and $N+2N\times N$ linear constraints for the worst case in the original model~\cite{bhusnurmath2008graph,LiHongDong10LPSeg}, whereas there are only $2N$ variables and $2N$ linear constraints in our model. The complexity of our model is therefore $O(N^3)$ which is the same as QP according to Eq.(\ref{EQ:SEG_BW}). The number of variables and constraints does not change when increasing the number of edges in the graph. We will compare the performance of the two formulations experimentally.

\subsection{Mathematical relationship between total variation and quasi-total variation}
In this subsection, we are particularly interested how tightly the total variation in the form of $l_1$-norm can be related to the quasi-total variation in the form of $l_1^+$-norm, such that the neat properties of the total variation can be shared by the quasi-total variation.

Let us consider the reduced QR factorization of the rectangular matrix $[\mathrm{diag}(\vec{w})\mathbf{D}]$ in the $l_1$-norm boundary term, i.e. $[\mathrm{diag}(\vec{w})\mathbf{D}] = \mathbf{Q}^{N^2\times N}\mathbf{R}^{N\times N}$, where
$\mathbf{Q}$ is an orthogonal matrix, such that $\mathbf{Q}^T\mathbf{Q} = \mathbf{I}^{N\times N}$, and $\mathbf{R}$ is an upper triangular matrix. The following fact will relate our $l_1^+$ relaxation to the original $l_1$-norm minimization.
\begin{theorem}\label{THM:U=R}
The upper triangular matrix $\mathbf{U}$ in the $l_1^+$-norm minimization model in Eq.(\ref{EQ:SEG_LBW}) is identical to the upper triangular matrix $\mathbf{R}$ in the QR factorization of $[\mathrm{diag}(\vec{w})\mathbf{D}]$ in the $l_1$-norm minimization model in Eq.(\ref{EQ:SEG_BW1})
\end{theorem}

The proof of this theorem is presented in the supplementary material. The above theorem implies several additional relationships between the $l_1$-norm and the $l_1^+$-norm.
\begin{corollary}
If we replace the $l_1$-norm with $l_2$-norm, we will have $\|\mathbf{U}\mathcal{L}\|_{l_2}=\|\mathrm{diag}(\vec{w})\mathbf{D}\mathcal{L}\|_{l_2}$.
\end{corollary}
In other words, $\mathbf{U}$ is an equivalent linear operator of $\mathrm{diag}(\vec{w})\mathbf{D}$ for $l_2$-norm.

\begin{corollary}
$\mathbf{U}\mathcal{L} = \mathbf{Q}^T\mathbf{Q}\mathbf{U}\mathcal{L} = \mathbf{Q}^T[\mathrm{diag}(\vec{w})\mathbf{D}\mathcal{L}]$.
\end{corollary}
The above equality implies that the $l_1^+$-norm is the $l_1$-norm of the linearly transformed weighted gradients, and the transformation matrix is $\mathbf{Q}$. The weighted variations in $\mathcal{L}$ are projected on the subspace of $\mathbf{Q}$ before calculating the total. Hence, we may also view our $l_1^+$-norm as a total subspace-variation. This observation implies that the quasi-total variation minimization may share the discontinuity preservability of the total variation minimization.

Besides, Theorem \ref{THM:U=R} offers us a stronger relationship between the two formulations in terms of a tight equivalence-of-norm bound.
\begin{theorem}\label{TH:NormEq}
The difference of $l_1^+$-norm and $l_1$-norm satisfies the following inequalities:
\begin{equation}
{1\over\|\mathbf{Q}\|_{l_1}}\|\mathrm{diag}(\vec{w})\mathbf{D}\mathcal{L}\|_{l_1}\leq\|\mathbf{U}\mathcal{L}\|_{l_1}\leq\|\mathbf{Q}^T\|_{l_1}\|\mathrm{diag}(\vec{w})\mathbf{D}\mathcal{L}\|_{l_1}
\end{equation}
\end{theorem}

The proof of this theorem is included in the supplementary material.

Comparing this above norm equivalence with Eq.(\ref{EQ:NormEq1}), we can conclude that the worst-case difference between the $l_1^+$-norm and the $l_1$-norm can be much smaller than that between $l_2$-norm and $l_1$-norm. This is because we are allowed to reduce $\|\mathbf{Q}\|_{l_1}$ and $\|\mathbf{Q}^T\|_{l_1}$ in order to improve the approximation, while the approximation in Eq.(\ref{EQ:NormEq1}) is hardy. In addition, according to Theorem \ref{TH:NormEq}, we consider our compact LP model in (\ref{EQ:LPSEG}) as a tight relaxation of the original $l_1$-norm boundary term.

\section{Experiments}\label{SEC:ExpRst}
In the experiment, we will evaluate our model for the boundary term. This evaluation is possible since the seed-initialized interactive object segmentation is generally formulated with only the boundary term, and the seed points (seeds) are incorporated as the hard constraints encoded by the unary potential \cite{BoykovJolly01GMM-MRF}. We compare our LP for $l_1^+$-norm minimization with the original graph cuts (GC)~\cite{Boykov01GraphCut}, the $l_1$-norm minimization via LP~\cite{LiHongDong10LPSeg}, and the $l_2$-norm minimization by QP~\cite{grady2006randomwalk,sinop2007seeded}.

\subsection{Experimental settings}
\paragraph{Data and performance measure}
We mainly experiment on two segmentation datasets. The first one will be the clownfish segmentation dataset constructed by the authors which contains 62 images of clownfish. The other is the Oxford interactive segmentation benchmark dataset \cite{Gulshan10StarGeoGraphCut}. Ground truth results and user input seeds on objects and backgrounds are provided in both datasets. The clownfish dataset may be simpler than the Oxford dataset because it contains less variations of objects. It can be considered as a controlled dataset, and the Oxford dataset is more like a natural dataset. We shall use the clownfish dataset for proof of concept, and then validate our model under more general situations in the Oxford dataset. The performance of the methods is measured by the overlapping ratio between the labeled region and the ground truth object region:
$$\Gamma = {\hbox{size}\big(\hbox{Result Region}\cap\hbox{True Region}\big)\over\hbox{size}\big(\hbox{Result Region}\cup\hbox{True Region}\big)}.$$

To evaluate the performance gain in terms of computation. We perform the conventional LP and our proposed LP on GPU for synthetic data. In this experiment, we randomly generate the model parameters and apply the interior point method to solving the LP. We are unable to experiment on images due to the limit on our hardware.

\paragraph{Implementation issues}
We adopt superpixelization \cite{Levinshtein2009TurboPixels} as a preprocessing to reduce the computational cost. The number of superpixels is around 800 for all test images. We choose the average color of each superpixel to represent the superpixel. We implement all the methods in MATLAB. We used the \verb"linprog" function and \verb"quadprog" function. We use default option settings of the functions. The graph cuts is based on Michael Rubinstein's implementation \footnote{\url{http://www.mathworks.com/matlabcentral/fileexchange/21310-maxflow}}. There are some parameters in the model for segmentation. We used $c=0.00001, \lambda=10$ in all the experiments. The threshold value for converting the continuous labels to binary labels is empirically chosen as 0.08. We also experiment on the effect of differnt threshold values. We perform the experiments on a PC with Intel Core i5-450M (2.4GHz) processor and 4GB memory.

\subsection{Results}

\begin{figure}[!h]
    \begin{center}
\begin{tabular}
{
@{\hspace{0mm}}c@{\hspace{0.5mm}}c@{\hspace{0.5mm}}c @{\hspace{0.5mm}}c
@{\hspace{0.5mm}}c@{\hspace{1mm}}c@{\hspace{1mm}}c @{\hspace{1mm}}c
@{\hspace{1mm}}c
}
\begin{sideways}\parbox{15mm}{\centering\footnotesize Input images with seeds}\end{sideways} &
  \includegraphics[height=1.6cm]{imgs/L2LPSelected/012_init.png}&
  \includegraphics[height=1.6cm]{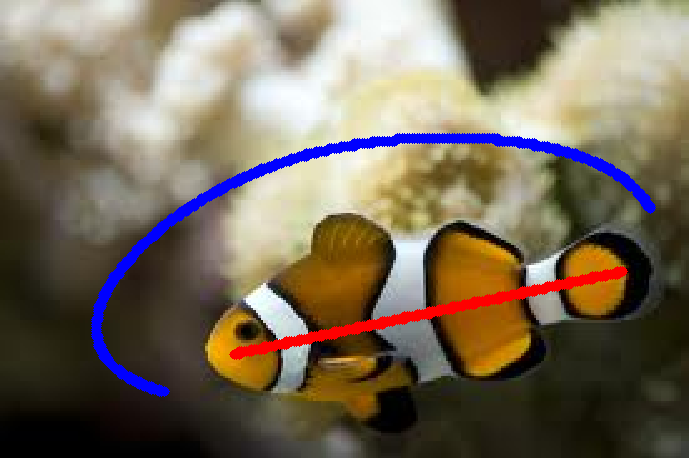}&
  \includegraphics[height=1.6cm]{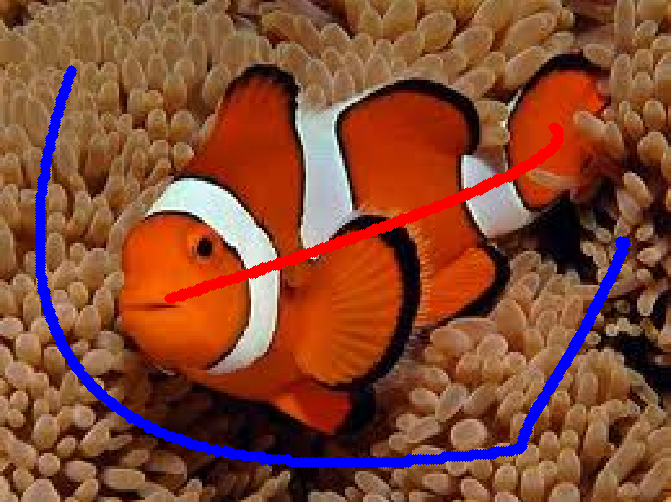}&
  \includegraphics[height=1.6cm]{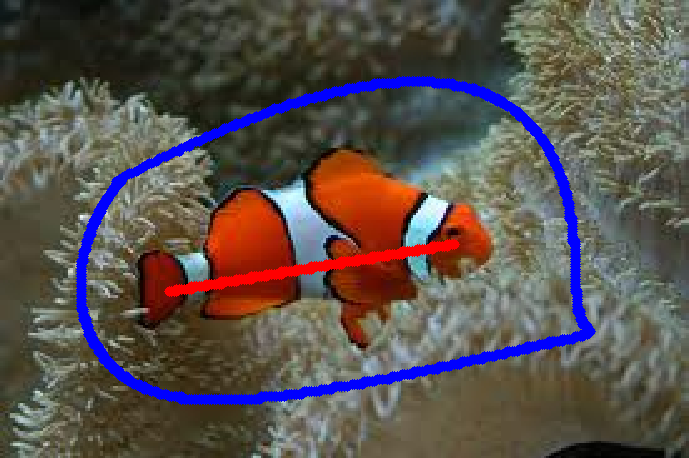}\\
  \begin{sideways}\parbox{15mm}{\centering\footnotesize GC~\cite{Boykov01GraphCut}}\end{sideways} &
  \includegraphics[height=1.6cm]{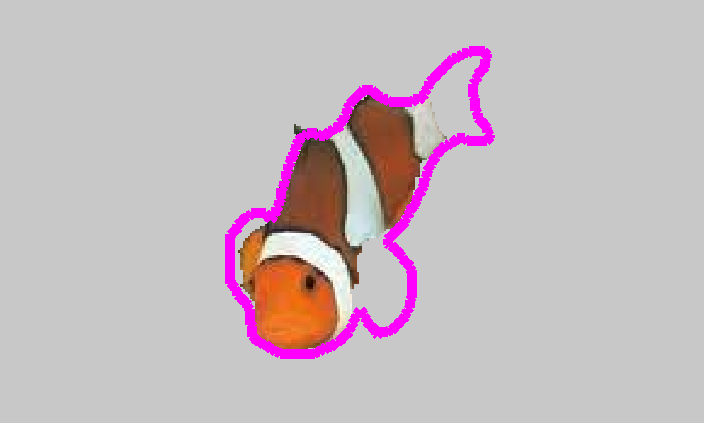}&
  \includegraphics[height=1.6cm]{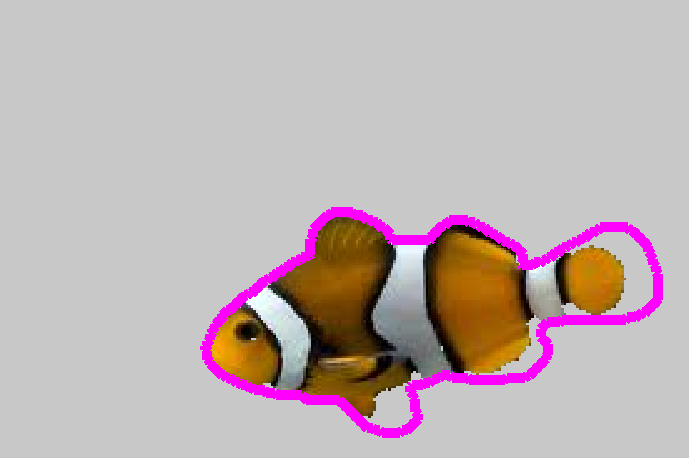}&
  \includegraphics[height=1.6cm]{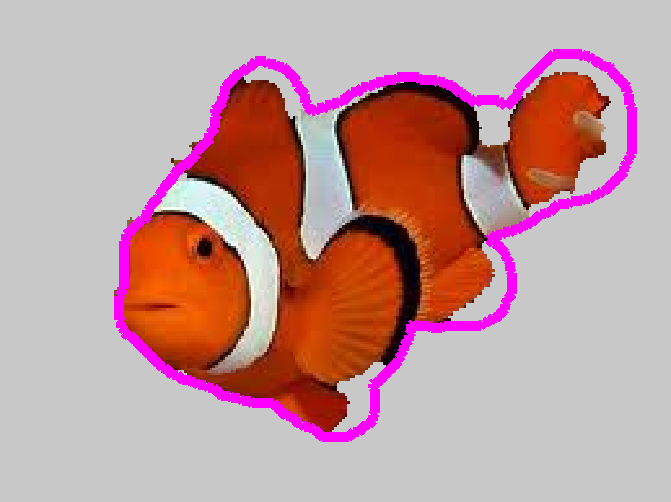}&
  \includegraphics[height=1.6cm]{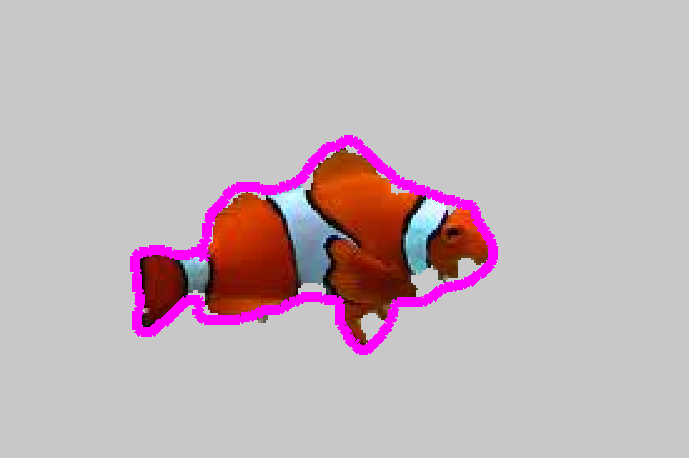}\\
\begin{sideways}\parbox{15mm}{\centering\footnotesize LP~\cite{LiHongDong10LPSeg}}\end{sideways} &
 \includegraphics[height=1.6cm]{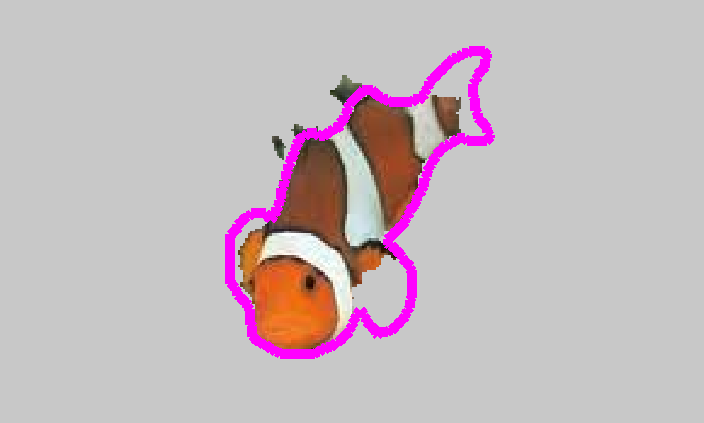}&
 \includegraphics[height=1.6cm]{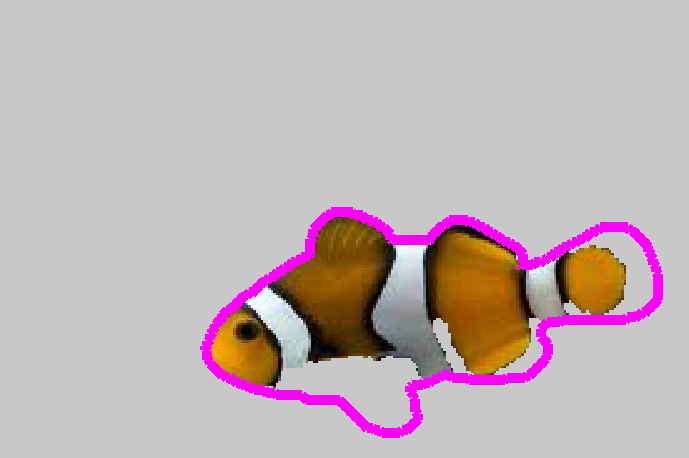}&
 \includegraphics[height=1.6cm]{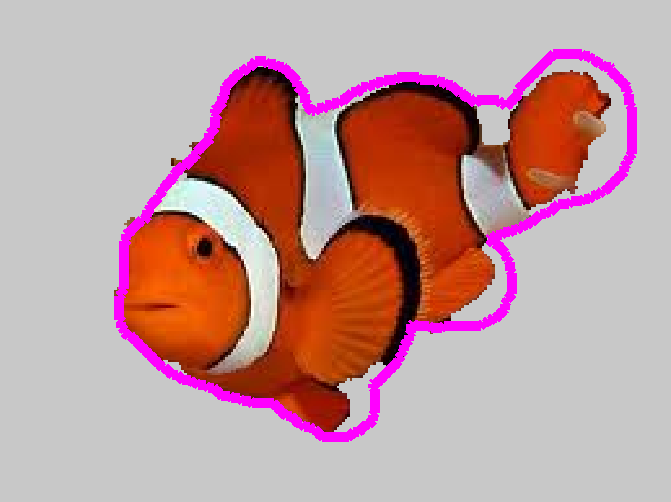}&
 \includegraphics[height=1.6cm]{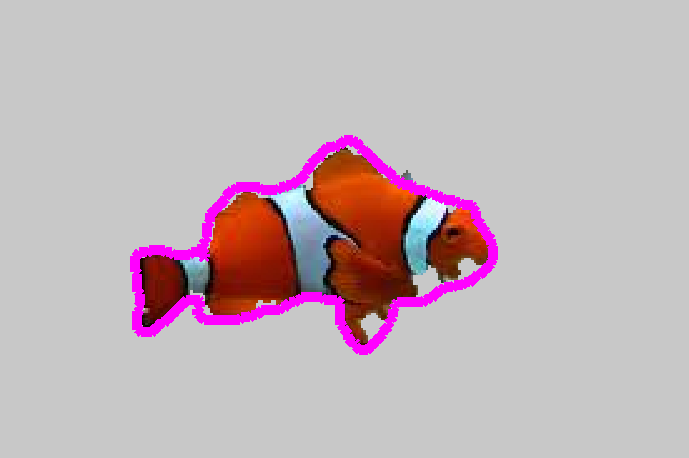}\\
\begin{sideways}\parbox{15mm}{\centering\footnotesize QP~\cite{grady2006randomwalk,sinop2007seeded}}\end{sideways} &
  \includegraphics[height=1.6cm]{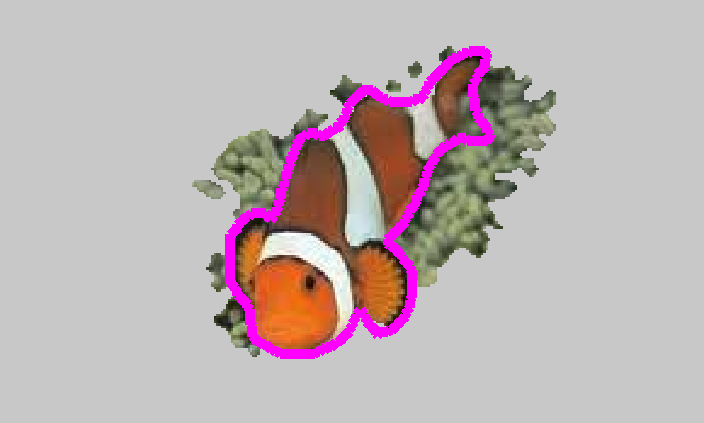}&
  \includegraphics[height=1.6cm]{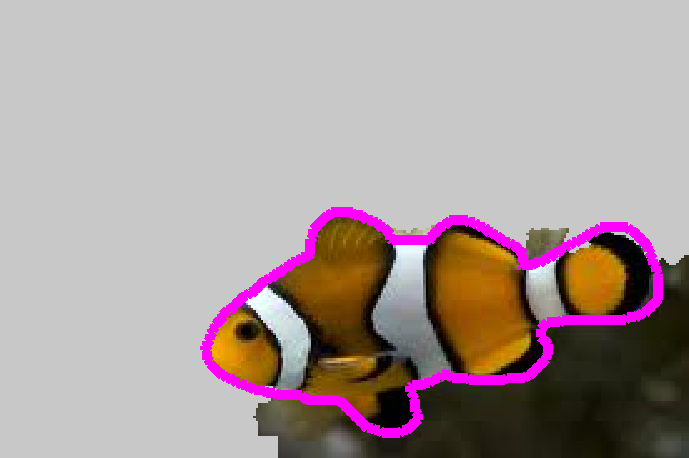}&
  \includegraphics[height=1.6cm]{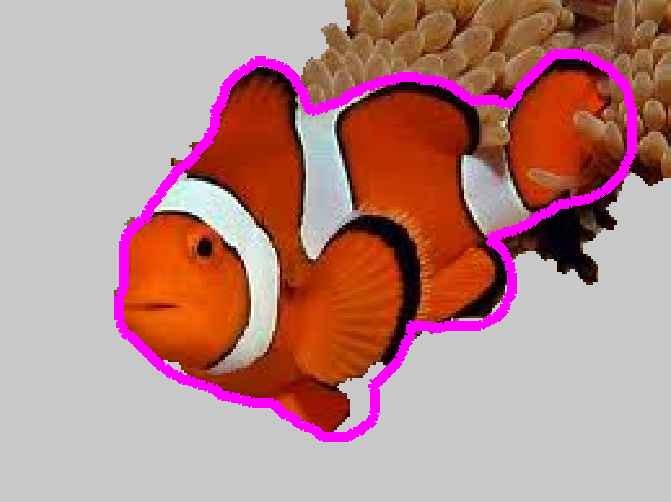}&
  \includegraphics[height=1.6cm]{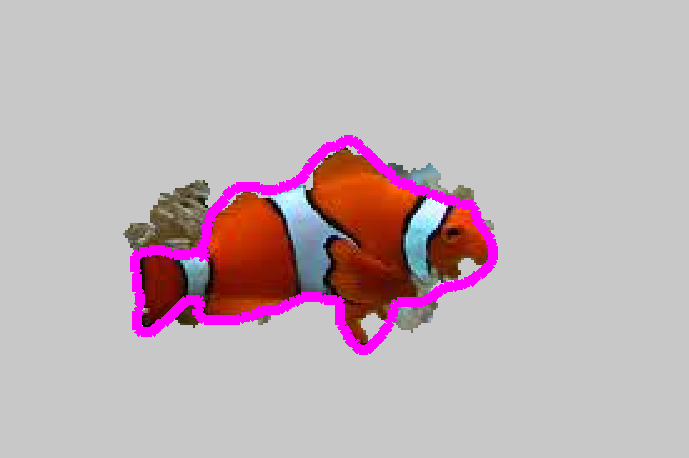}\\
\begin{sideways}\parbox{15mm}{\centering\footnotesize Our method}\end{sideways} &
  \includegraphics[height=1.6cm]{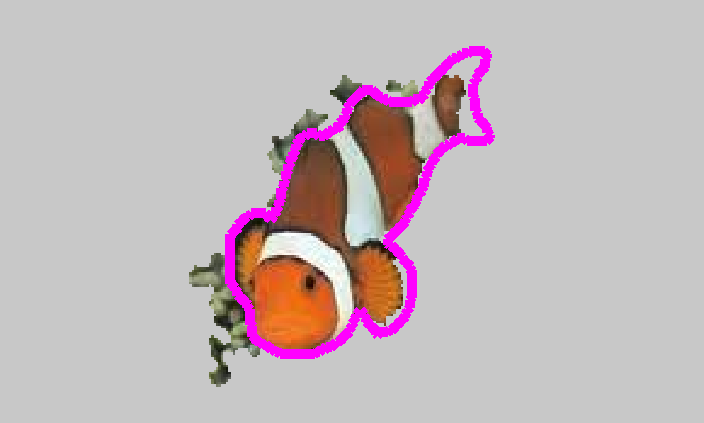}&
  \includegraphics[height=1.6cm]{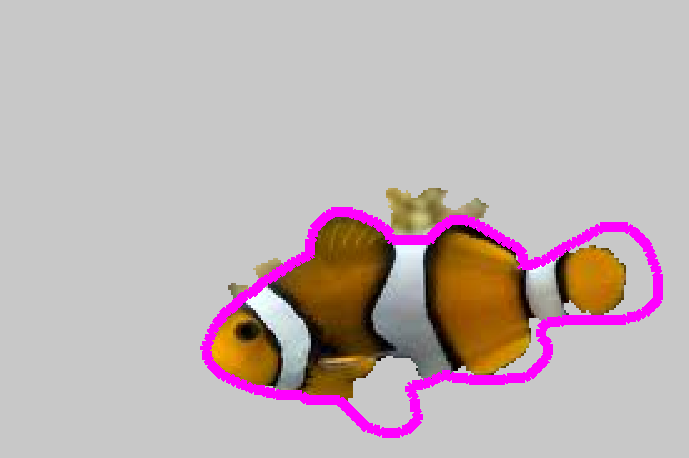}&
  \includegraphics[height=1.6cm]{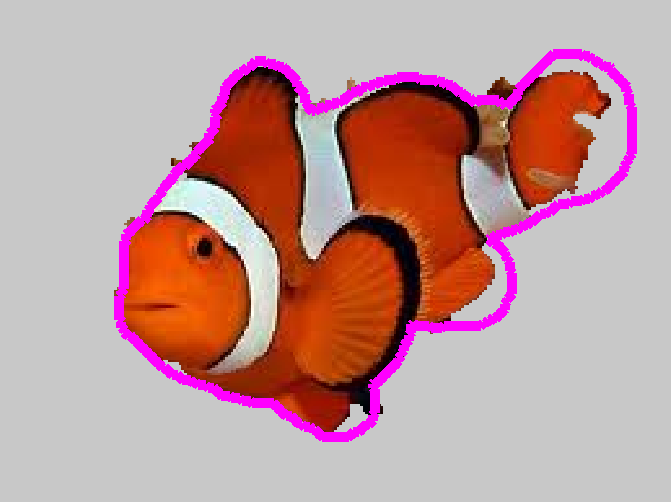}&
  \includegraphics[height=1.6cm]{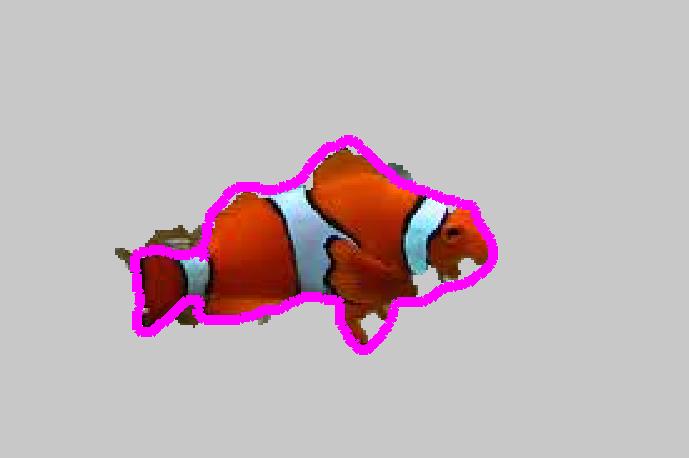}\\
   \end{tabular}
  \end{center}\vspace{-0.6cm}
  \caption{Example results of the seed-initialized interactive segmentation on clownfish dataset. The results are shown as extracted image regions against the ground truth shape contours in purple.}\label{FIG:CMP_fish}
\end{figure}

\begin{figure}[!h]
    \begin{center}
\begin{tabular}
{
@{\hspace{0mm}}c@{\hspace{0.5mm}}c@{\hspace{0.5mm}}c @{\hspace{0.5mm}}c
@{\hspace{0.5mm}}c@{\hspace{1mm}}c@{\hspace{1mm}}c @{\hspace{1mm}}c
@{\hspace{1mm}}c
}
\begin{sideways}\parbox{15mm}{\centering\footnotesize LP~\cite{LiHongDong10LPSeg}}\end{sideways} &
 \includegraphics[height=2cm]{imgs/L1LPSelected/012_Label.png}&
 \includegraphics[height=2cm]{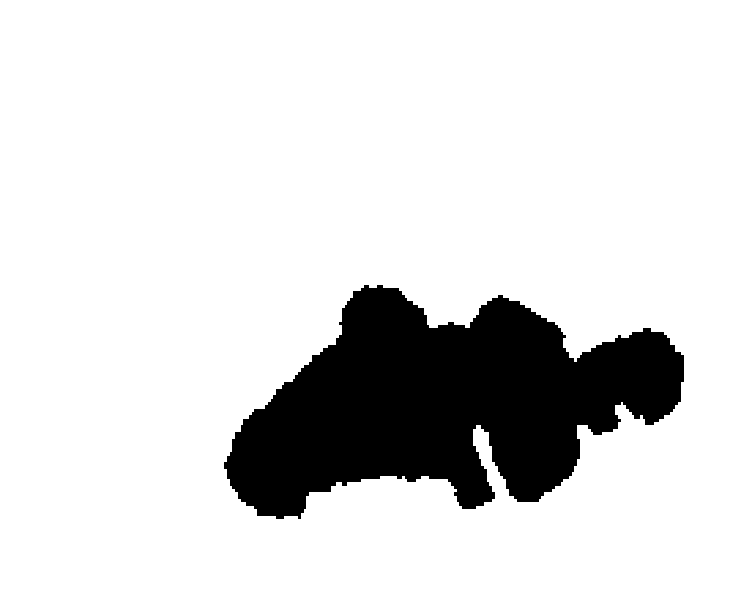}&
 \includegraphics[height=2cm]{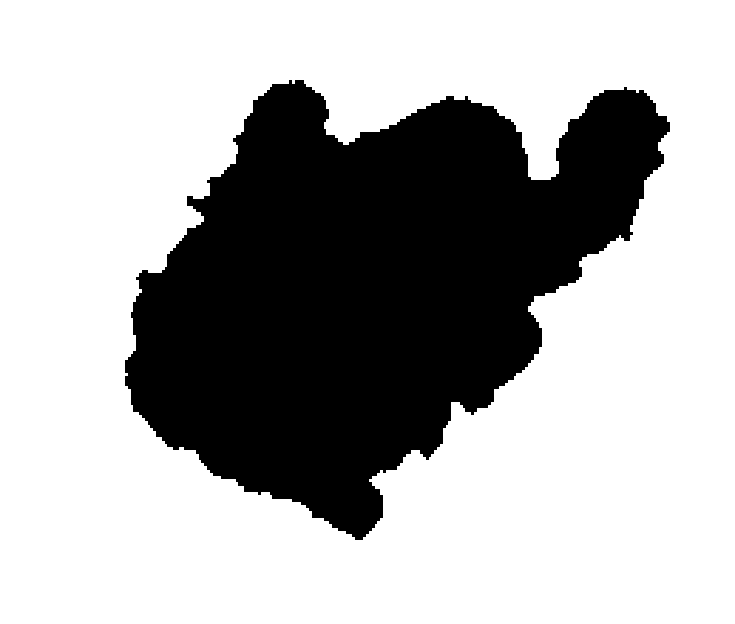}&
 \includegraphics[height=2cm]{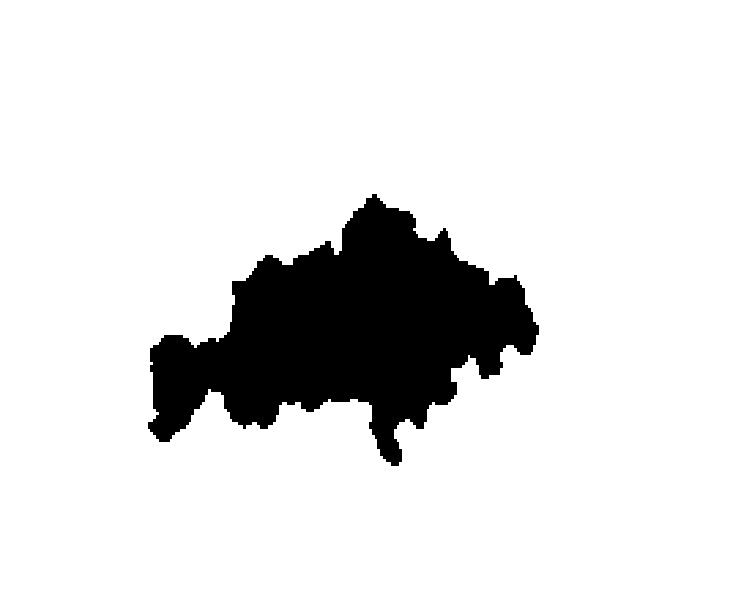}\\
\begin{sideways}\parbox{15mm}{\centering\footnotesize QP~\cite{grady2006randomwalk,sinop2007seeded}}\end{sideways} &
 \includegraphics[height=2cm]{imgs/QPSelected/012_Label.png}&
 \includegraphics[height=2cm]{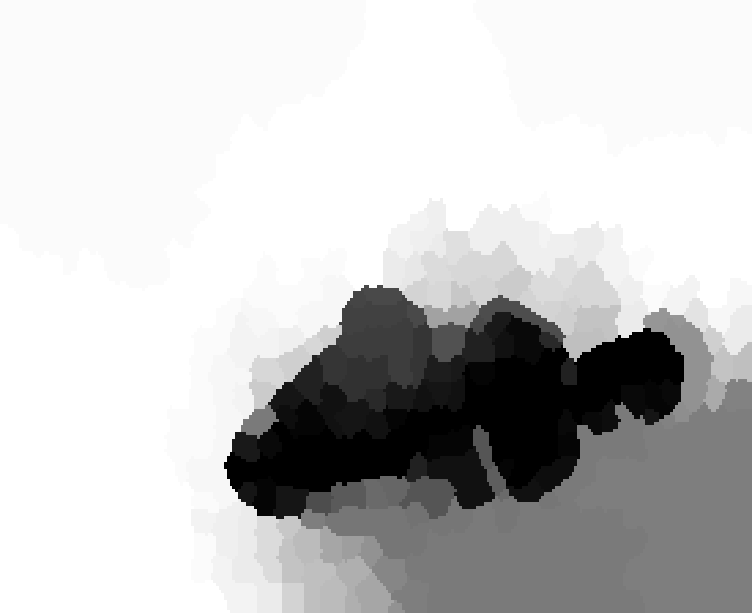}&
 \includegraphics[height=2cm]{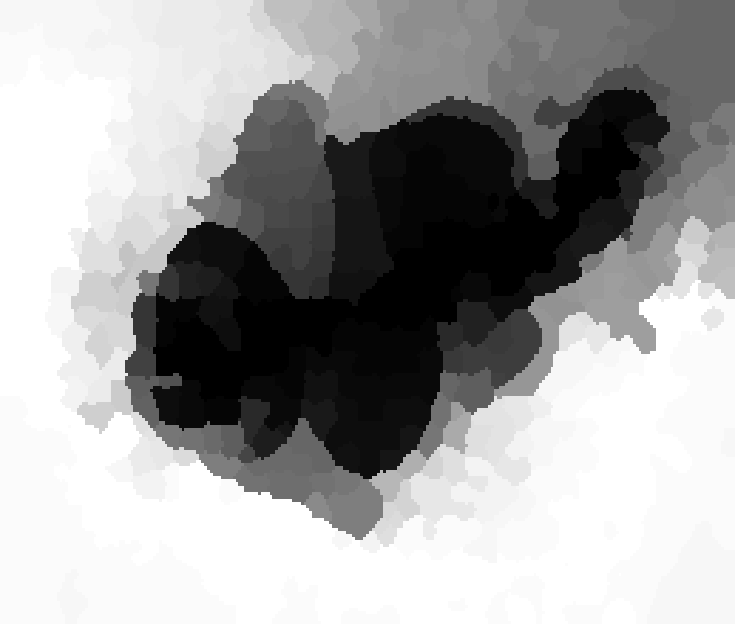}&
 \includegraphics[height=2cm]{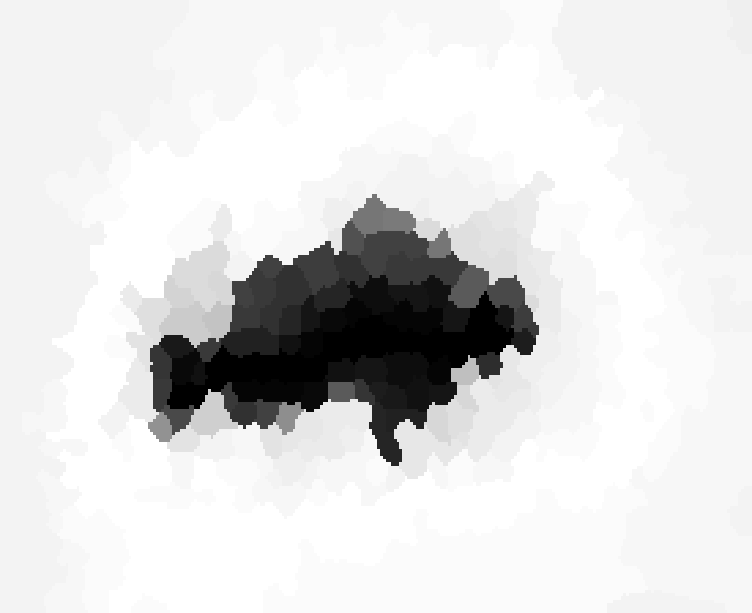}\\
\begin{sideways}\parbox{15mm}{\centering\footnotesize Our method}\end{sideways} &
 \includegraphics[height=2cm]{imgs/L2LPSelected/012_Label.png}&
 \includegraphics[height=2cm]{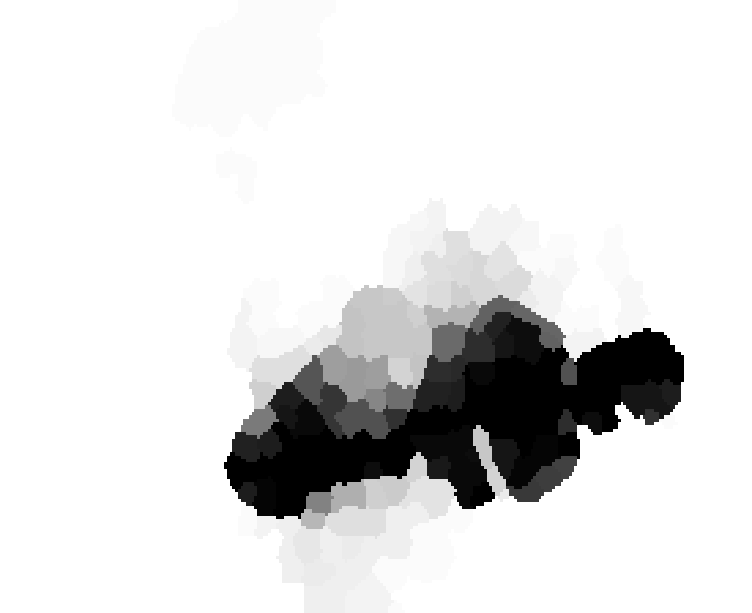}&
 \includegraphics[height=2cm]{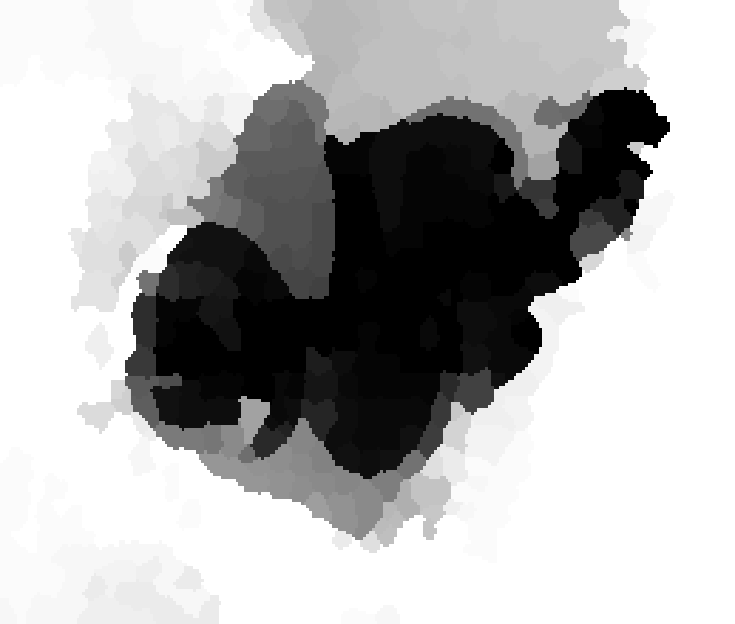}&
 \includegraphics[height=2cm]{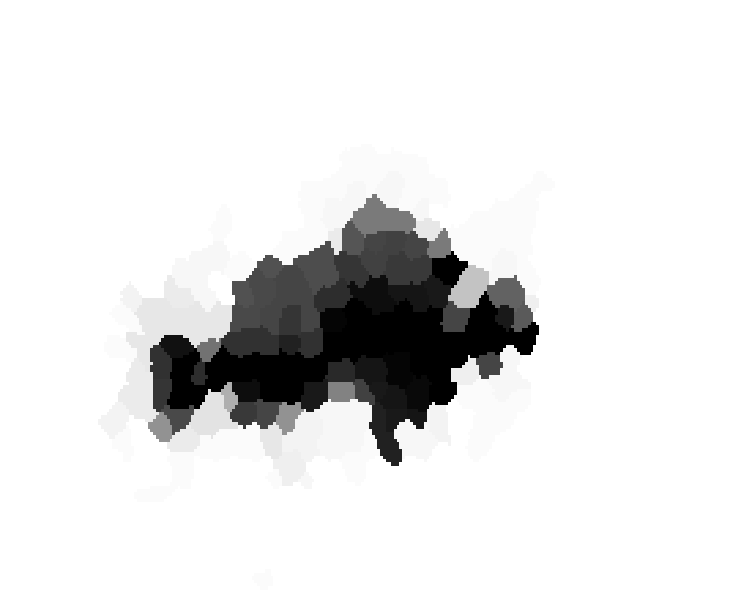}\\
    \end{tabular}
  \end{center}\vspace{-0.7cm}
\caption{Example labels of segmentation results in Fig.~\ref{FIG:CMP_fish}.}\label{FIG:Clabel_fish}
\end{figure}

\begin{figure}
  \centering
  \includegraphics[width=0.6\columnwidth]{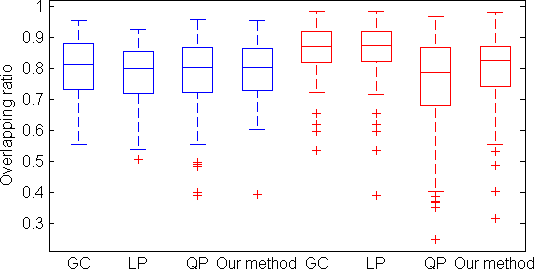}\\\vspace{-0.2cm}
  \caption{Quantitative results of the experiments. Blue boxes are the results for clownfish dataset and the red boxes are the results for Oxford dataset.}\label{FIG:QC_all}
\end{figure}

\paragraph{The clownfish dataset.}
We first present and analyze the experimental results for the clownfish dataset. See Fig.~\ref{FIG:CMP_fish} for example segmentation results and input seeds (refer supplementary materials for additional results). In addition to the manually drawn background seeds, we include the points at the image border as the background seeds in this experiment. From the results, we can see that the results of the conventional LP is very similar to those by graph cuts as expected. A characteristic of them is that they suffer from the small-cut problem. In contrast, QP may produce larger regions due to the possible diffusion of labels at the boundaries. Thus, the resultant regions can be larger than the desired region. Our method compromises the two types of methods and the overall results may outperform the others, e.g., when LP suffers from small-cut problem and/or QP suffers from large-cut problem. We also present the label map of conventional LP, QP and our LP in Fig.~\ref{FIG:Clabel_fish}. As expected, the solutions of LP are binary without thresholding, and the solutions of QP can be over-smooth. The boundaries in the solutions of our LP are clearer than QP, and the solutions are smoother than LP. Quantitative segmentation results of the clownfish dataset are shown in Fig.~\ref{FIG:QC_all}. The results show that QP slightly outperforms the conventional LP on this dataset, and our method slightly outperforms the others. From Table.~\ref{TAB:CTSeedISeg}, we can see that the computational cost of our compact LP model is comparable to QP and requires significantly less computational expenses compared to conventional LP.

\begin{figure*}
    \begin{center}
\begin{tabular}
{
@{\hspace{0mm}}c@{\hspace{0.5mm}}c@{\hspace{0.5mm}}c@{\hspace{0.5mm}}c@{\hspace{0.5mm}}c@{\hspace{0.5mm}}c @{\hspace{0.5mm}}c
@{\hspace{1mm}}c@{\hspace{1mm}}c@{\hspace{1mm}}c@{\hspace{1mm}}c@{\hspace{1mm}}c @{\hspace{1mm}}c
} \begin{sideways}\parbox{15mm}{\centering\footnotesize GC~\cite{Boykov01GraphCut}}\end{sideways} &
  \includegraphics[height=4cm]{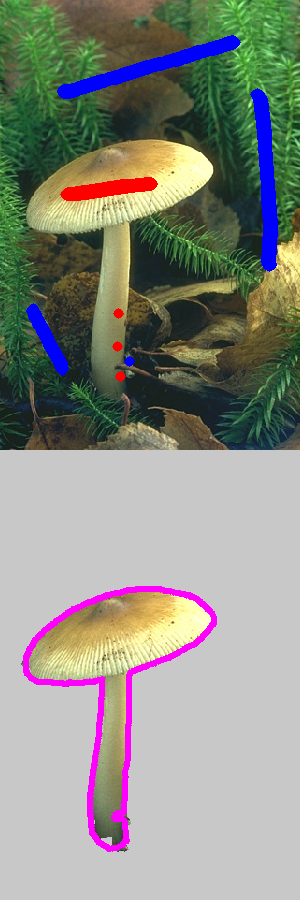}&
  \includegraphics[height=4cm]{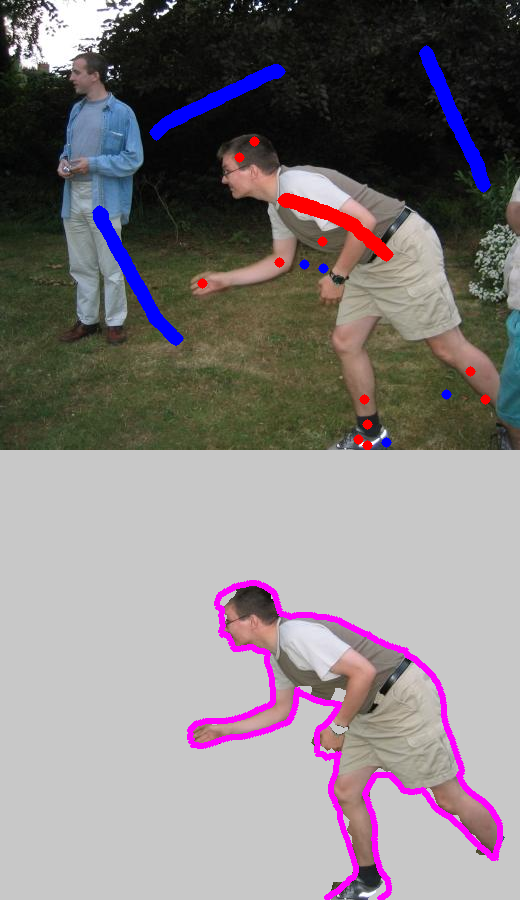}&
  \includegraphics[height=4cm]{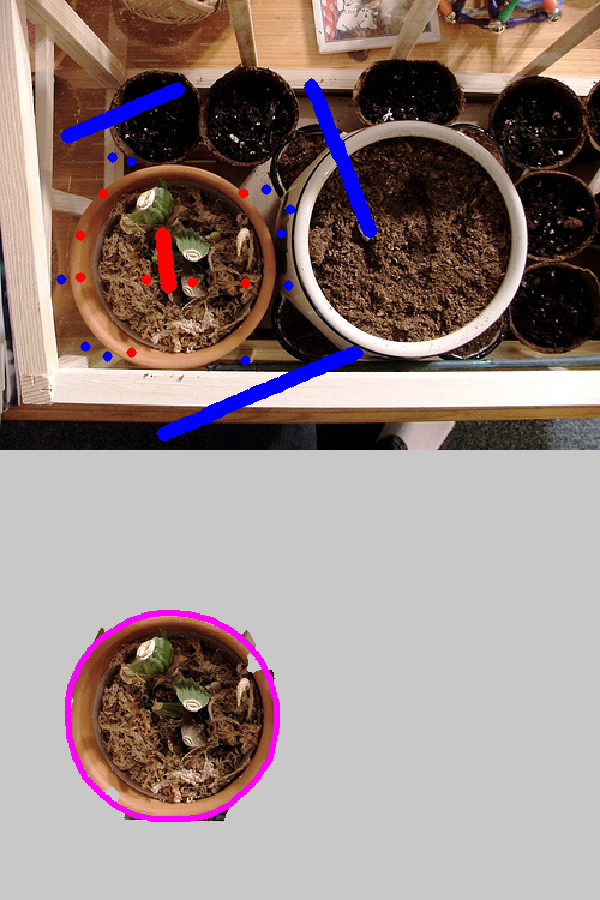}&
  \includegraphics[height=4cm]{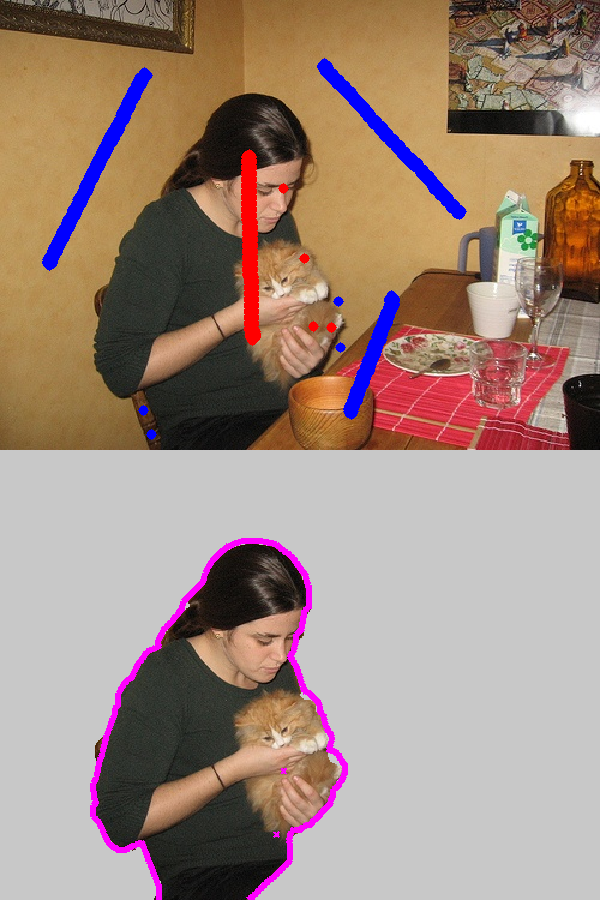}&  \includegraphics[height=4cm]{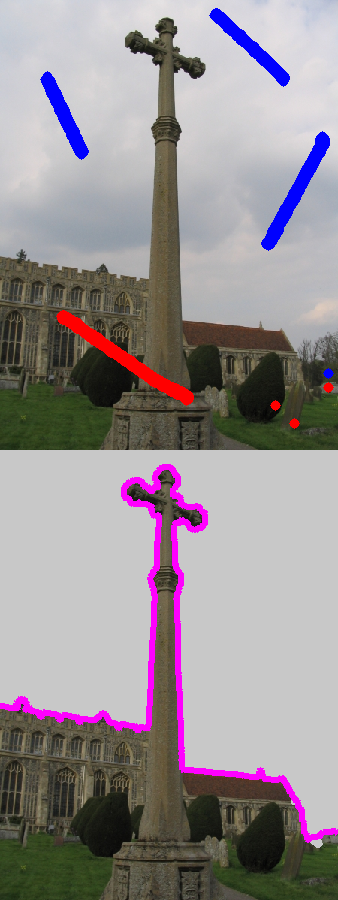}\\
  \begin{sideways}\parbox{15mm}{\centering\footnotesize LP~\cite{LiHongDong10LPSeg}}\end{sideways} &
  \includegraphics[height=4cm]{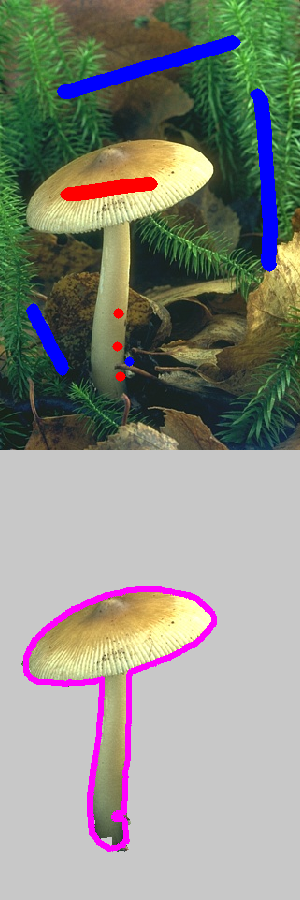}&
  \includegraphics[height=4cm]{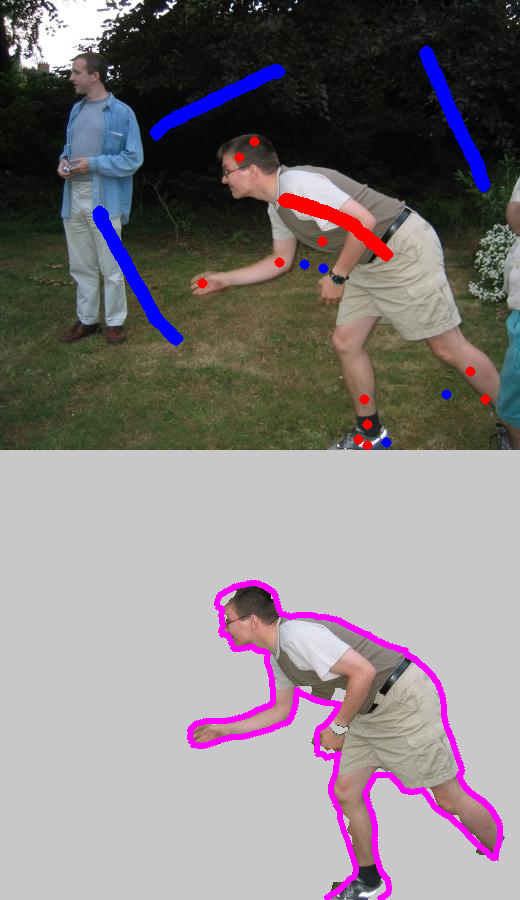}&
  \includegraphics[height=4cm]{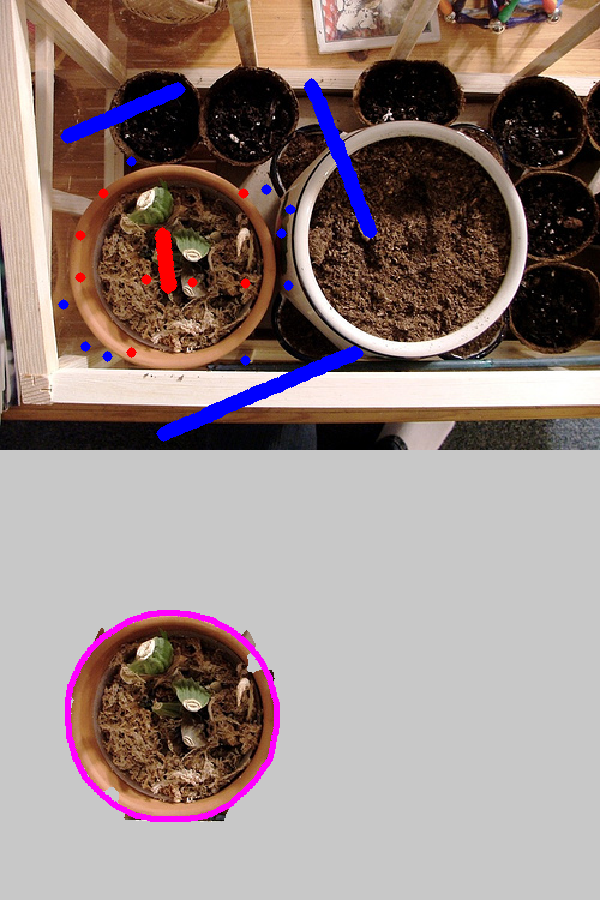}&
  \includegraphics[height=4cm]{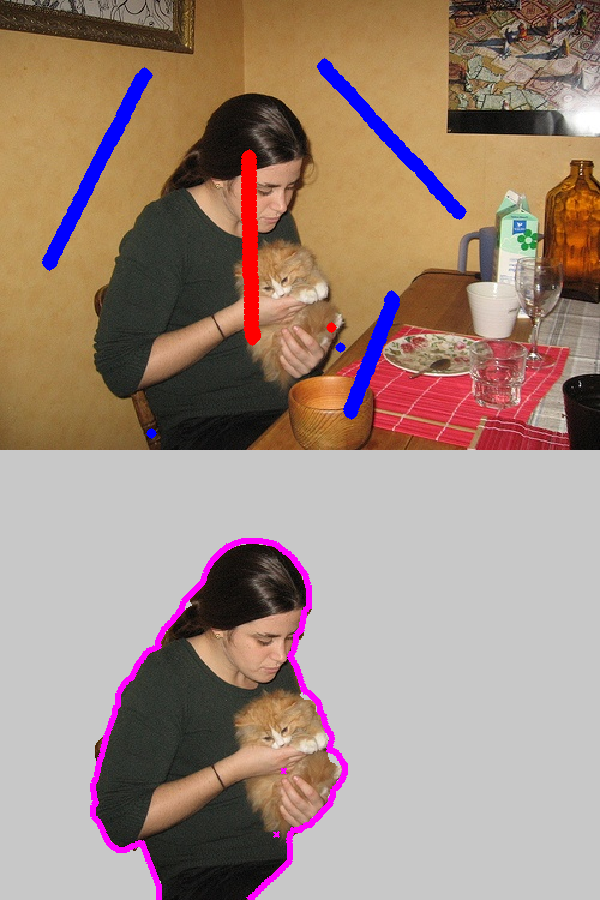}&
  \includegraphics[height=4cm]{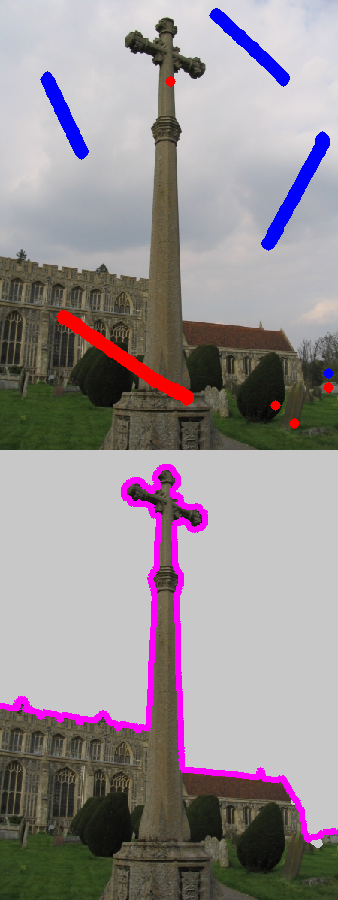}\\
  \begin{sideways}\parbox{15mm}{\centering\footnotesize QP~\cite{grady2006randomwalk,sinop2007seeded}}\end{sideways} &
  \includegraphics[height=4cm]{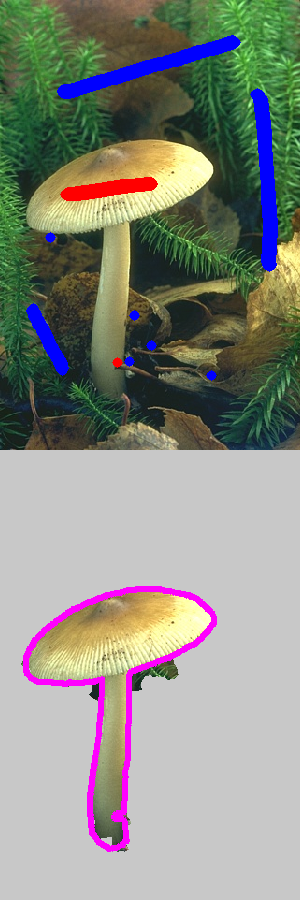}&
  \includegraphics[height=4cm]{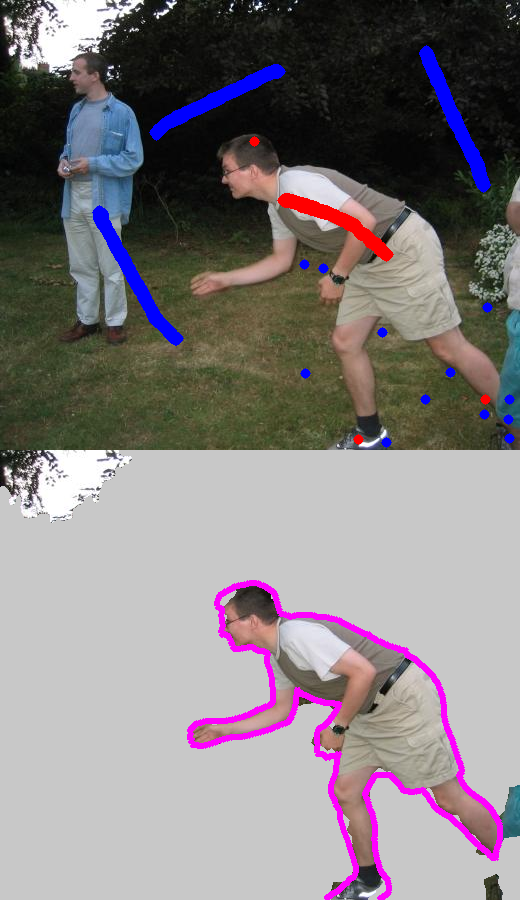}&
  \includegraphics[height=4cm]{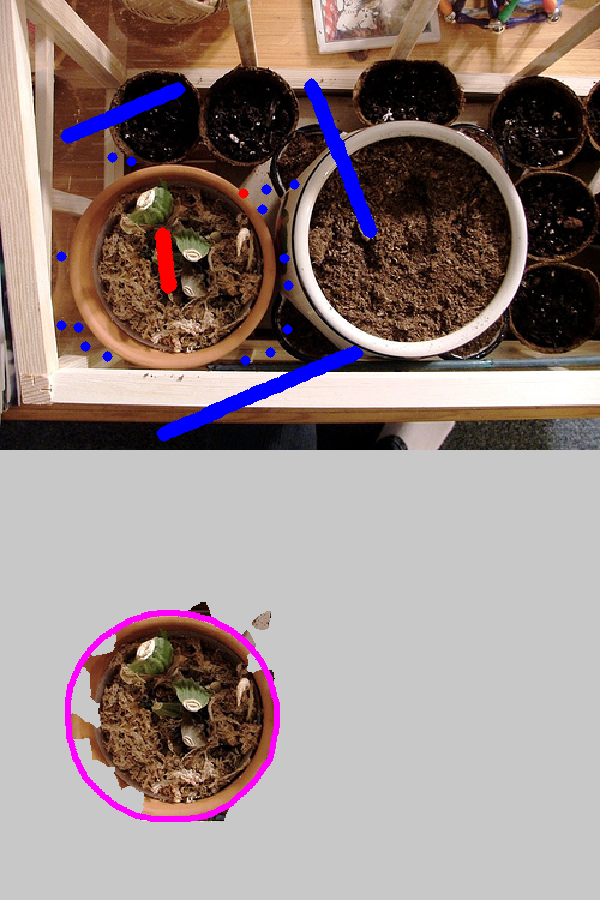}&
  \includegraphics[height=4cm]{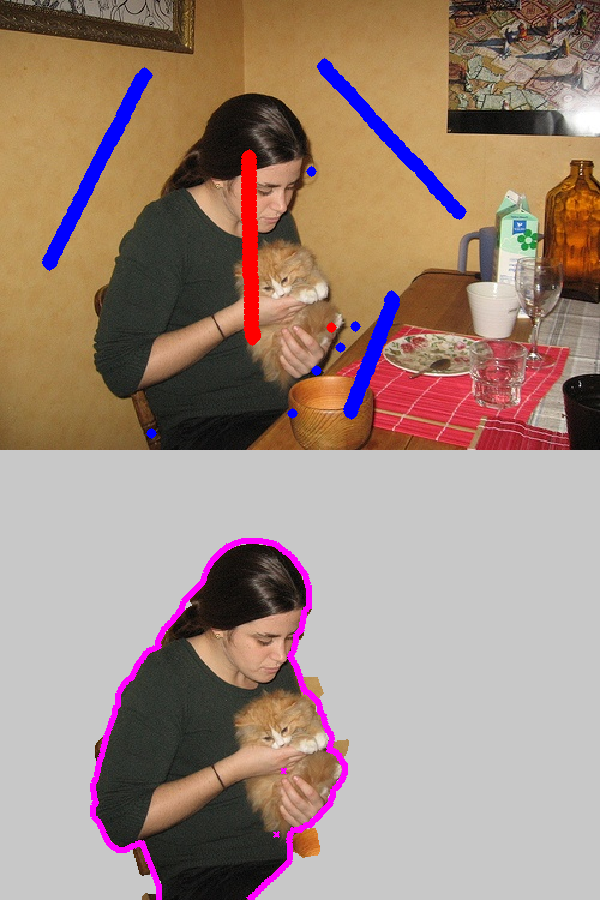}&
  \includegraphics[height=4cm]{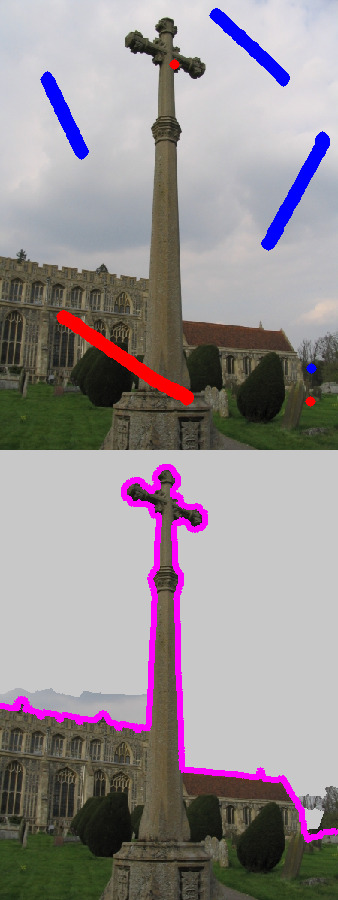}\\
  \begin{sideways}\parbox{15mm}{\centering\footnotesize Our method}\end{sideways} &
  \includegraphics[height=4cm]{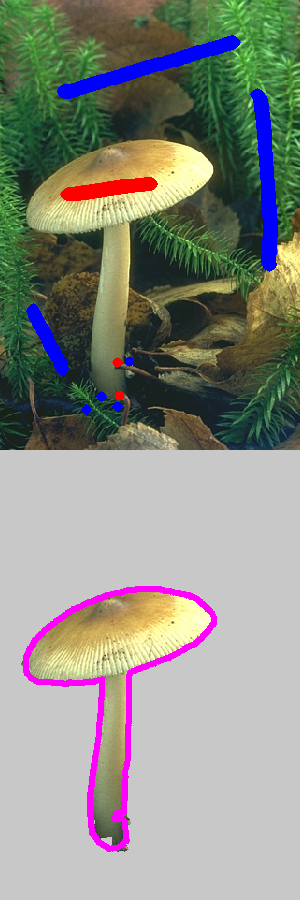}&
  \includegraphics[height=4cm]{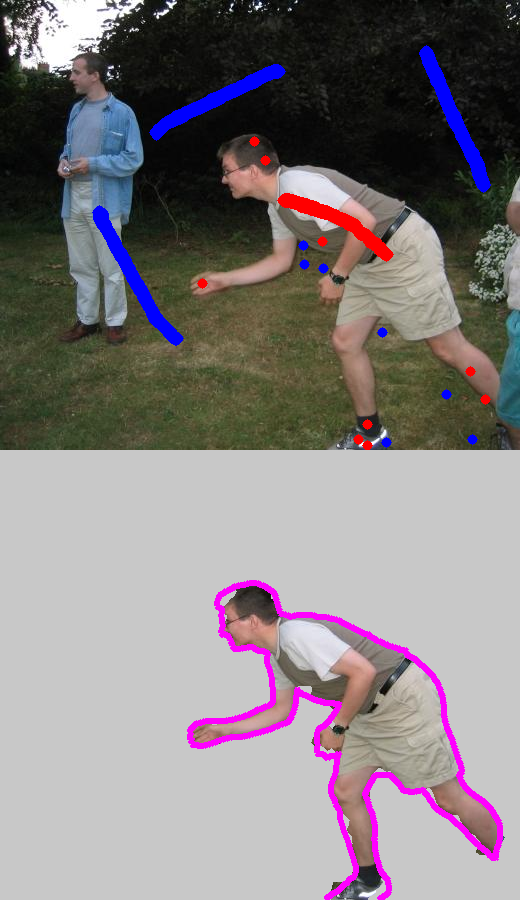}&
  \includegraphics[height=4cm]{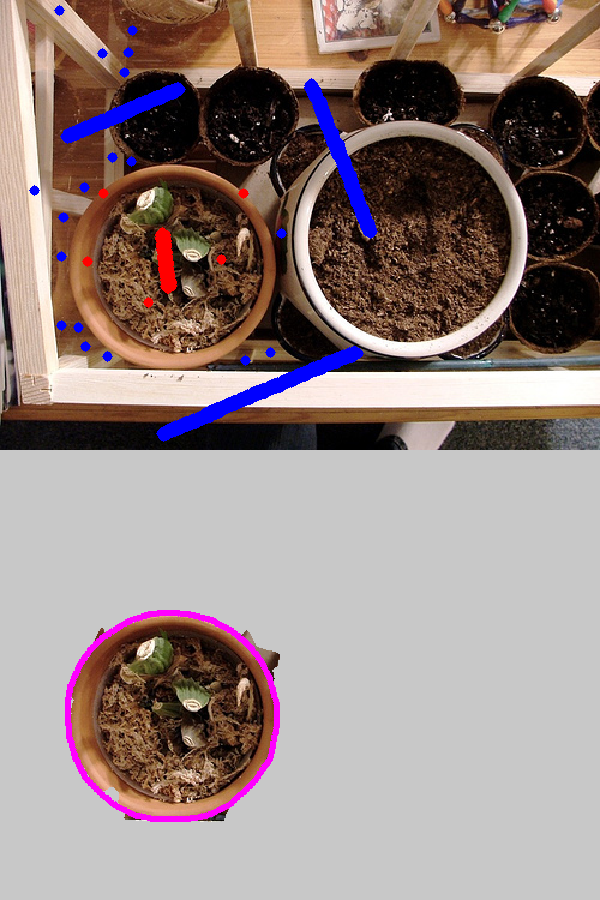}&
  \includegraphics[height=4cm]{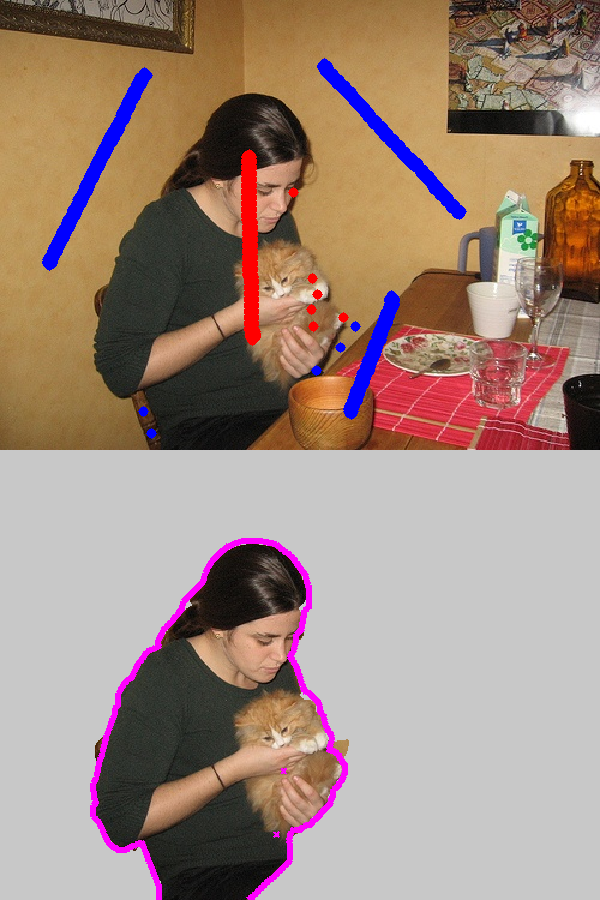}&
  \includegraphics[height=4cm]{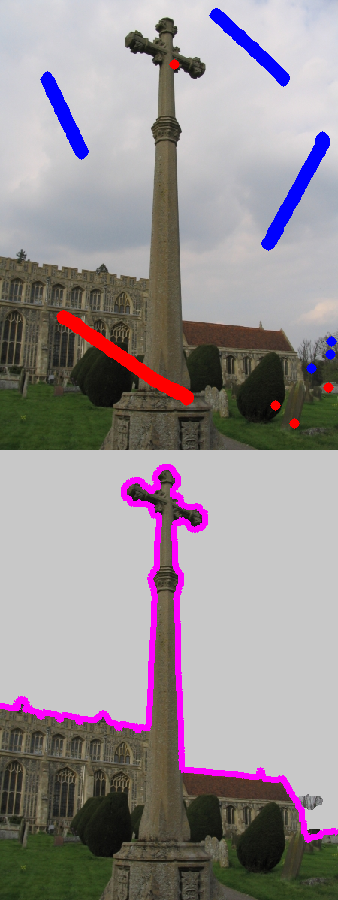}\\
    \end{tabular}
  \end{center}\vspace{-0.5cm}
  \caption{Comparison of segmentation performance on Oxford dataset. The top row show the input images overlaid with input seeds. The bottom rows show extracted image regions against the ground truth shape contours in purple.}\label{FIG:OxfordSeg}
\end{figure*}

\begin{figure}
    \begin{center}
\begin{tabular}
{
@{\hspace{0mm}}c@{\hspace{0.5mm}}c@{\hspace{0.5mm}}c@{\hspace{0.5mm}}c@{\hspace{0.5mm}}c@{\hspace{0.5mm}}c @{\hspace{0.5mm}}c
@{\hspace{1mm}}c@{\hspace{1mm}}c@{\hspace{1mm}}c@{\hspace{1mm}}c@{\hspace{1mm}}c @{\hspace{1mm}}c
}
  \begin{sideways}\parbox{15mm}{\centering\footnotesize LP~\cite{LiHongDong10LPSeg}}\end{sideways} &
  \includegraphics[height=1.8cm]{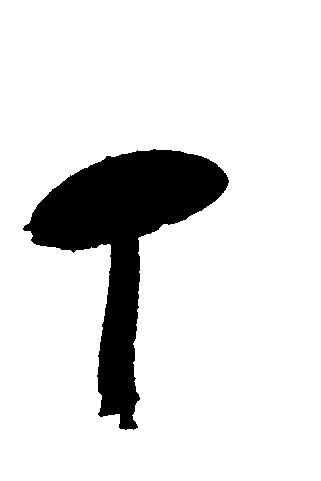}&
  \includegraphics[height=1.8cm]{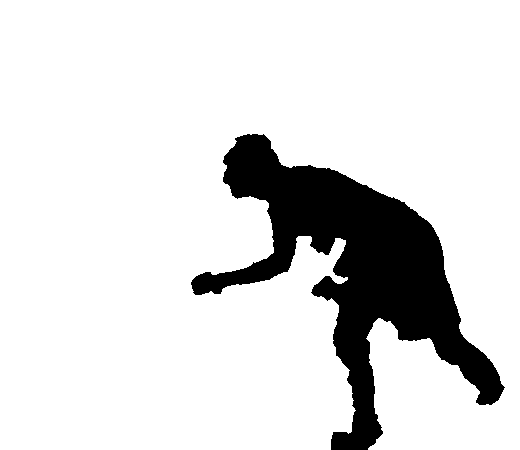}&
  \includegraphics[height=1.8cm]{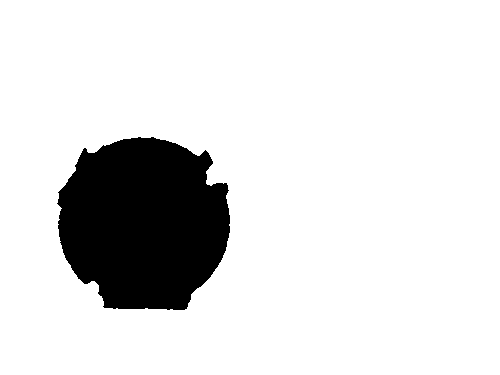}&
  \includegraphics[height=1.8cm]{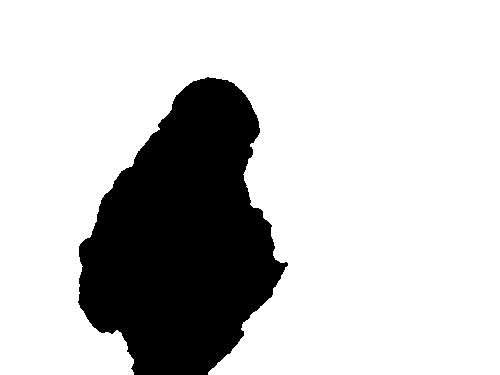}&
  \includegraphics[height=1.8cm]{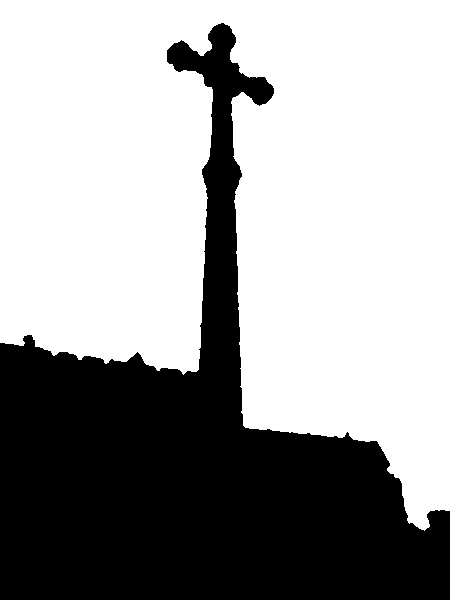}\\
  \begin{sideways}\parbox{15mm}{\centering\footnotesize QP~\cite{grady2006randomwalk,sinop2007seeded}}\end{sideways} &
  \includegraphics[height=1.8cm]{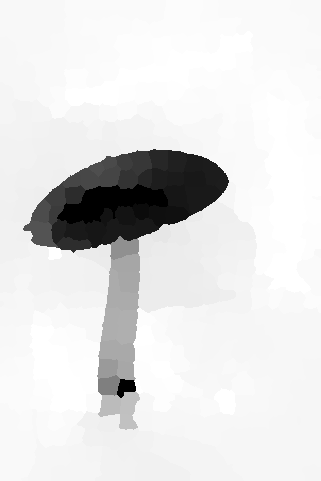}&
  \includegraphics[height=1.8cm]{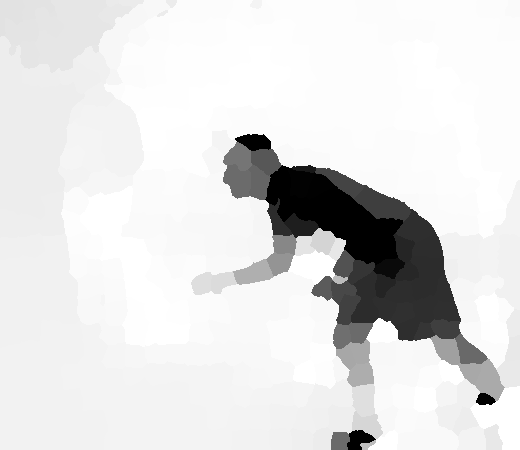}&
  \includegraphics[height=1.8cm]{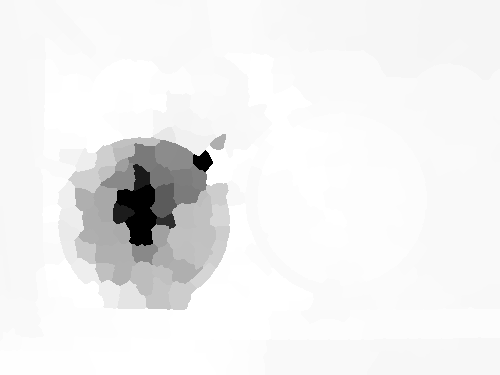}&
  \includegraphics[height=1.8cm]{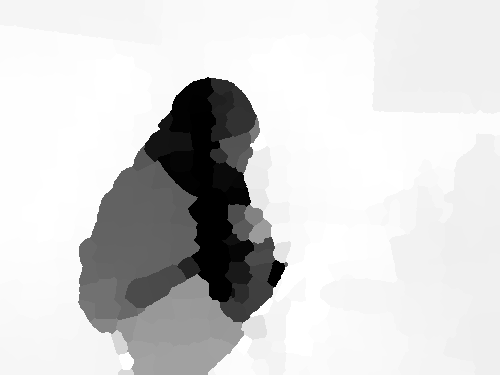}&
  \includegraphics[height=1.8cm]{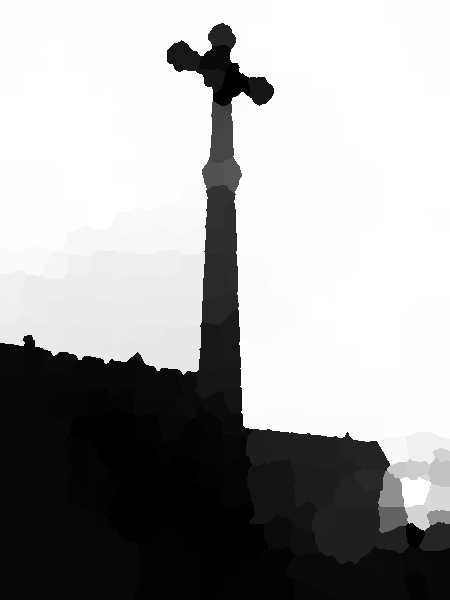}\\
  \begin{sideways}\parbox{15mm}{\centering\footnotesize Our method}\end{sideways} &
  \includegraphics[height=1.8cm]{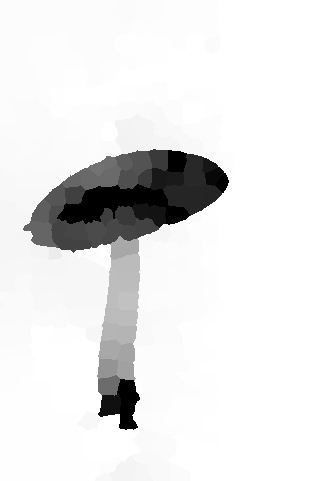}&
  \includegraphics[height=1.8cm]{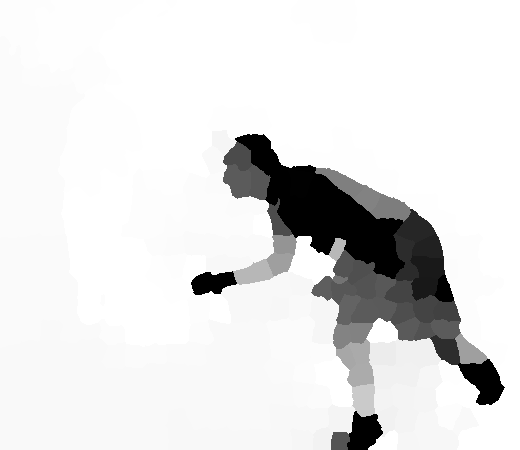}&
  \includegraphics[height=1.8cm]{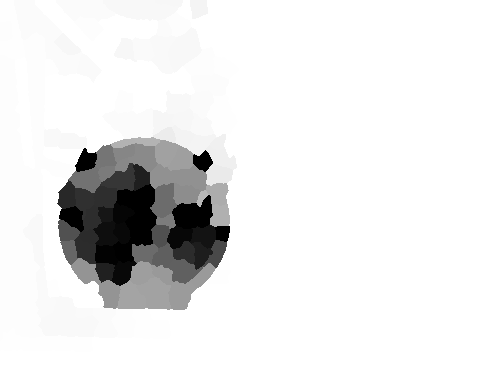}&
  \includegraphics[height=1.8cm]{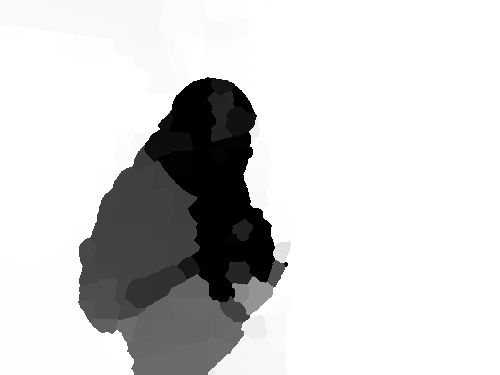}&
  \includegraphics[height=1.8cm]{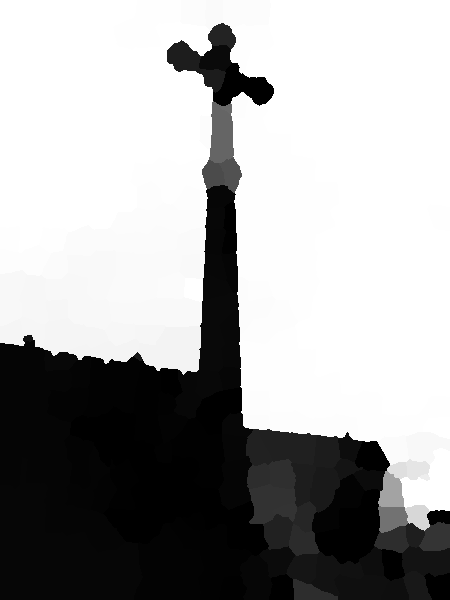}\\
    \end{tabular}
  \end{center}\vspace{-0.7cm}  \caption{Continuous labels before thresholding from LP, QP and our method on example inputs in Fig.~\ref{FIG:OxfordSeg}}\label{FIG:OxfordSeg_labels}
\end{figure}

\paragraph{The Oxford dataset.}
After proof of concept by the clown fish dataset, we validate our model on the Oxford dataset. The user input seeds provided in this dataset are generally insufficient for producing a satisfactory segmentation. We adopt the robotuser~\cite{Gulshan10StarGeoGraphCut} to simulate the additional user interactions.  By increasing the number of interactions, the segmentation results can finally become satisfactory. The maximum number of user interactions is set to 20 in our experiments. See Fig.~\ref{FIG:OxfordSeg} for example results and supplementary materials for additional results. We can observe that GC and LP performs quite alike, while QP may produce larger regions. In most of the situations our methods produce more accurate segmentation results than QP. We present the solutions of QP and our method before thresholding in Fig.~\ref{FIG:OxfordSeg_labels}. The LP produces binary labels as expected, the QP produces smooth labels near the object boundaries and our method produces piecewise smooth labels with relatively clear discontinuities at the boundaries. The quantitative results are shown as red boxes in Fig.~\ref{FIG:QC_all}. The statistics of the computational costs are shown in Table~\ref{TAB:CTSeedISeg}. Very recently, a fast optimization approach has been proposed for solving a similar segmentation model~\cite{WangPCVPR13FastSDP}. However, the computational cost of their approach for 760 superpixels is 2 times of our cost on a PC better than ours.

\begin{figure}
  \centering
  \includegraphics[width=0.6\columnwidth]{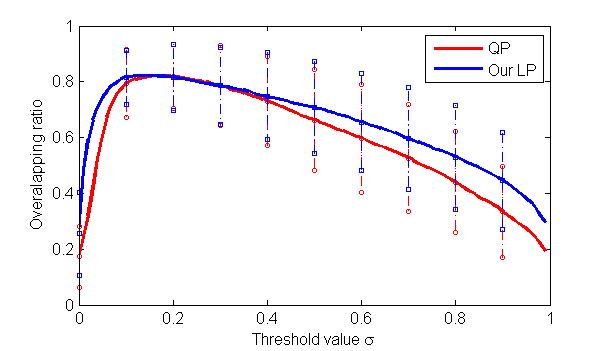}\\\vspace{-0.3cm}
  \caption{Comparison of QP and our LP with different threshold values on the entire Oxford dataset.}\label{FIG:OxfordSeg_thrd}
\end{figure}

To quantitatively reveal the effect of the discontinuity preservability of our method, we further consider the robustness of the segmentation to threshold values. We hypothesize that the continuous labels with clear discontinuities at the boundaries will be robust to different threshold values. Therefore, we generate a vector of 100 threshold values equally spaced in $[0,1]$ for the evaluation. We apply all these threshold values to the continuous labels of QP and our method. Surprisingly, we observe that our method overwhelmingly outperforms the QP for almost all the threshold values in the sense of average overlapping ratio. See Fig.~\ref{FIG:OxfordSeg_thrd} for the plots of mean performance with standard deviation.

\begin{table}[!h]
\caption{Comparison of computational costs.}\label{TAB:CTSeedISeg}
\begin{center}
\begin{tabular}{c| c c c}
  \hline
 &LP \cite{LiHongDong10LPSeg} & QP \cite{grady2006randomwalk,sinop2007seeded} & Our method\\\hline
 Worst-case complexity & $O(N^6)$ & $O(N^3)$ & $ O(N^3)$\\
 Time (s) &$72.35\pm 9.33$ & $1.13\pm 0.48$ & $12.9\pm2.61$ \\ \hline
\end{tabular}
\end{center}
\end{table}

\section{Conclusion}\label{SEC:Concl}
In this paper, we proposed a novel LP relaxation for the binary sub-modular MRF model. Our LP relaxation is based on a novel $l_1^+$-norm minimization in this paper, and it contains significantly fewer variables and constraints compared to the conventional LP. We also show that our $l_1^+$-norm minimization is tightly related to the total variation minimization, according to which we argue that the discontinuities in the solution at the object boundaries can be well preserved. Experimental results show that our method is significantly faster than the conventional LP. Besides, given the same order of computational complexity, our method uniformly outperforms the QP when converting the continuous labels to binary labels using various threshold values on the entire Oxford dataset. Our model may be of use to many other problems modeled by MRF.

{\small
\bibliographystyle{plain}
\bibliography{Matching,MRFseg,LevelSetActiveContours,math,Otherseg}
}

\newpage
\renewcommand\thesection{\Alph{section}}
\setcounter{section}{0}
\renewcommand{\theequation}{\Alph{section}-\arabic{equation}}
\setcounter{equation}{0}  
\section{Appendix}
In this supplementary material, we include the lengthy proofs, derivations and additional experimental results that we omitted in the paper due to the page limit.

To evaluate the performance gain in terms of computation, we perform the conventional LP and our proposed LP on matlab GPU for synthetic data. In this experiment, we randomly generate the model parameters for the Eq. (3), we factorize the boundary term to arrive at Eq. (13) and we apply the interior point method to solving both the LP problems. We are unable to experiment on images due to the limit on our hardware. Our graphics card is the mobile NVIDIA Geforce 780M with 3GB graphics memory on our laptop. It thus does not allow the GPU implementation for the conventional LP on images containing hundreds of superpixels. Let's consider 200 superpixels in an image. We will have a constraint matrix containing $(200+200\times 200)\times(200+2\times 200\times200)=3224040000$ elements. Since the matlab GPU only supports full matrix, we will need 24 GB graphics memory for storing this full matrix, which is frankly impossible.

\begin{figure}
  \centering
  \includegraphics[width=0.6\columnwidth]{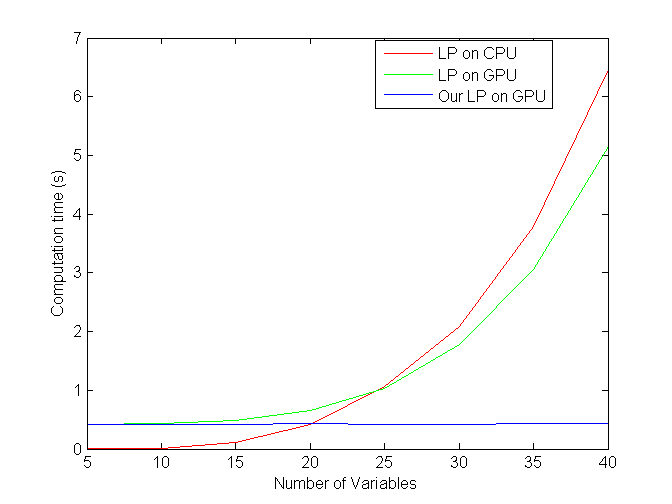}\\\vspace{-0.3cm}
  \caption{Comparison of computational times on GPU.}\label{FIG:GPU}
\end{figure}

The average computational costs on GPU are shown in Fig. \ref{FIG:GPU}. We also plot the baseline computational cost of LP on CPU. It can be seen that the computational cost of LP on GPU tends to be lower than that of the LP on CPU when the number of variables increase. It is interesting to see that the computational cost of our LP on GPU remains steady w.r.t. the number of variables.

\subsection{Proofs and derivations}
\subsubsection{Derivation of Eq. (9)}
\begin{equation}
\begin{split}
\mathbf{B}(\mathcal{L}) &= \sum_{j,j'} E_{jj'}{B}_{jj'}|\mathcal{L}_j-\mathcal{L}_{j'}|^2 \\
&=\sum_{j,j'} w_{jj'}(\mathcal{L}^2_j+\mathcal{L}^2_{j'}-2\mathcal{L}_j\mathcal{L}_{j'}) \\
&=\sum_{j}\mathcal{L}^2_{j} \sum_{j'} w_{jj'}+\sum_{j'}\mathcal{L}^2_{j'}\sum_iw_{jj'}\\
&\hspace{10pt}-2\sum_{j,j'}w_{jj'}\mathcal{L}_j\mathcal{L}_{j'}\\
&=\sum_{j}\mathcal{L}^2_j \bar{w}_{j}+\sum_{j'}\mathcal{L}^2_{j'}\hat{w}_{j'}-2\sum_{j,j'}w_{jj'}\mathcal{L}_j\mathcal{L}_{j'}\\
&=\mathcal{L}^T\mathbf{\widetilde{W}}\mathcal{L}
\end{split}
\end{equation}
where $\mathbf{\widetilde{W}}=diag(\bar{w})+diag(\hat{w})-2\mathbf{W}$, $\mathbf{W}=[w_{jj'}]=[E_{jj'}{B}_{jj'}]$.
\subsubsection{Proof of Proposition 4.1}
\begin{proof}
The definition of $\mathbf{\widetilde{W}}$ is as follows.
\begin{equation}
\mathbf{\widetilde{W}} = diag(\bar{w})+diag(\hat{w})-2\mathbf{W}
\end{equation}
where
\begin{equation}
\bar{w}_j=\sum_k w_{jl} = \sum_k w_{lj'} = \hat{w}_{j'}, \hbox{ if } j=j'.
\end{equation}
In short $diag(\bar{w}) = diag(\hat{w})$. Note that $w_{jj'}=0$ for $j=j'$. Hence, we have the following.
\begin{equation}
\mathbf{\widetilde{W}}_{jj'} = \left\{\begin{array}{cc}
                                       2\bar{w}_j, & \hbox{ for } j=j' \\
                                       -2w_{jj'}, & \hbox{ otherwise}
                                     \end{array}\right.
\end{equation}
Therefore, matrix $\mathbf{\widetilde{W}}$ is a symmetric diagonal dominant matrix, and the diagonal elements are nonnegative. Such matrix is a positive semi-definite matrix.\qed
\end{proof}

\subsubsection{Equivalence between $L_1$ norm minimization and linear programming}
\begin{equation}\label{EQ:l1norm2LP}
\begin{split}
\sum_i |{y}_i| =   \min_{y^+} &~~~~ \sum_i y^+_i\\
                                     \hbox{s.t.} &~~~~ \forall i, -y^+_i\leq y_i\leq y^+_i\\
                                     &~~~~ \forall i, y^+_i>0.
\end{split}
\end{equation}

\subsubsection{Proof of Theorem 4.2}
\begin{proof}
Substituting $[\mathrm{diag}(\vec{w})\mathbf{D}] = \mathbf{Q}^{N^2\times N}\mathbf{R}^{N\times N}$ into Eq.(4), we obtain the following form of the boundary term.
\begin{equation}
\mathbf{B}_{l_1}(\mathcal{L}) = \|\mathbf{Q}\mathbf{R}\mathcal{L}\|_{l_1}
\end{equation}
where we applied the QR factorization. The $l_2$ relaxation of this form will lead to
\begin{equation}
\begin{split}
\mathbf{B}_{l_2}(\mathcal{L}) &= \big(\mathcal{L}^T\mathbf{R}^T\mathbf{Q}^T\mathbf{Q}\mathbf{R}\mathcal{L}\big)^{1/2}\\
&= \big(\mathcal{L}^T\mathbf{R}^T\mathbf{R}\mathcal{L}\big)^{1/2}\\
&= \|\mathbf{R}\mathcal{L}\|_{l_2}
\end{split}
\end{equation}

The corresponding $l_1^+$-norm minimization is therefore the following
\begin{equation}
\mathbf{B}_{l_1^+}(\mathcal{L}) = \|\mathbf{R}\mathcal{L}\|_{l_1}
\end{equation}
Note that the Cholesky decomposition is unique and $\mathbf{R}$ is upper-triangular. We can conclude that $\mathbf{U}=\mathbf{R}$.\qed
\end{proof}

\subsubsection{Proof of Theorem 4.5}
\begin{proof}
We prove the left hand side first.
\begin{equation}
\begin{split}
\|\mathrm{diag}(\vec{w})\mathbf{D}\mathcal{L}\|_{l_1}&=\|\mathbf{Q}\mathbf{U}\mathcal{L}\|_{l_1}\\
&\leq\|\mathbf{Q}\|_{l_1}\|\mathbf{U}\mathcal{L}\|_{l_1}\\
&\Leftrightarrow\\
{1\over\|\mathbf{Q}\|_{l_1}}\|\mathrm{diag}(\vec{w})\mathbf{D}&\mathcal{L}\|_{l_1}\leq\|\mathbf{U}\mathcal{L}\|_{l_1}
\end{split}
\end{equation}
where we have replaced $\mathbf{R}$ with $\mathbf{U}$. The right hand side is likewise.
\begin{equation}
\begin{split}
\|\mathbf{U}\mathcal{L}\|_{l_1}&=\|\mathbf{Q}^T\mathbf{Q}\mathbf{U}\mathcal{L}\|_{l_1}\\
&\leq\|\mathbf{Q}^T\|_{l_1}\|\mathbf{Q}\mathbf{U}\mathcal{L}\|_{l_1}\\
&=\|\mathbf{Q}^T\|_{l_1}\|\mathrm{diag}(\vec{w})\mathbf{D}\mathcal{L}\|_{l_1},
\end{split}
\end{equation}
which completes the proof.\qed
\end{proof}

\subsection{Extra experimental results}
\begin{figure*}
    \begin{center}
\begin{tabular}
{
@{\hspace{0mm}}c@{\hspace{0.5mm}}c@{\hspace{0.5mm}}c@{\hspace{0.5mm}}c@{\hspace{0.5mm}}c@{\hspace{0.5mm}}c @{\hspace{0.5mm}}c
@{\hspace{1mm}}c@{\hspace{1mm}}c@{\hspace{1mm}}c@{\hspace{1mm}}c@{\hspace{1mm}}c @{\hspace{1mm}}c
} \begin{sideways}\parbox{15mm}{\centering\footnotesize GC~\cite{Boykov01GraphCut}}\end{sideways} &
  \includegraphics[height=3.5cm]{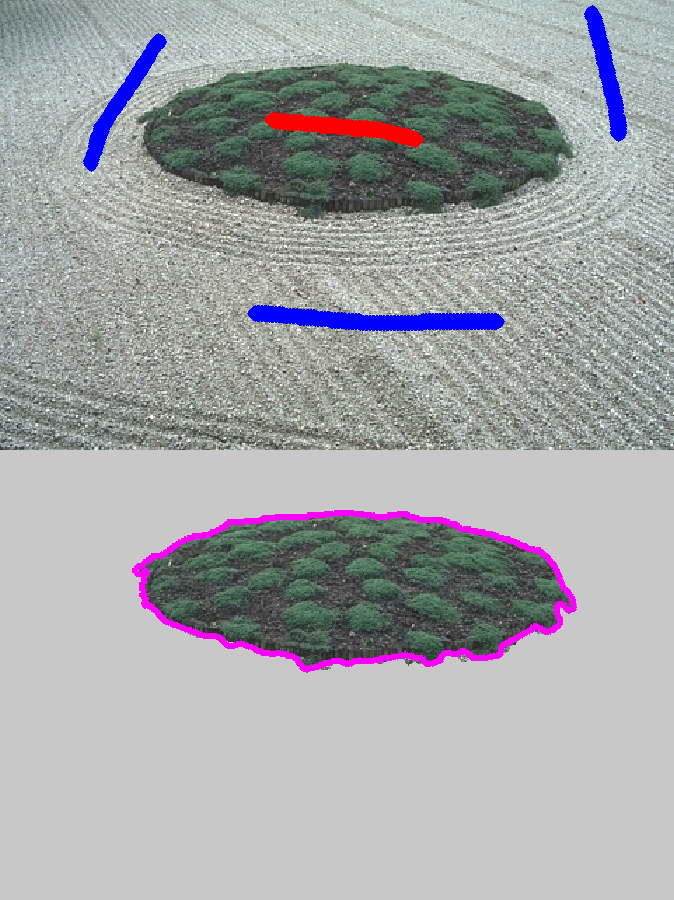}&
  \includegraphics[height=3.5cm]{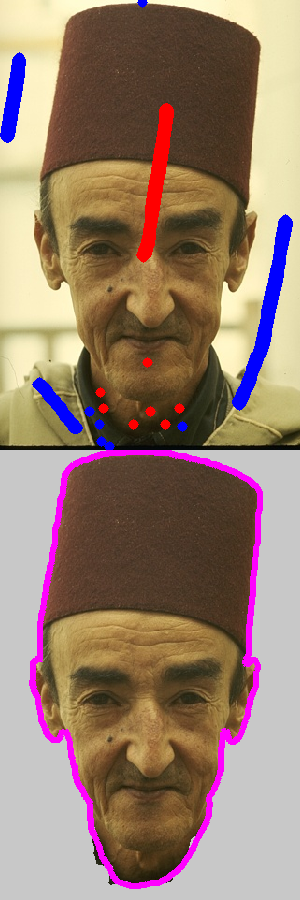}&
  \includegraphics[height=3.5cm]{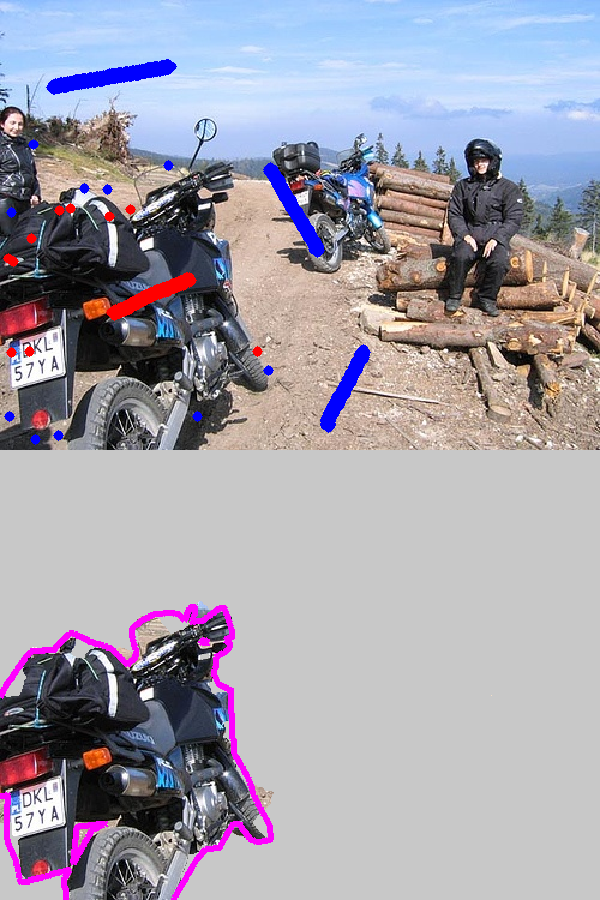}&
  \includegraphics[height=3.5cm]{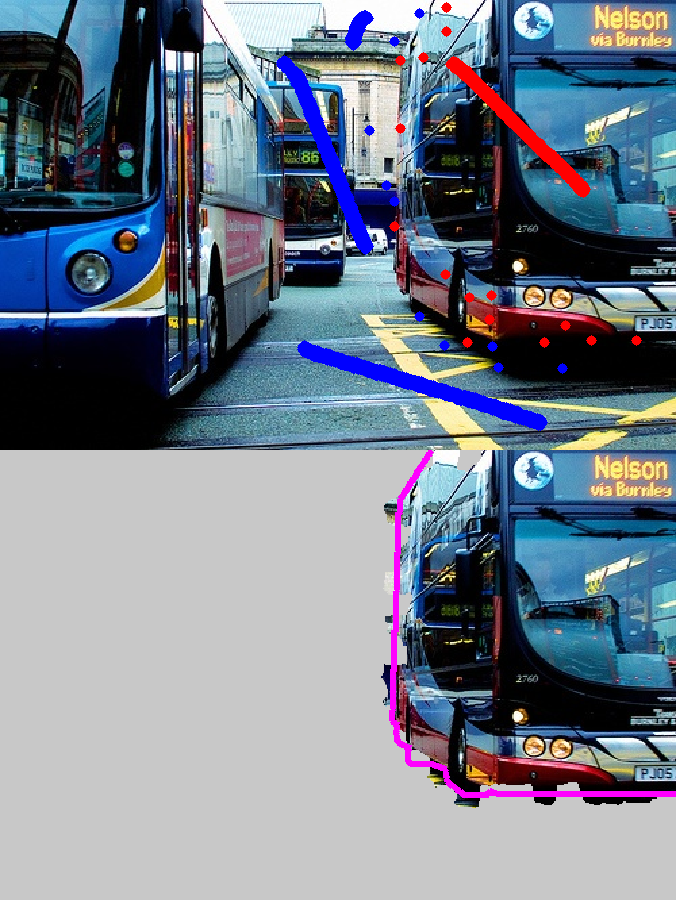}&  \includegraphics[height=3.5cm]{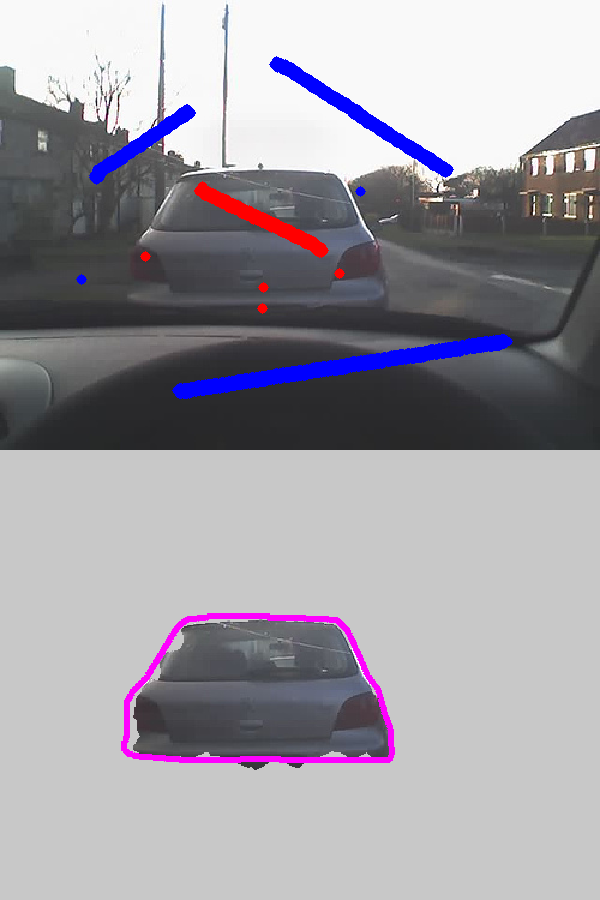}\\
  \begin{sideways}\parbox{15mm}{\centering\footnotesize LP~\cite{LiHongDong10LPSeg}}\end{sideways} &
  \includegraphics[height=3.5cm]{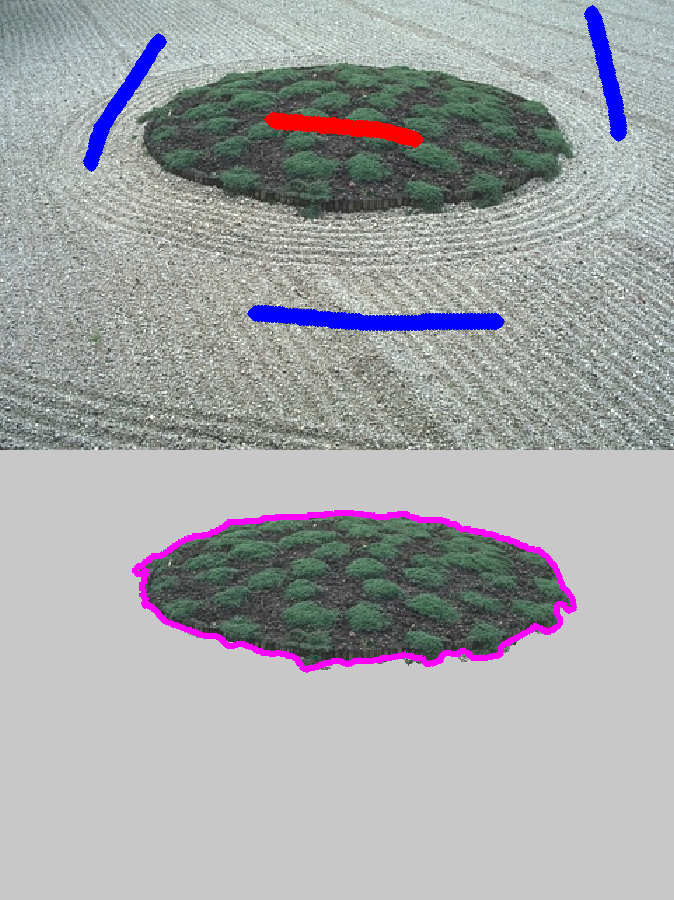}&
  \includegraphics[height=3.5cm]{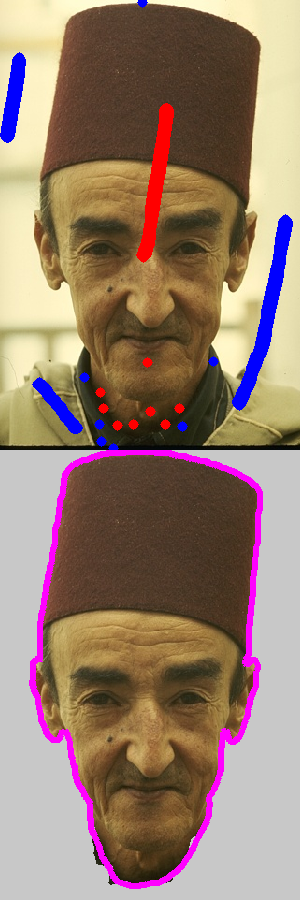}&
  \includegraphics[height=3.5cm]{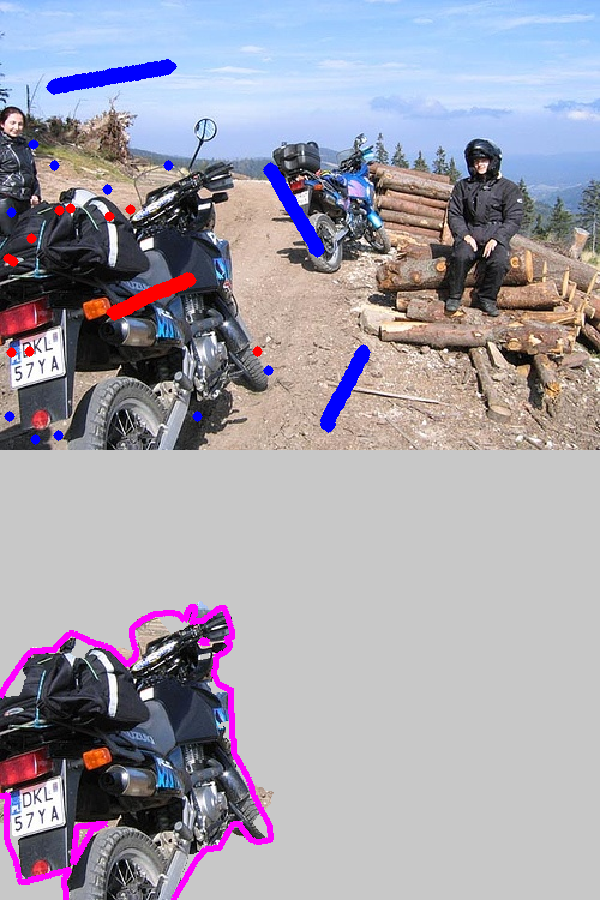}&
  \includegraphics[height=3.5cm]{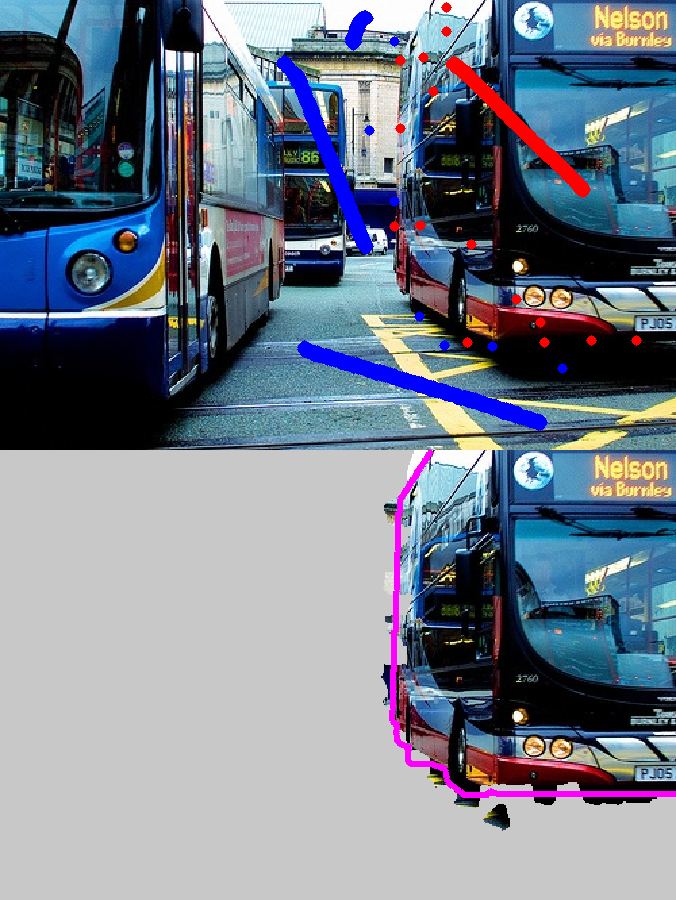}&
  \includegraphics[height=3.5cm]{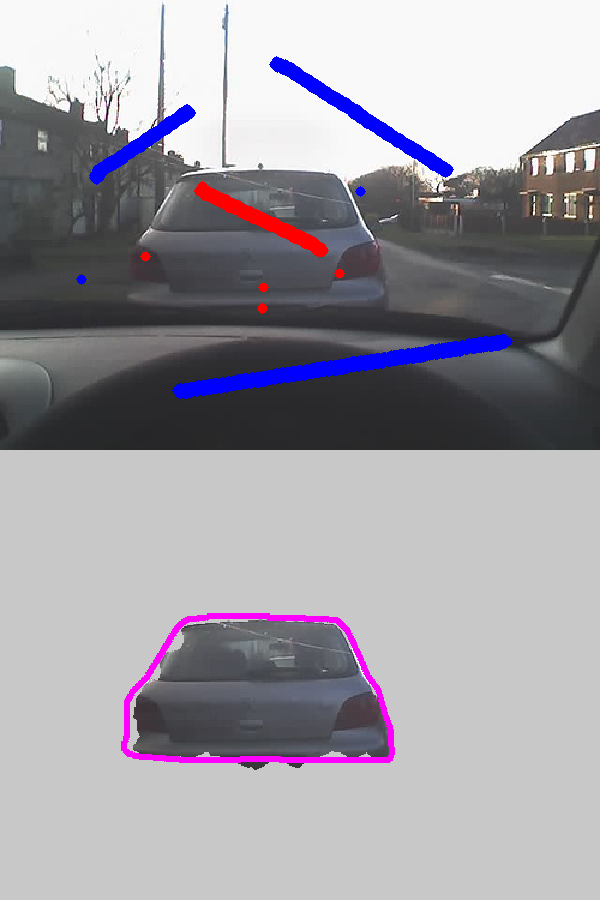}\\
  \begin{sideways}\parbox{15mm}{\centering\footnotesize QP~\cite{grady2006randomwalk,sinop2007seeded}}\end{sideways} &
  \includegraphics[height=3.5cm]{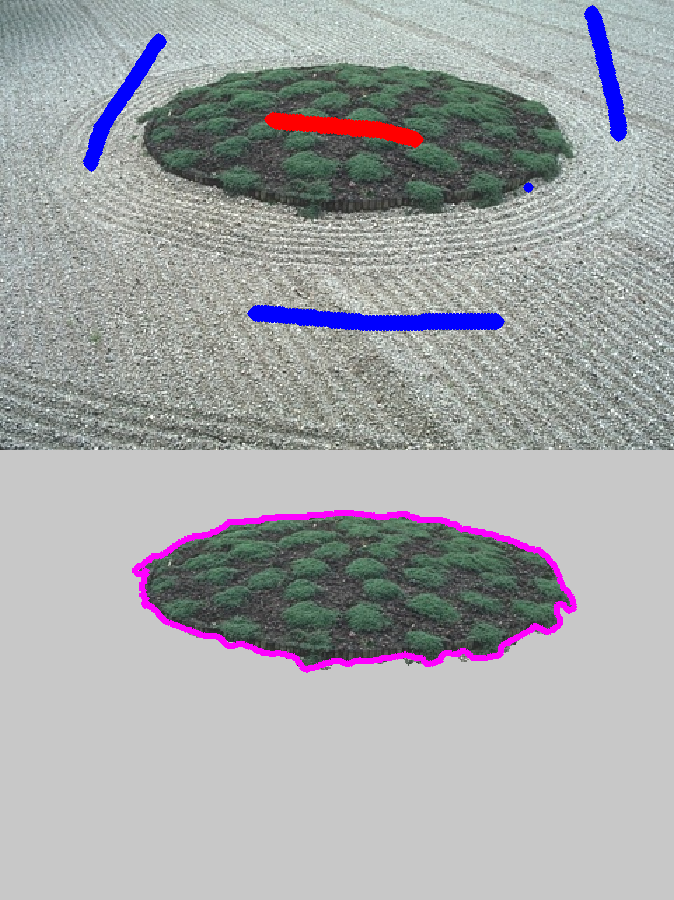}&
  \includegraphics[height=3.5cm]{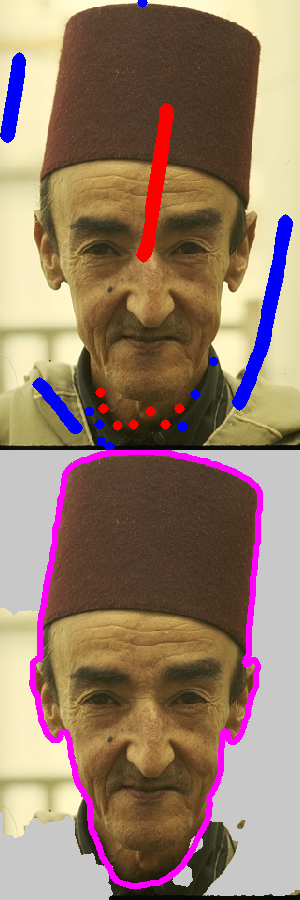}&
  \includegraphics[height=3.5cm]{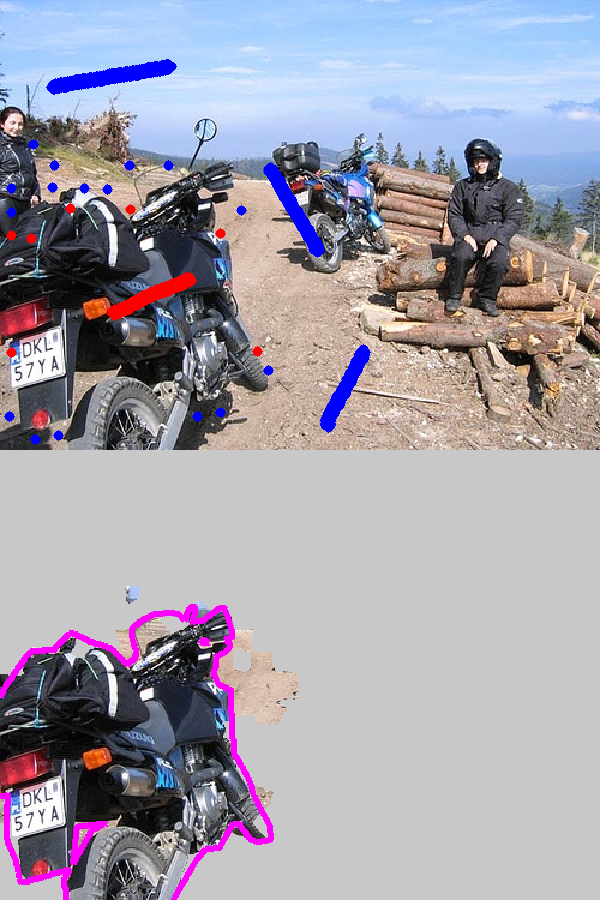}&
  \includegraphics[height=3.5cm]{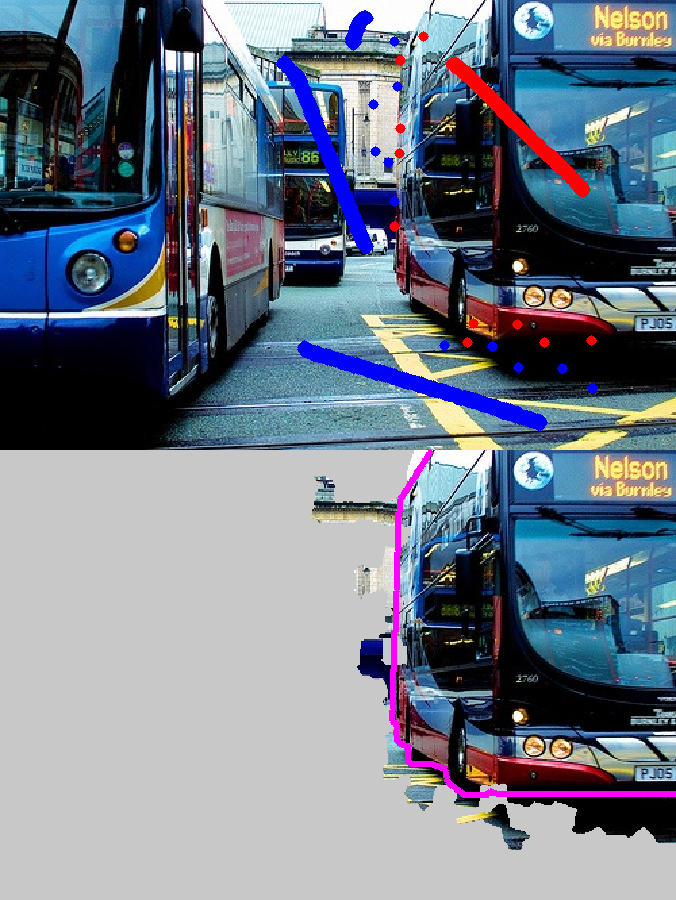}&
  \includegraphics[height=3.5cm]{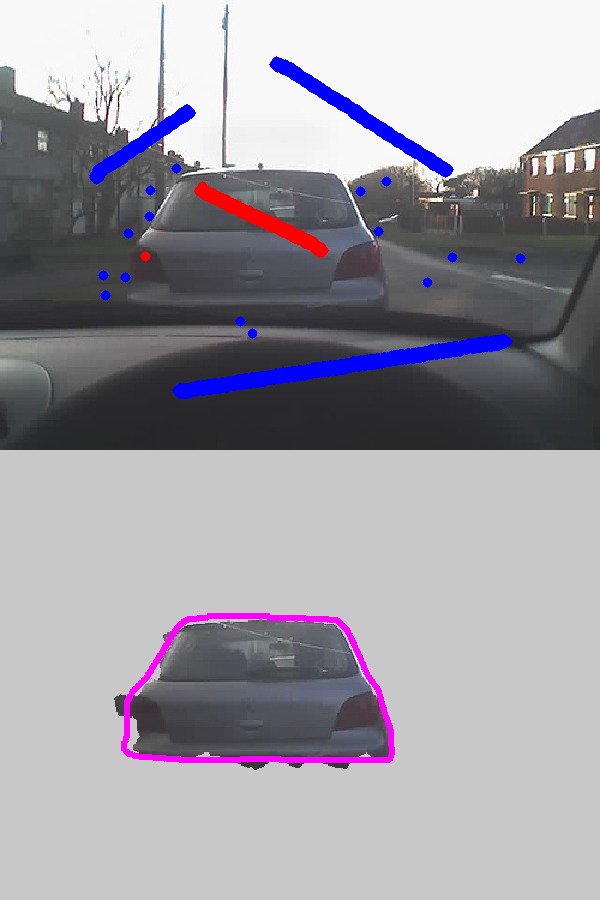}\\
  \begin{sideways}\parbox{15mm}{\centering\footnotesize Our method}\end{sideways} &
  \includegraphics[height=3.5cm]{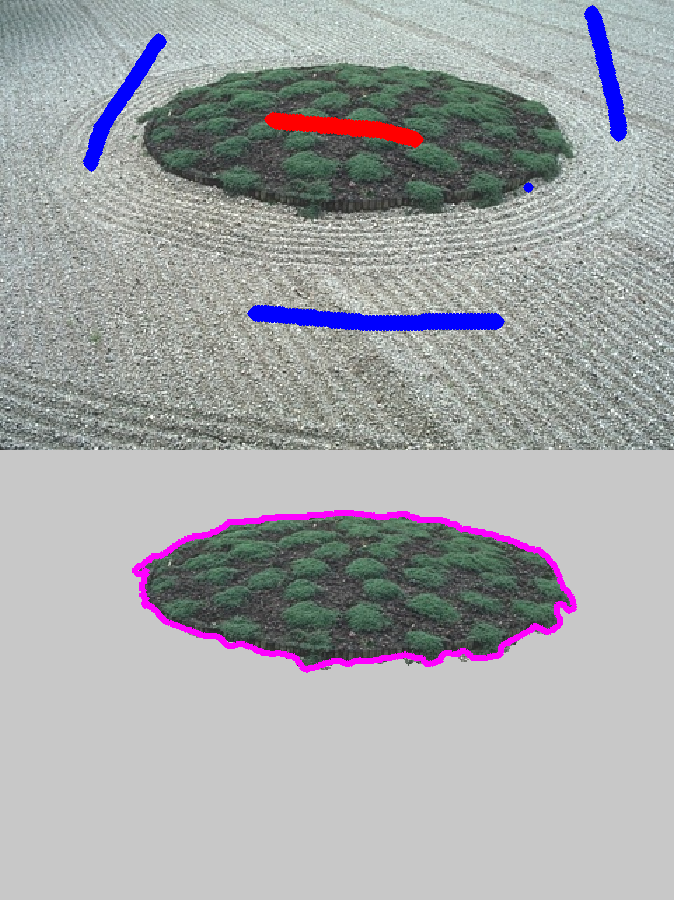}&
  \includegraphics[height=3.5cm]{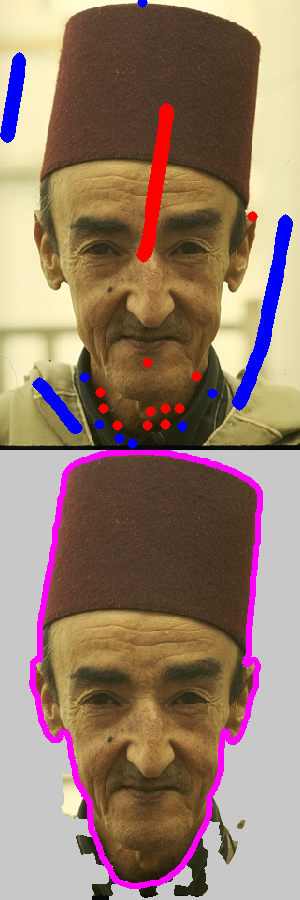}&
  \includegraphics[height=3.5cm]{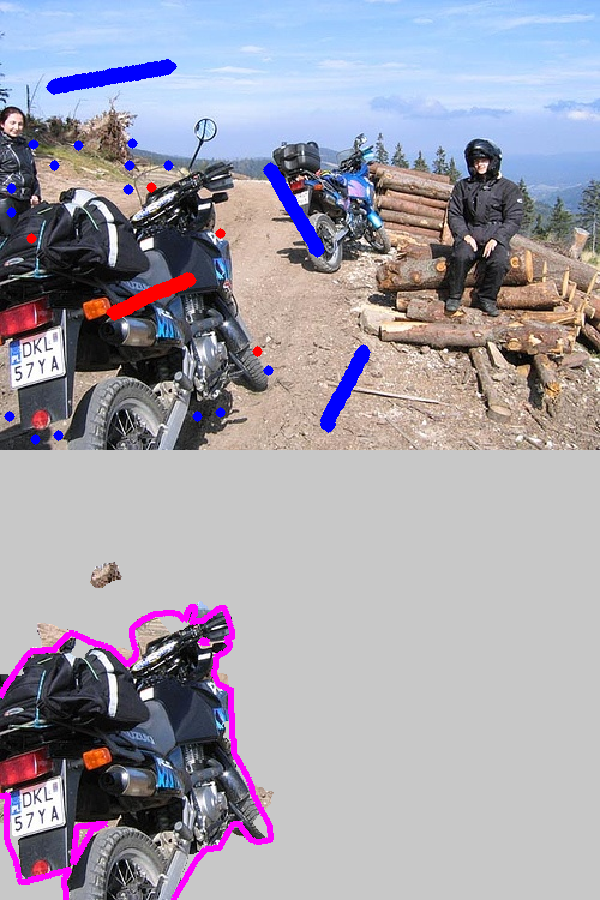}&
  \includegraphics[height=3.5cm]{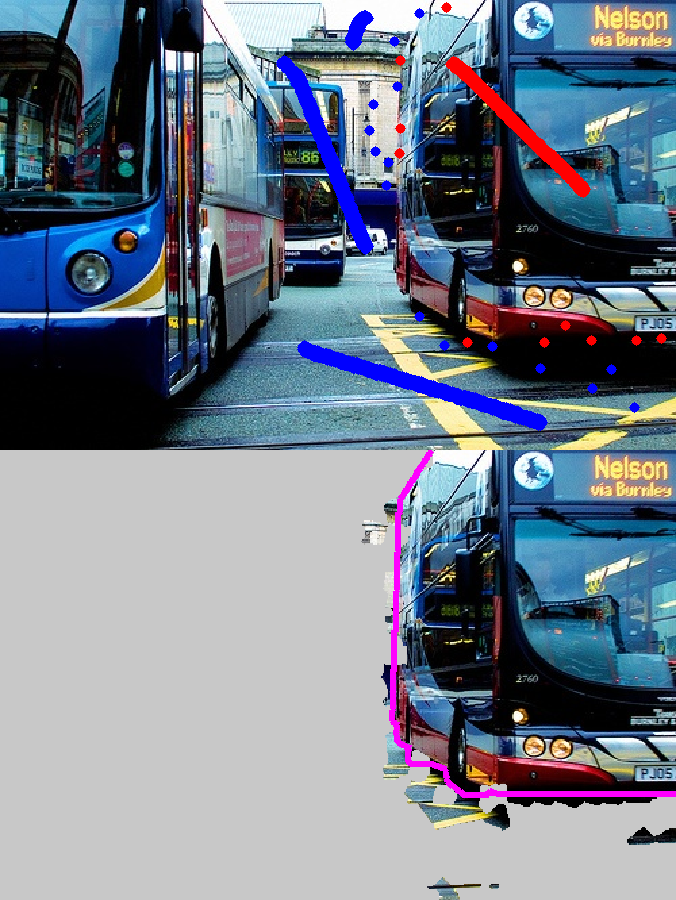}&
  \includegraphics[height=3.5cm]{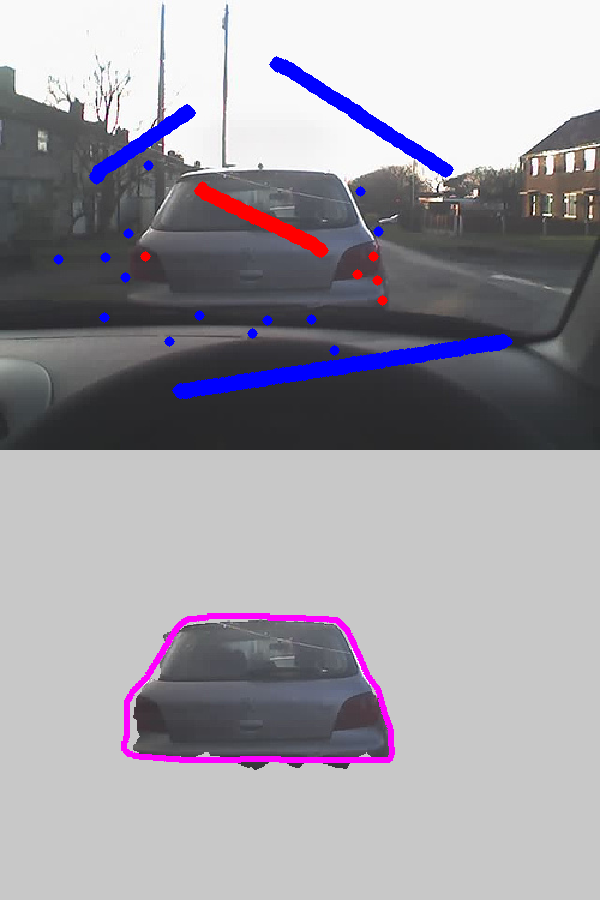}\\
    \end{tabular}
  \end{center}
  \caption{Additional experimental results of segmentation on Oxford dataset-I. The input and result images are shown as top-bottom pairs. The top rows show the input images overlaid with input seeds. The bottom rows show extracted image regions against the ground truth shape contours in purple.}\label{FIG:OxfordSeg1}
\end{figure*}

\begin{figure*}
    \begin{center}
\begin{tabular}
{
@{\hspace{0mm}}c@{\hspace{0.5mm}}c@{\hspace{0.5mm}}c@{\hspace{0.5mm}}c@{\hspace{0.5mm}}c@{\hspace{0.5mm}}c @{\hspace{0.5mm}}c
@{\hspace{1mm}}c@{\hspace{1mm}}c@{\hspace{1mm}}c@{\hspace{1mm}}c@{\hspace{1mm}}c @{\hspace{1mm}}c
} \begin{sideways}\parbox{15mm}{\centering\footnotesize GC~\cite{Boykov01GraphCut}}\end{sideways} &
  \includegraphics[height=4cm]{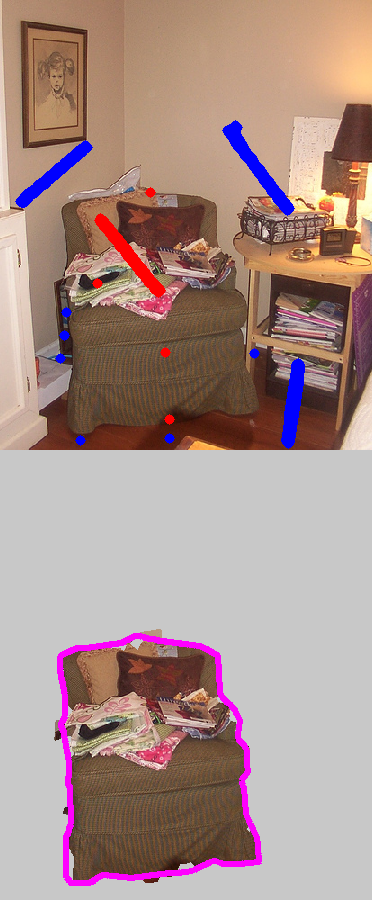}&
  \includegraphics[height=4cm]{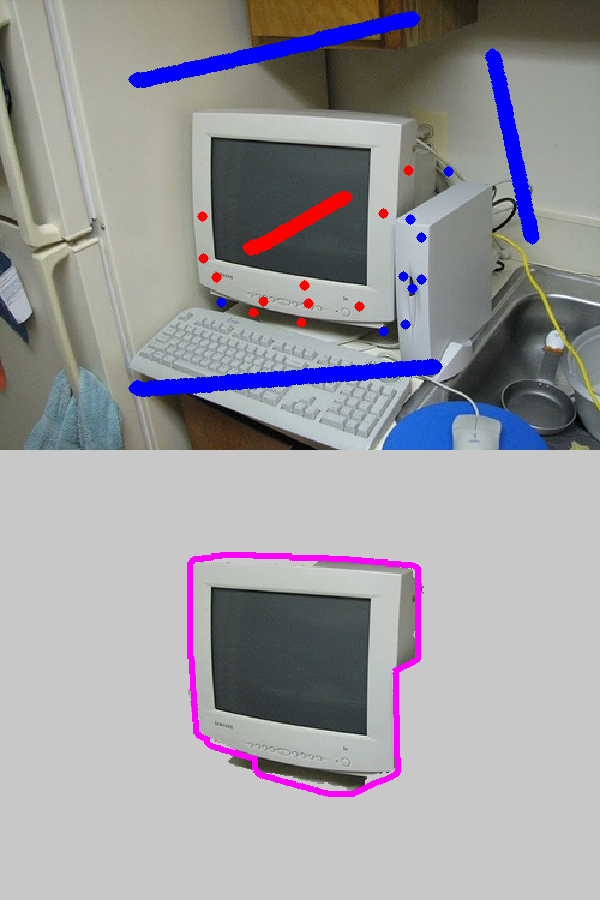}&
  \includegraphics[height=4cm]{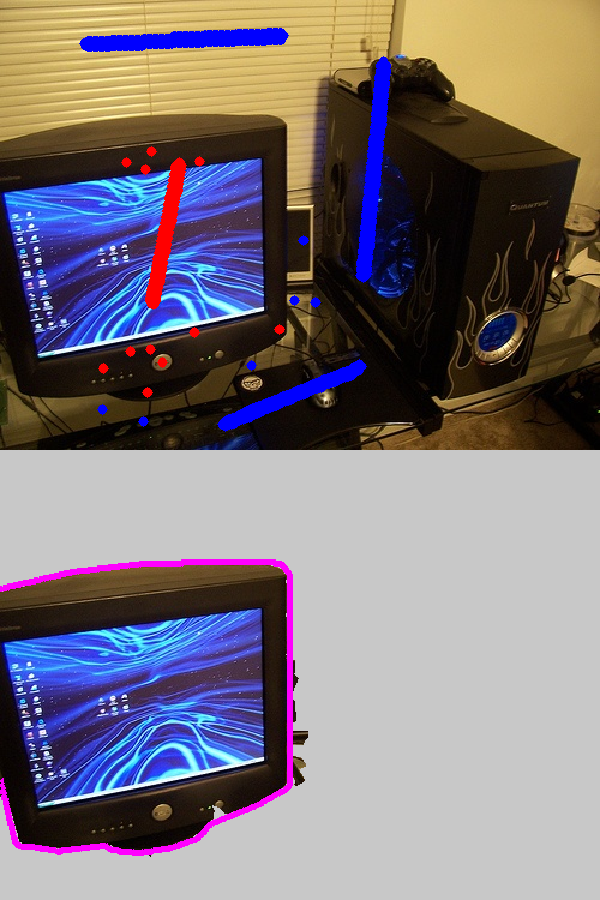}&
  \includegraphics[height=4cm]{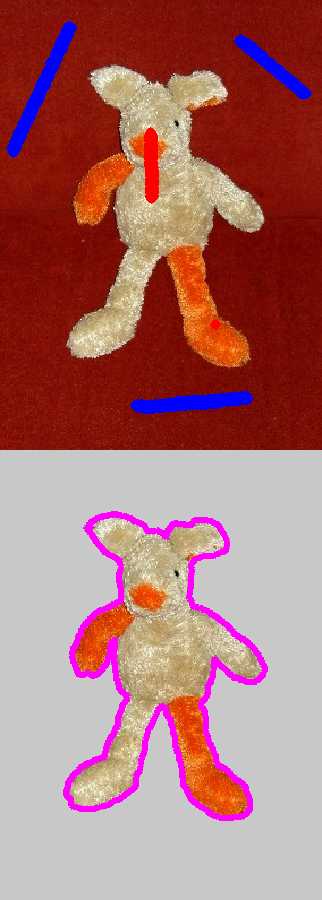}&  \includegraphics[height=4cm]{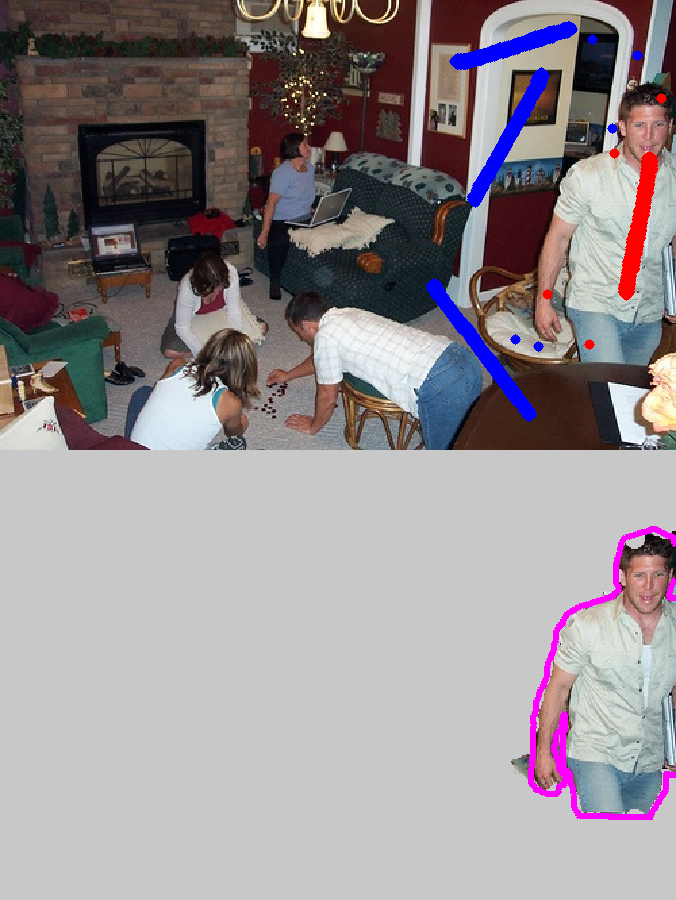}\\
  \begin{sideways}\parbox{15mm}{\centering\footnotesize LP~\cite{LiHongDong10LPSeg}}\end{sideways} &
  \includegraphics[height=4cm]{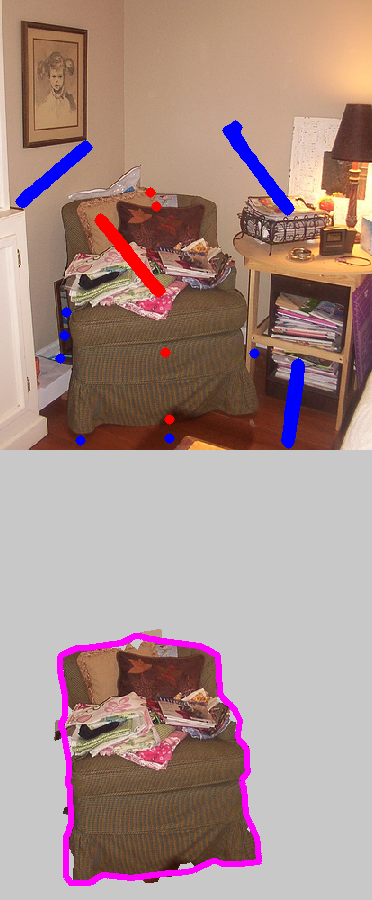}&
  \includegraphics[height=4cm]{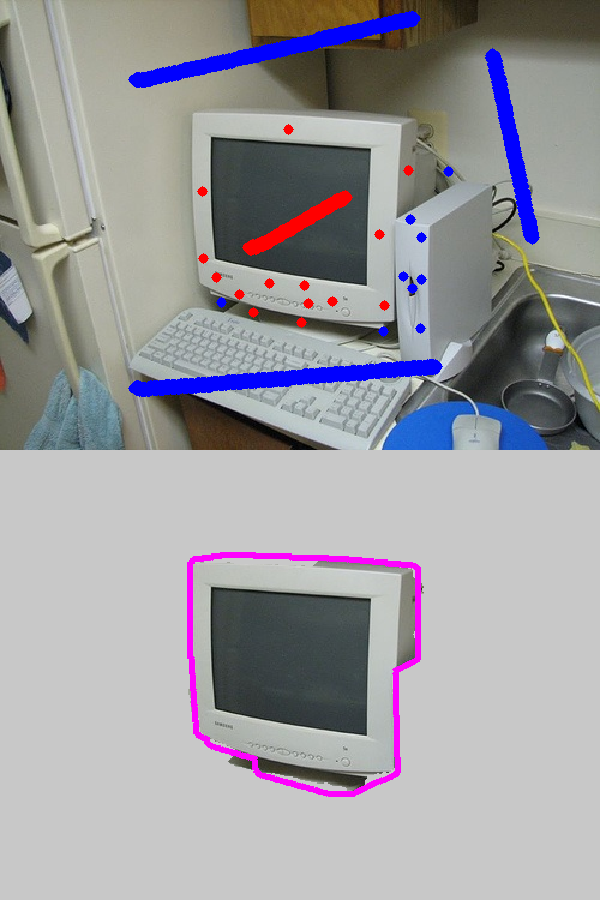}&
  \includegraphics[height=4cm]{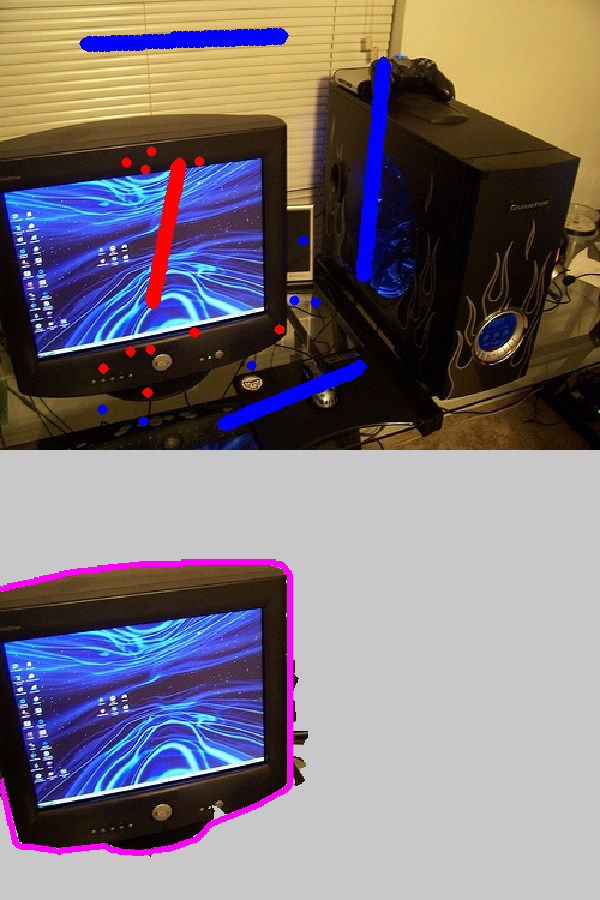}&
  \includegraphics[height=4cm]{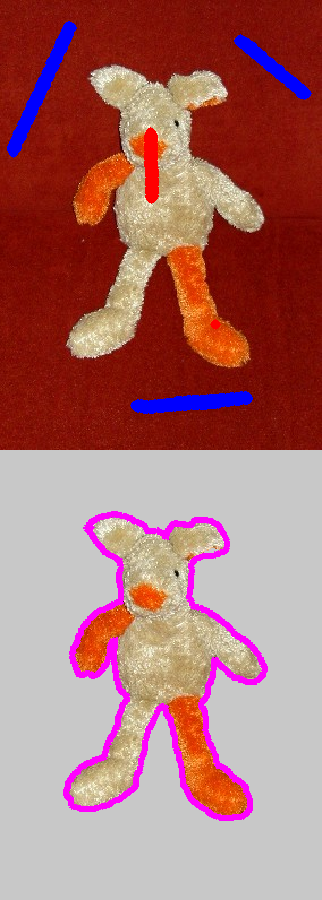}&
  \includegraphics[height=4cm]{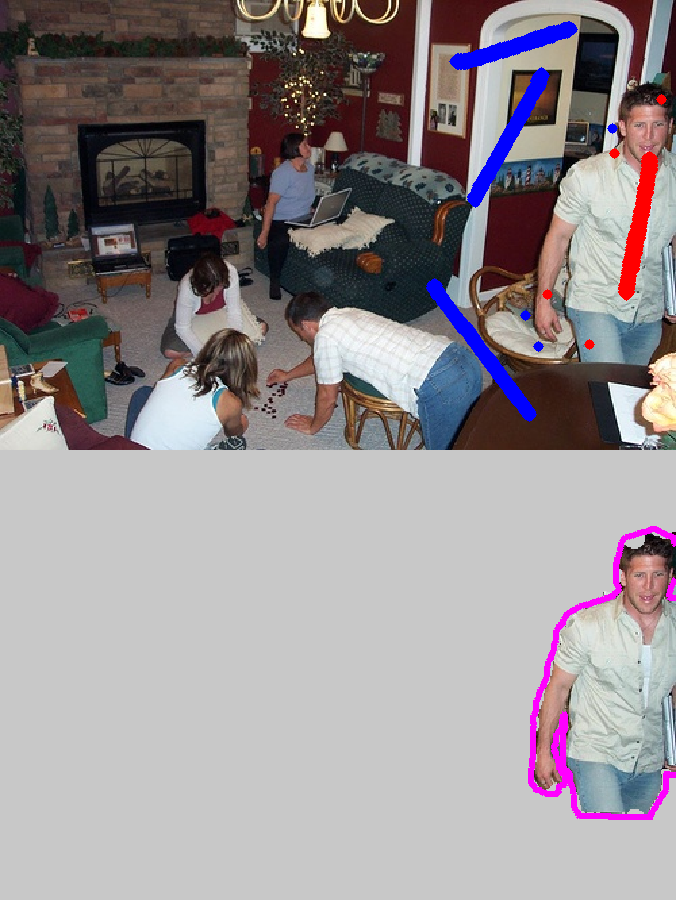}\\
  \begin{sideways}\parbox{15mm}{\centering\footnotesize QP~\cite{grady2006randomwalk,sinop2007seeded}}\end{sideways} &
  \includegraphics[height=4cm]{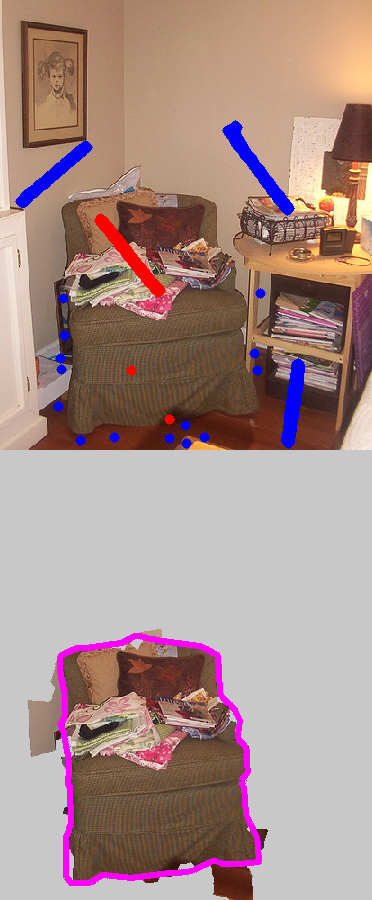}&
  \includegraphics[height=4cm]{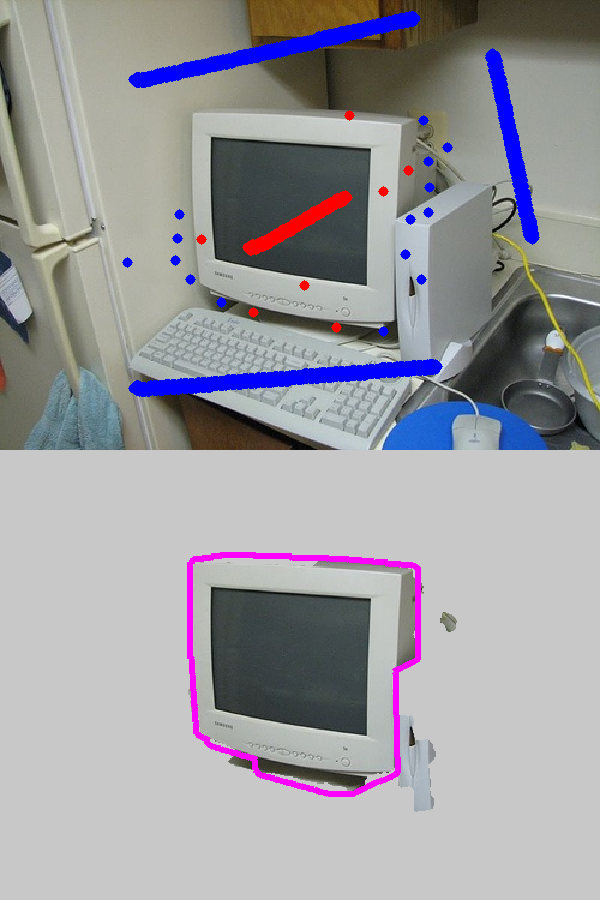}&
  \includegraphics[height=4cm]{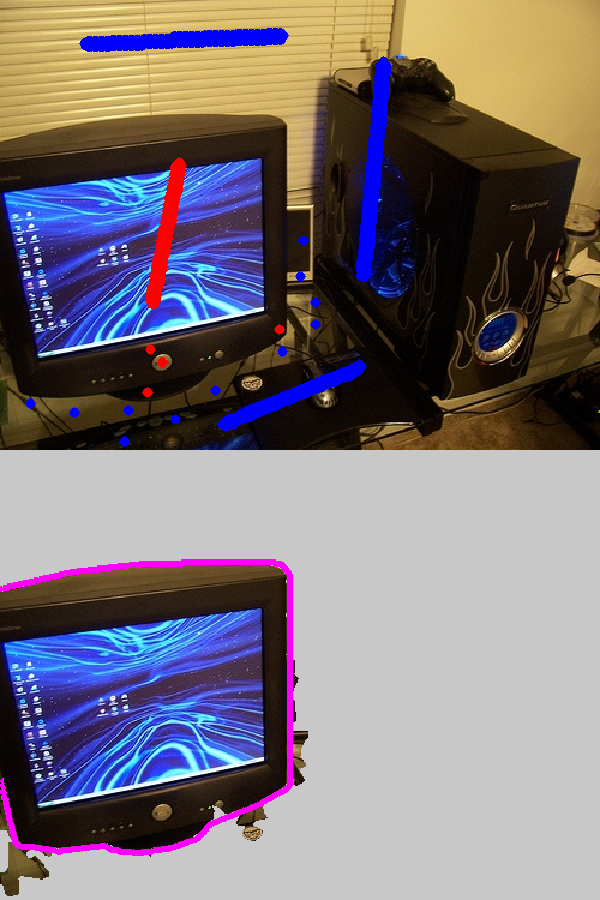}&
  \includegraphics[height=4cm]{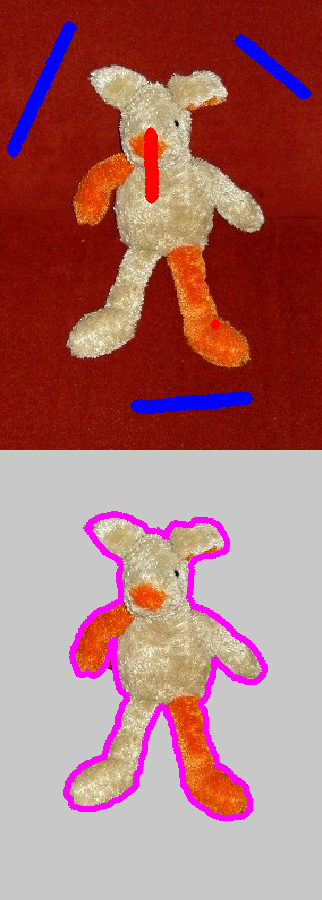}&
  \includegraphics[height=4cm]{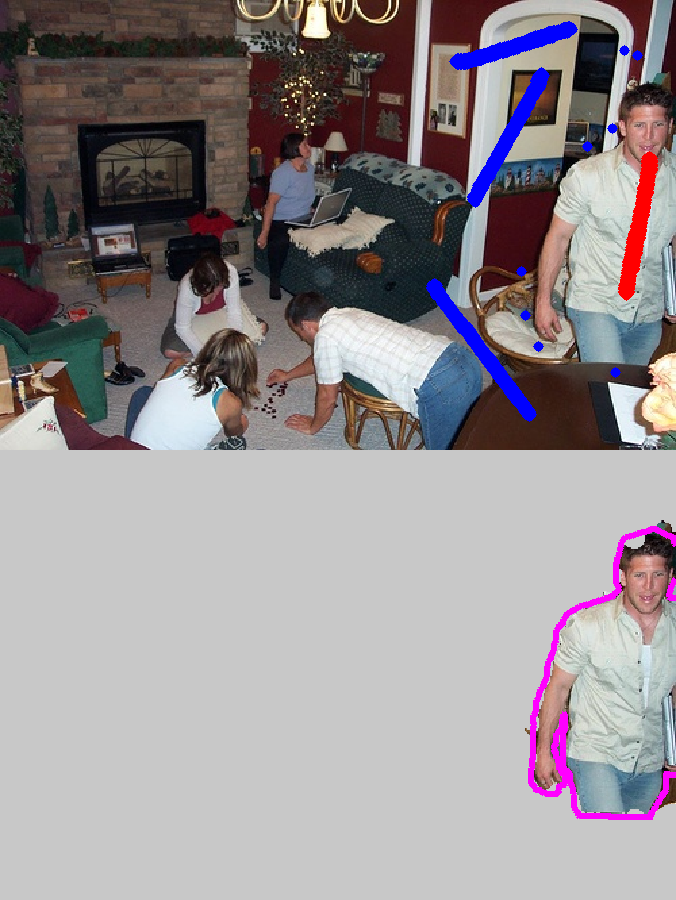}\\
  \begin{sideways}\parbox{15mm}{\centering\footnotesize Our method}\end{sideways} &
  \includegraphics[height=4cm]{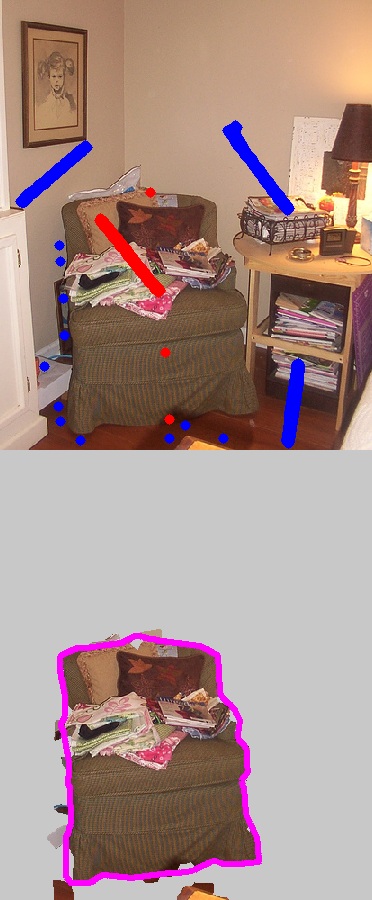}&
  \includegraphics[height=4cm]{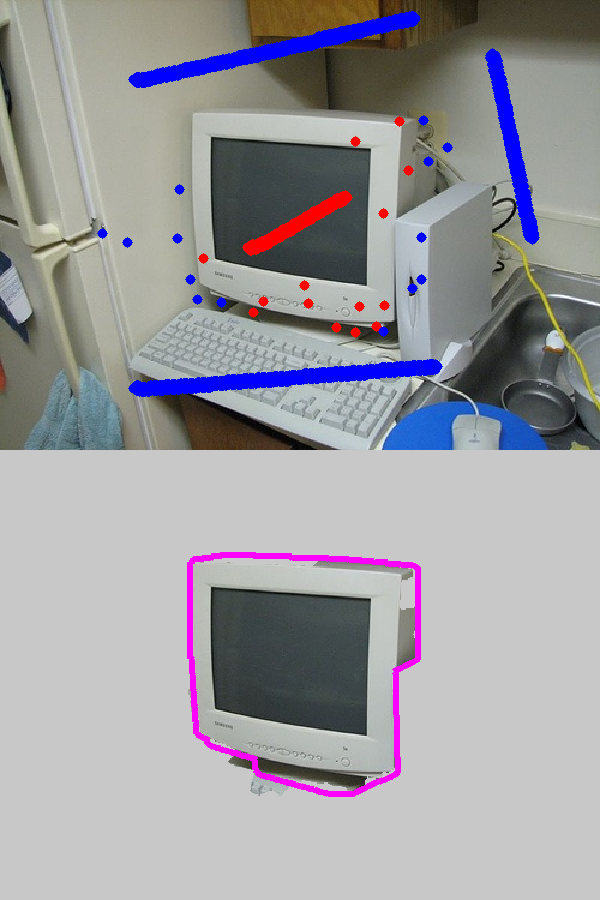}&
  \includegraphics[height=4cm]{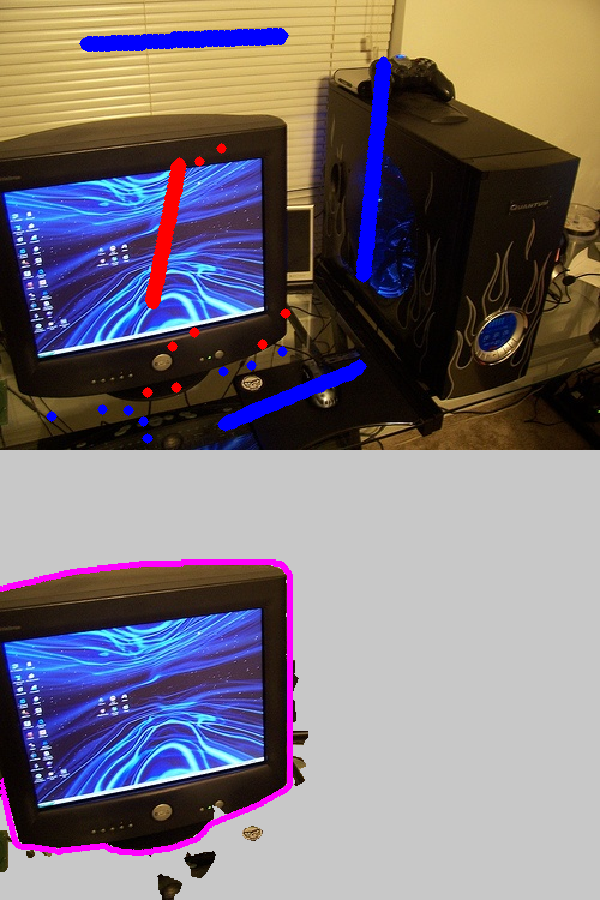}&
  \includegraphics[height=4cm]{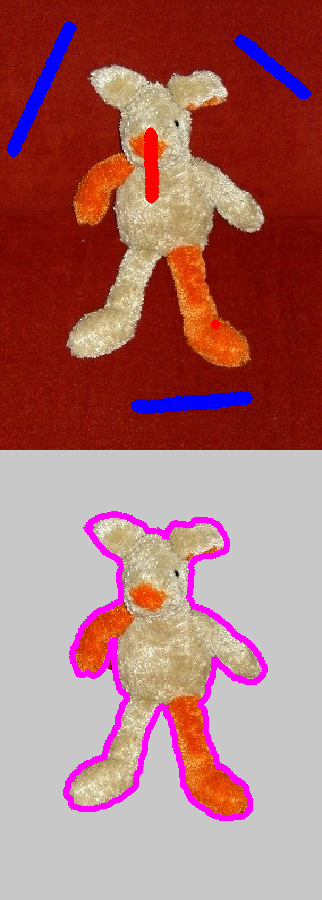}&
  \includegraphics[height=4cm]{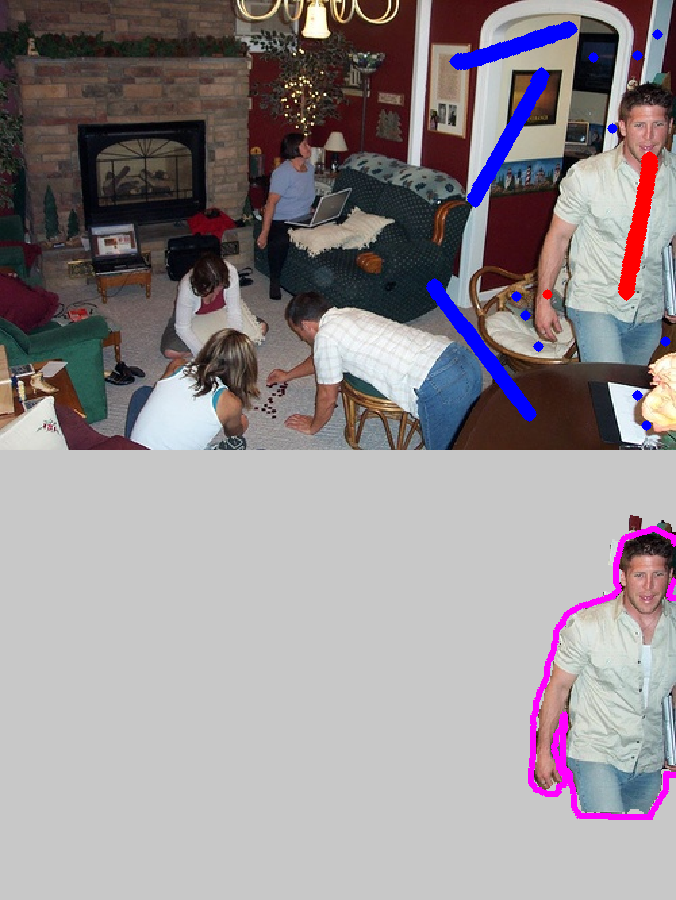}\\
    \end{tabular}
  \end{center}
  \caption{Additional experimental results of segmentation on Oxford dataset-II}\label{FIG:OxfordSeg2}
\end{figure*}

\begin{figure*}
    \begin{center}
\begin{tabular}
{
@{\hspace{0mm}}c@{\hspace{0.5mm}}c@{\hspace{0.5mm}}c@{\hspace{0.5mm}}c@{\hspace{0.5mm}}c@{\hspace{0.5mm}}c @{\hspace{0.5mm}}c
@{\hspace{1mm}}c@{\hspace{1mm}}c@{\hspace{1mm}}c@{\hspace{1mm}}c@{\hspace{1mm}}c @{\hspace{1mm}}c
} \begin{sideways}\parbox{15mm}{\centering\footnotesize GC~\cite{Boykov01GraphCut}}\end{sideways} &
  \includegraphics[height=3.5cm]{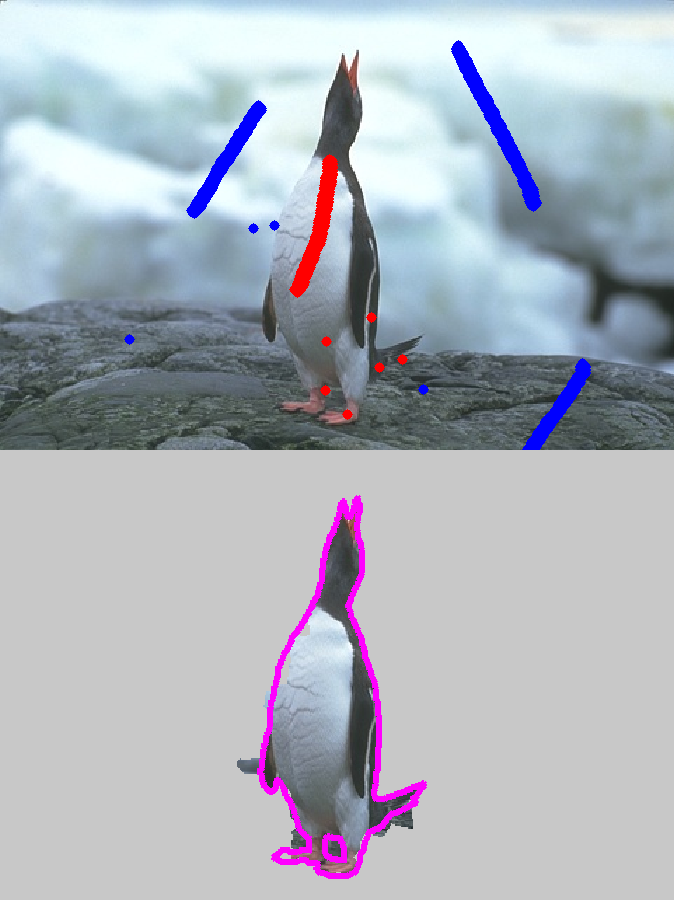}&
  \includegraphics[height=3.5cm]{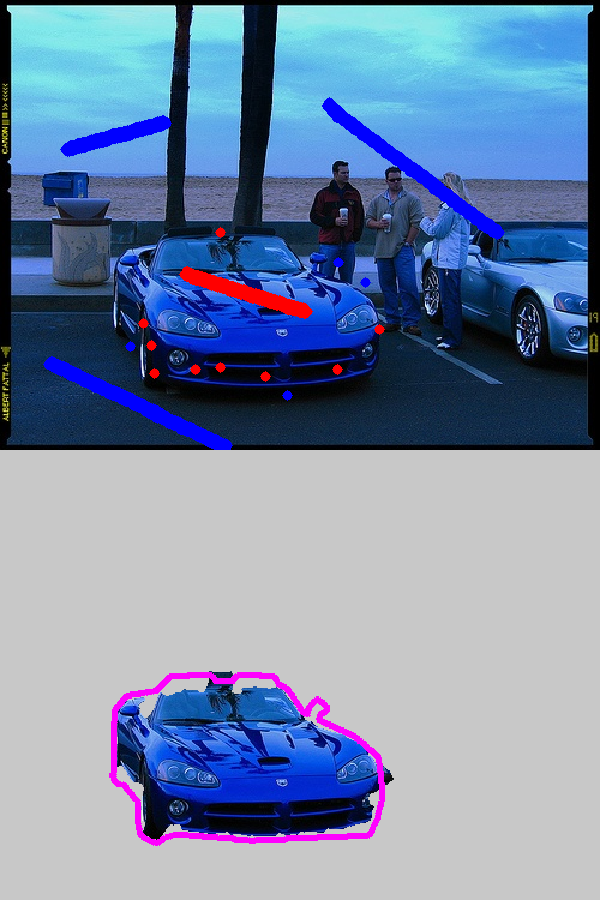}&
  \includegraphics[height=3.5cm]{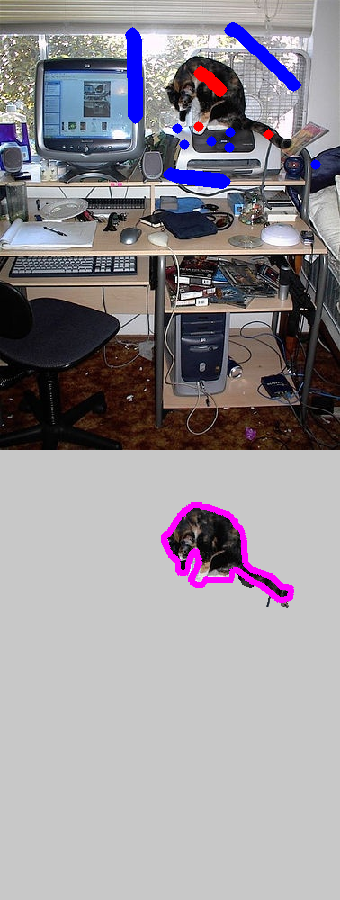}&
  \includegraphics[height=3.5cm]{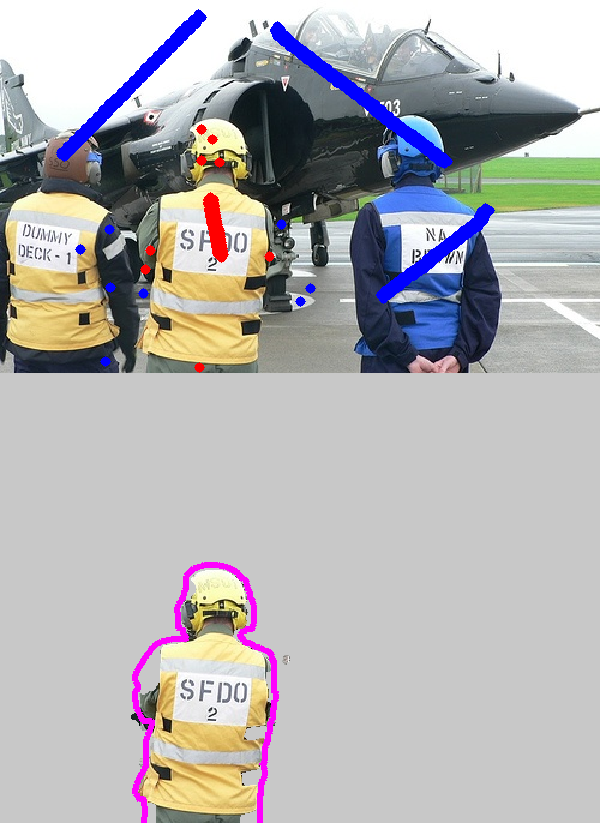}&  \includegraphics[height=3.5cm]{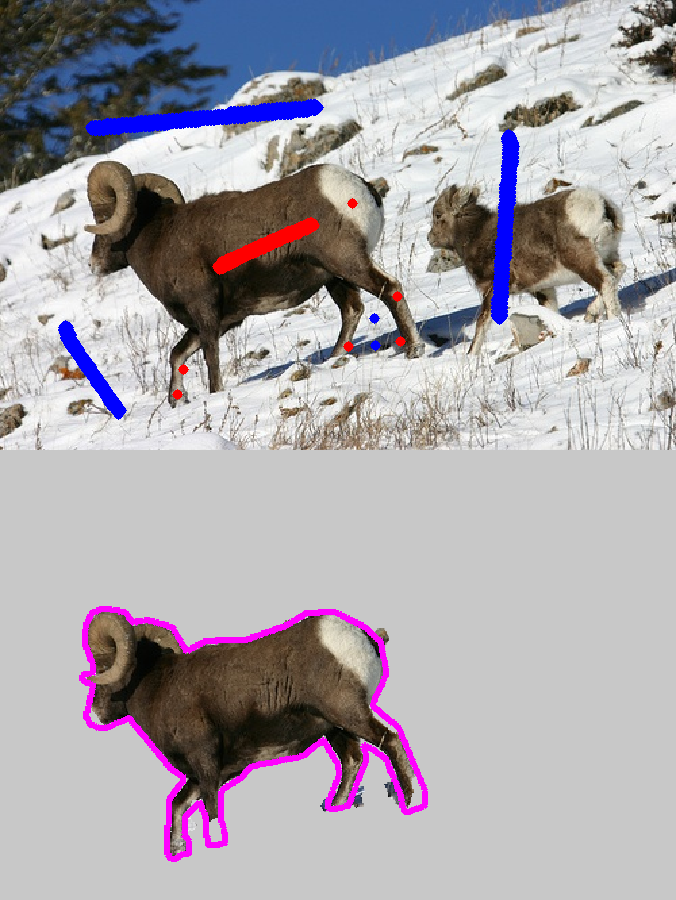}\\
  \begin{sideways}\parbox{15mm}{\centering\footnotesize LP~\cite{LiHongDong10LPSeg}}\end{sideways} &
  \includegraphics[height=3.5cm]{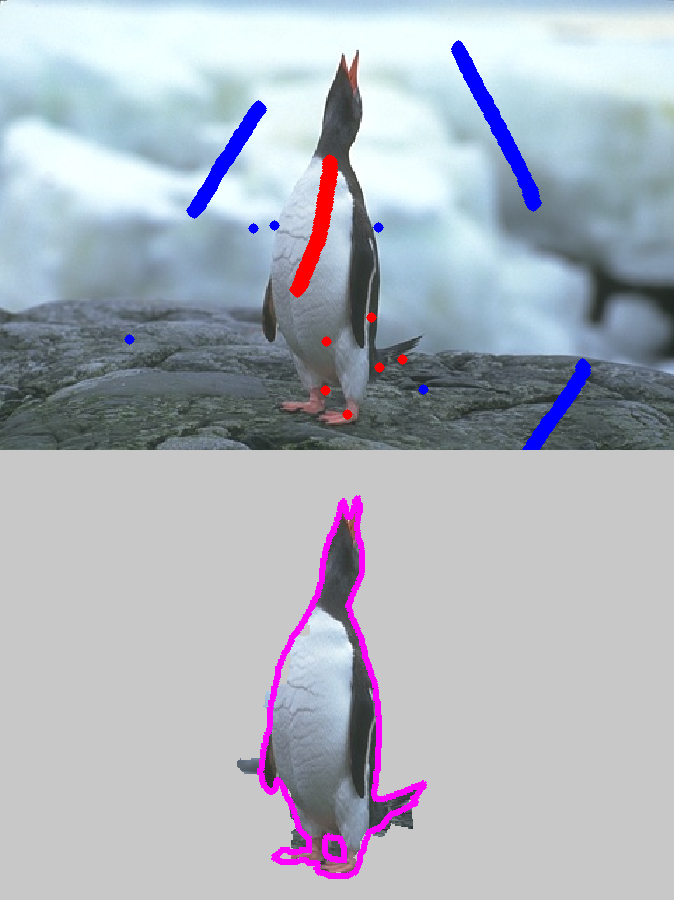}&
  \includegraphics[height=3.5cm]{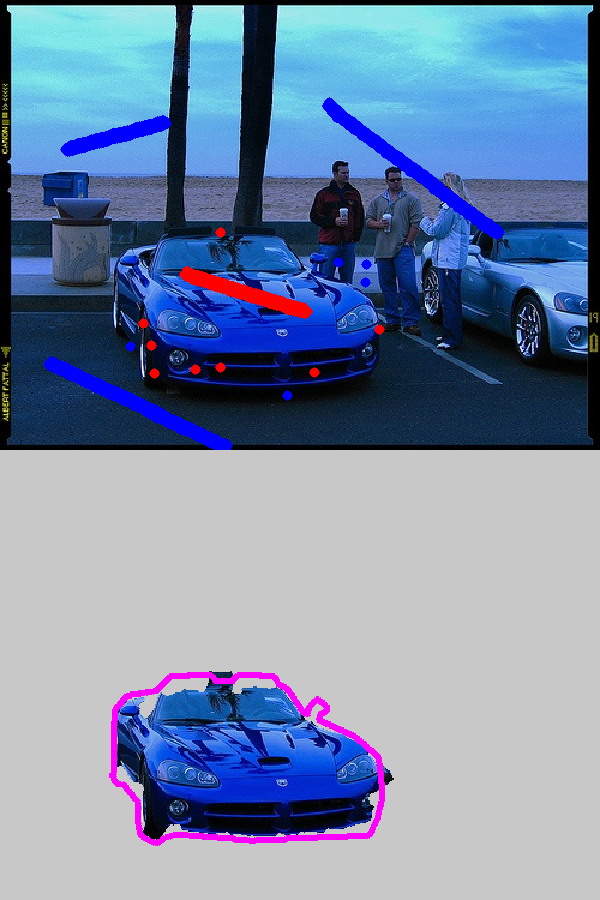}&
  \includegraphics[height=3.5cm]{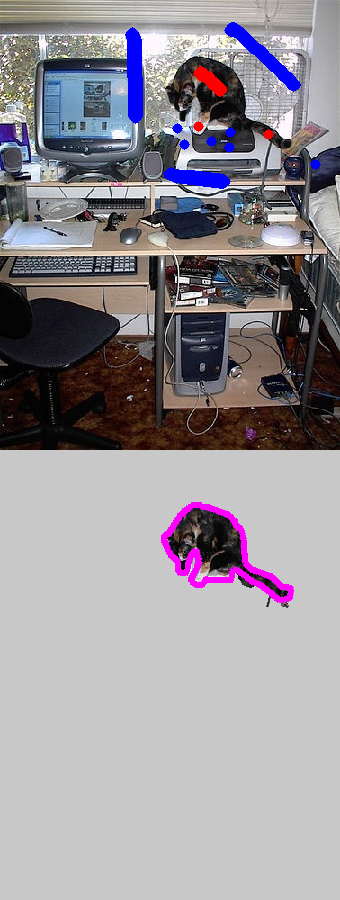}&
  \includegraphics[height=3.5cm]{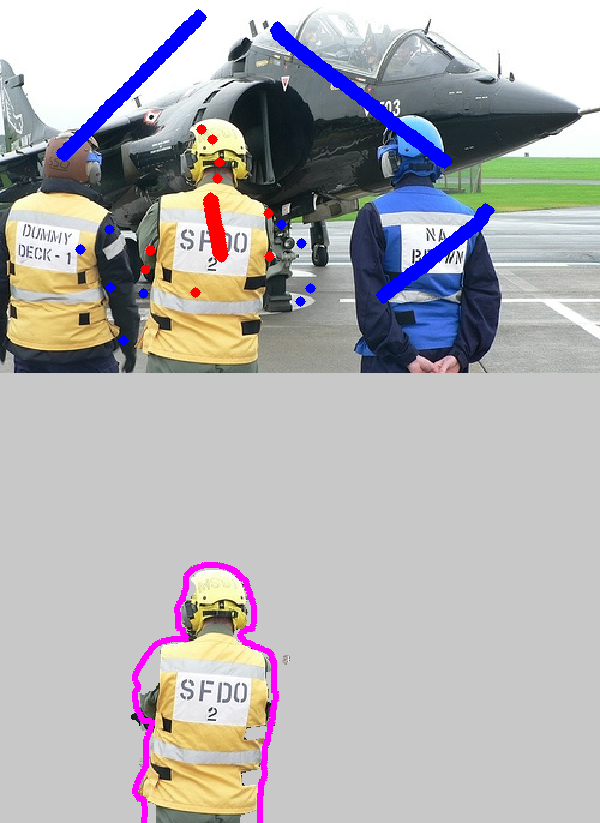}&
  \includegraphics[height=3.5cm]{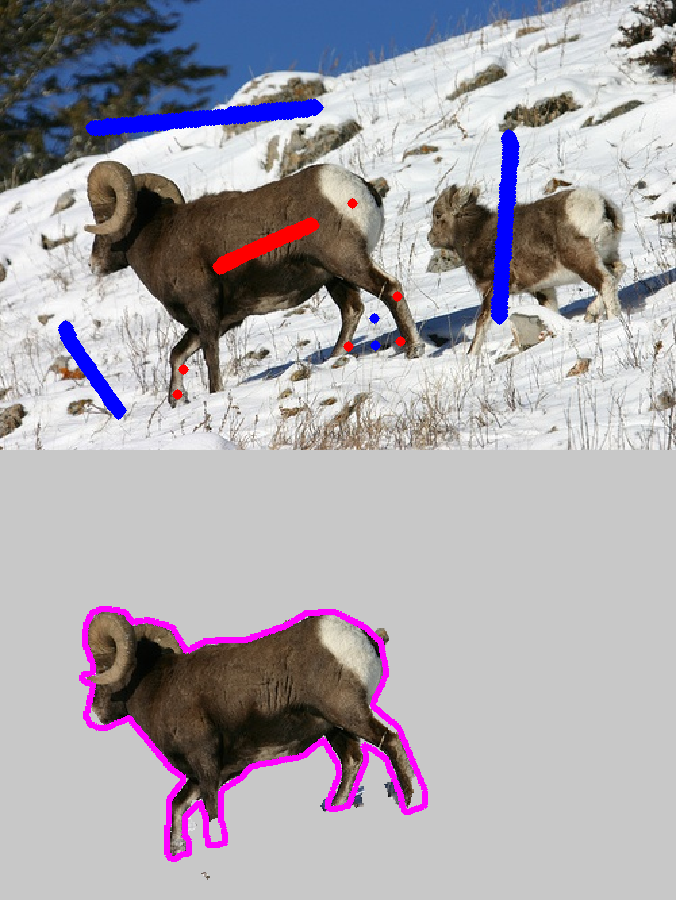}\\
  \begin{sideways}\parbox{15mm}{\centering\footnotesize QP~\cite{grady2006randomwalk,sinop2007seeded}}\end{sideways} &
  \includegraphics[height=3.5cm]{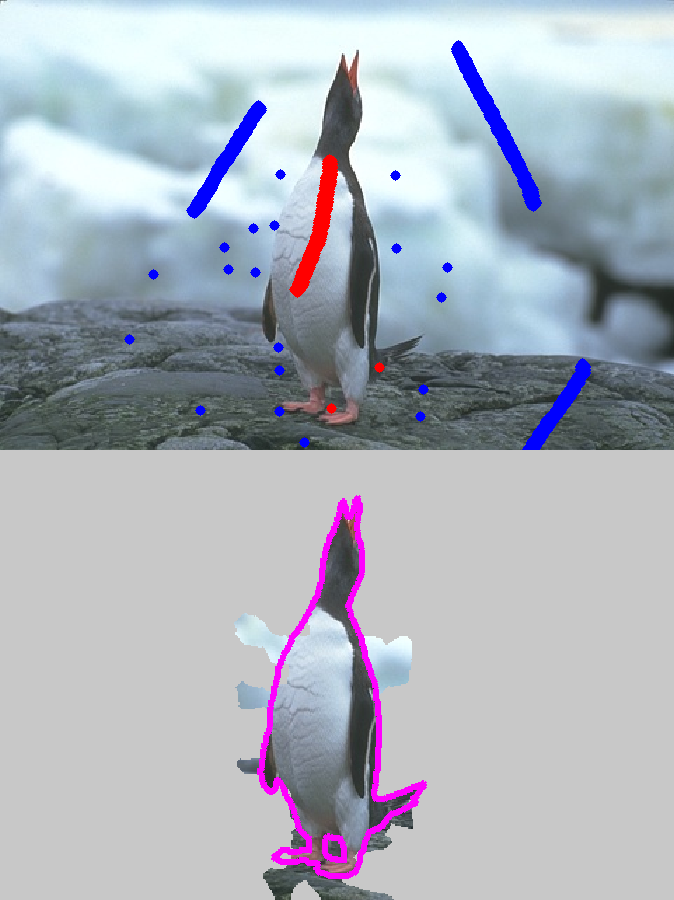}&
  \includegraphics[height=3.5cm]{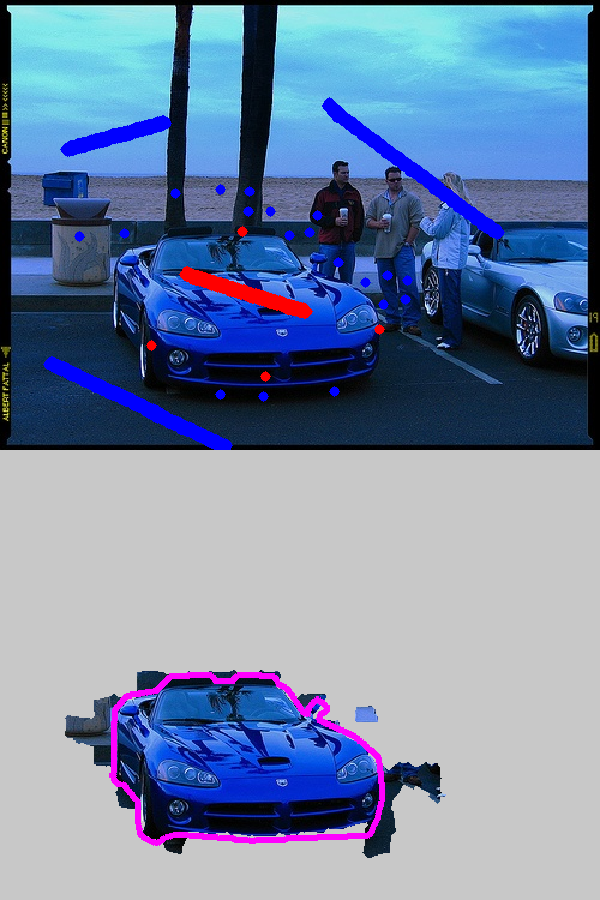}&
  \includegraphics[height=3.5cm]{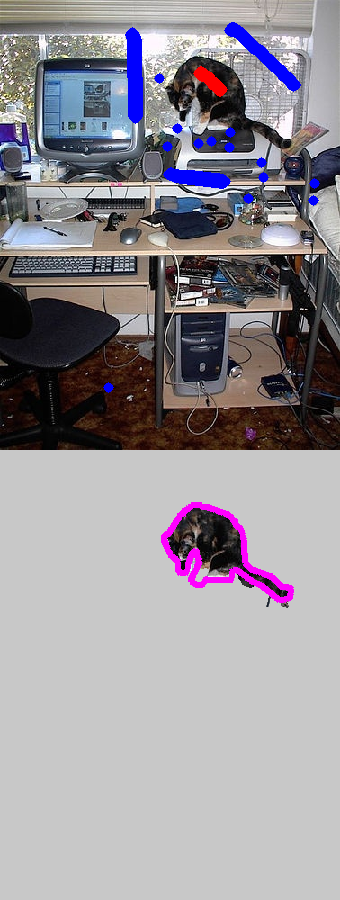}&
  \includegraphics[height=3.5cm]{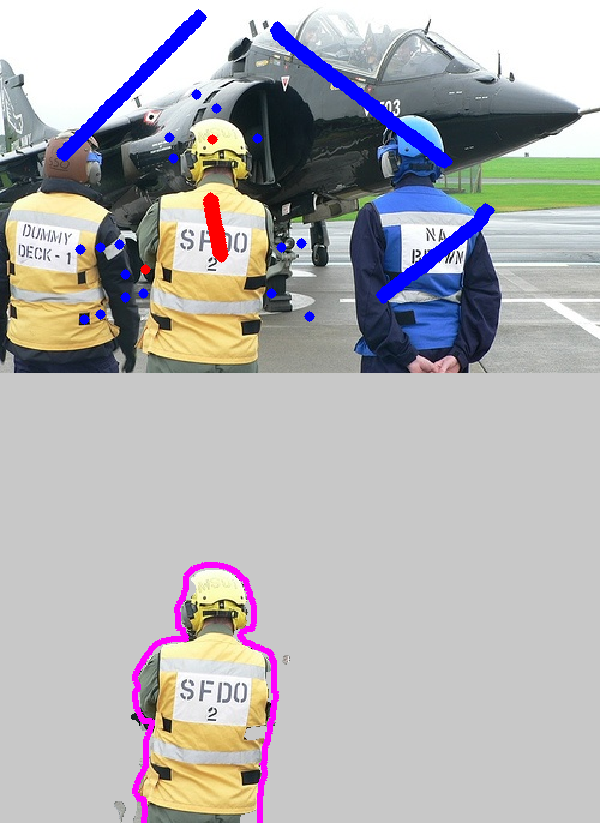}&
  \includegraphics[height=3.5cm]{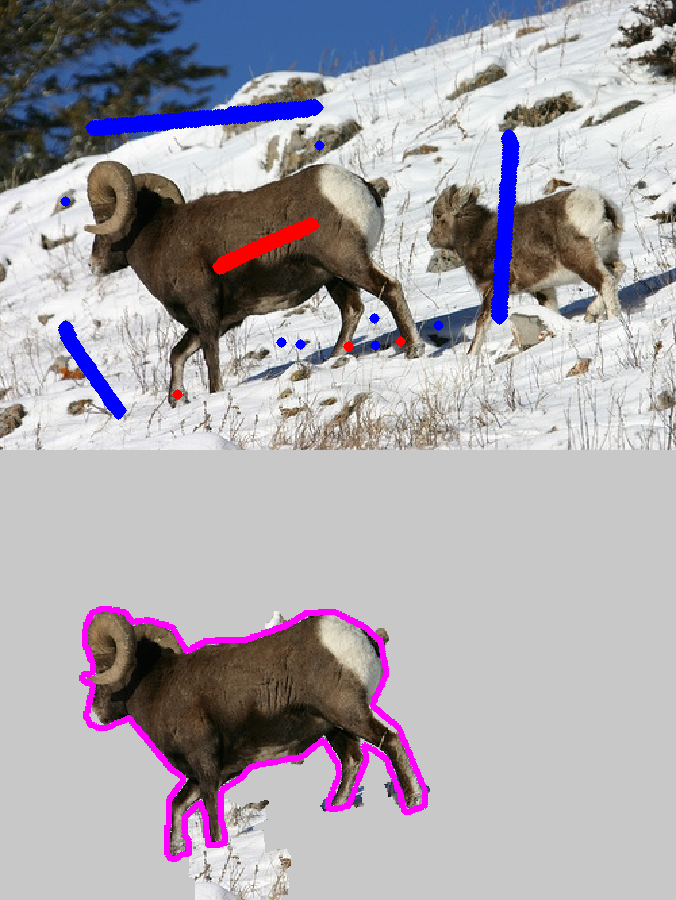}\\
  \begin{sideways}\parbox{15mm}{\centering\footnotesize Our method}\end{sideways} &
  \includegraphics[height=3.5cm]{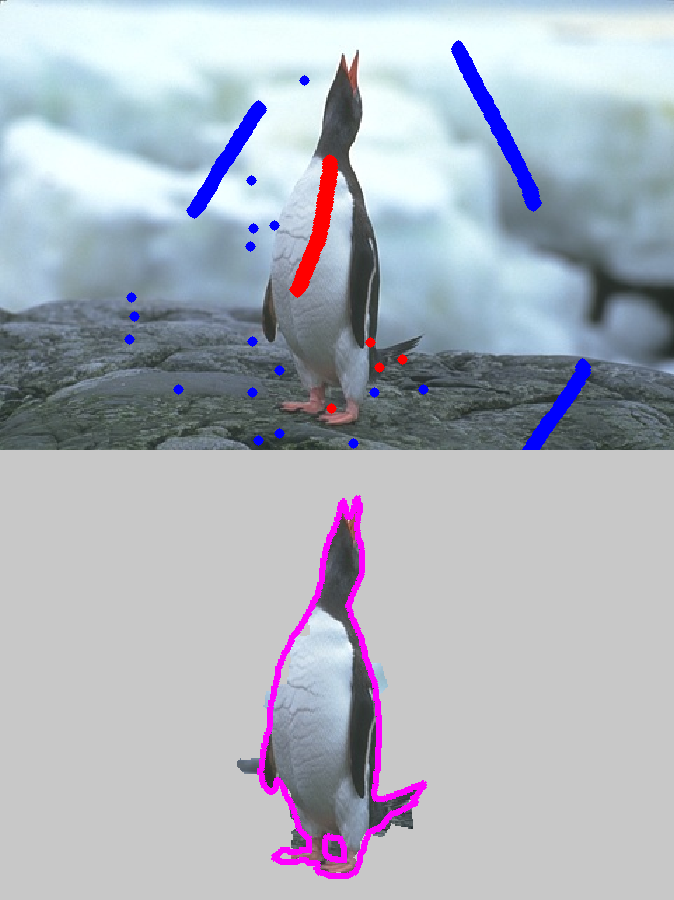}&
  \includegraphics[height=3.5cm]{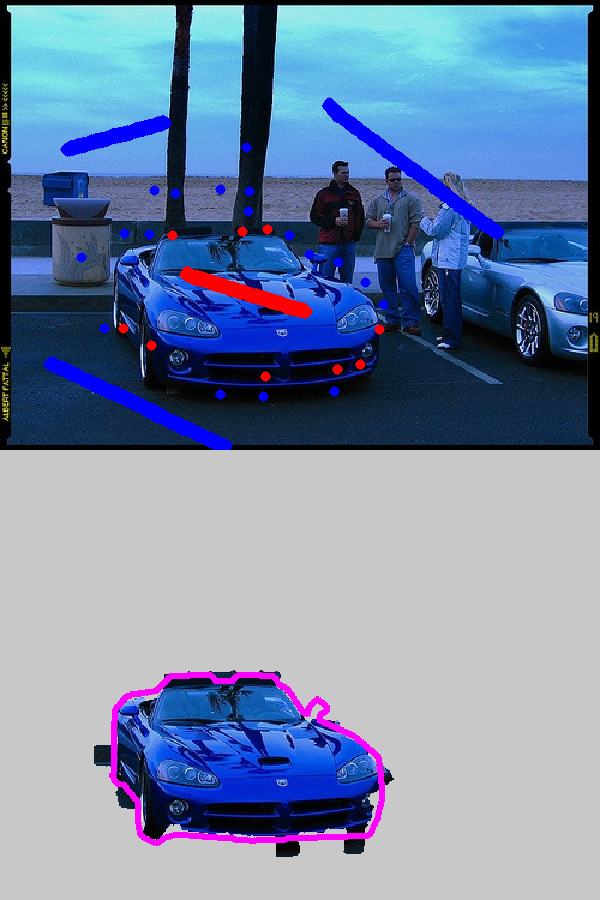}&
  \includegraphics[height=3.5cm]{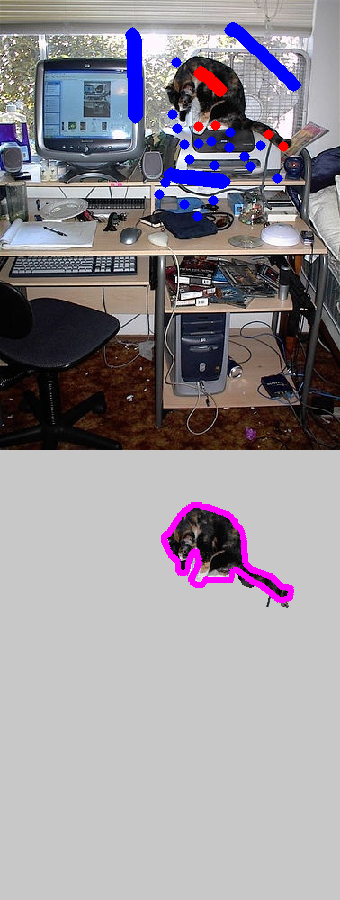}&
  \includegraphics[height=3.5cm]{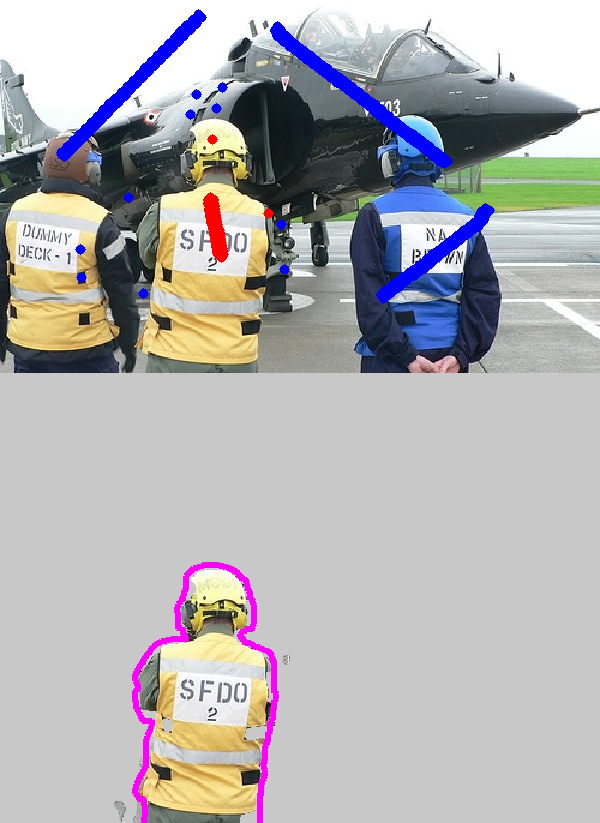}&
  \includegraphics[height=3.5cm]{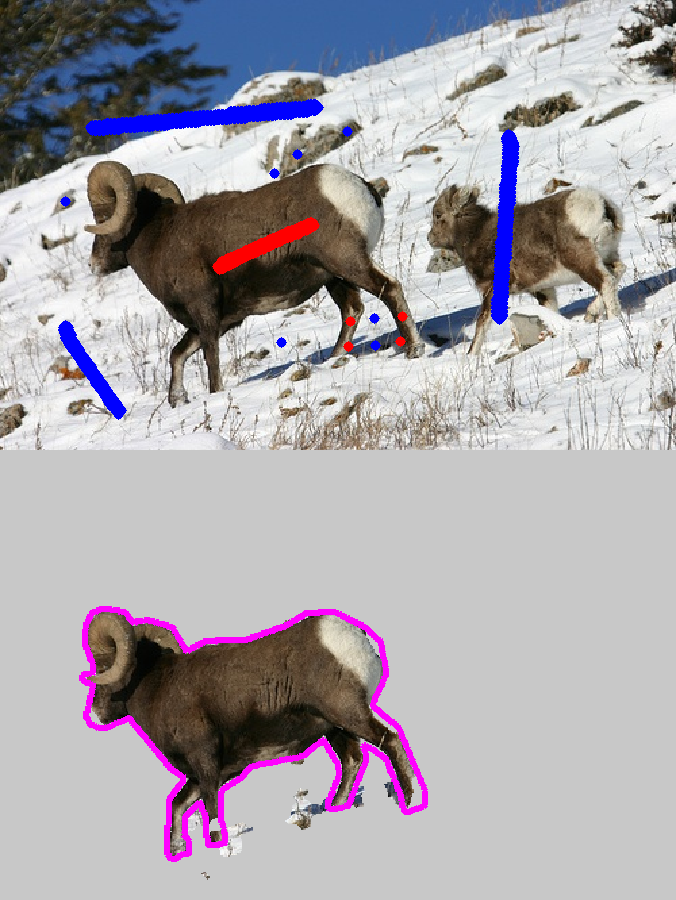}\\
    \end{tabular}
  \end{center}
  \caption{Additional experimental results of segmentation on Oxford dataset-III}\label{FIG:OxfordSeg3}
\end{figure*}

\begin{figure}
    \begin{center}
\begin{tabular}
{
@{\hspace{0mm}}c@{\hspace{0.5mm}}c@{\hspace{0.5mm}}c@{\hspace{0.5mm}}c@{\hspace{0.5mm}}c@{\hspace{0.5mm}}c @{\hspace{0.5mm}}c
@{\hspace{1mm}}c@{\hspace{1mm}}c@{\hspace{1mm}}c@{\hspace{1mm}}c@{\hspace{1mm}}c @{\hspace{1mm}}c
}
  \begin{sideways}\parbox{15mm}{\centering\footnotesize LP~\cite{LiHongDong10LPSeg}}\end{sideways} &
  \includegraphics[height=1.8cm]{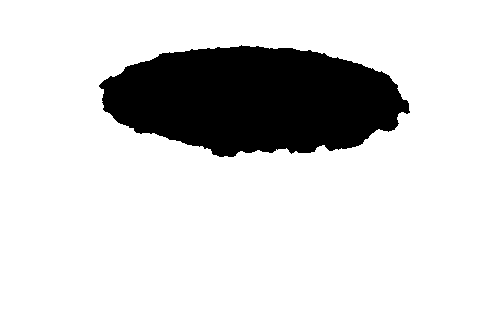}&
  \includegraphics[height=1.8cm]{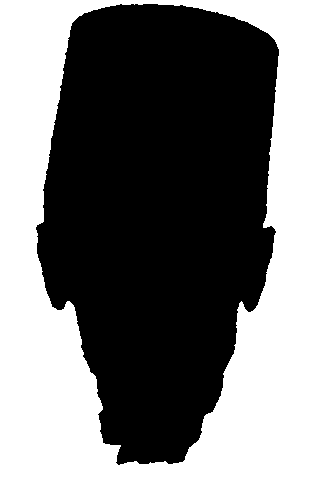}&
  \includegraphics[height=1.8cm]{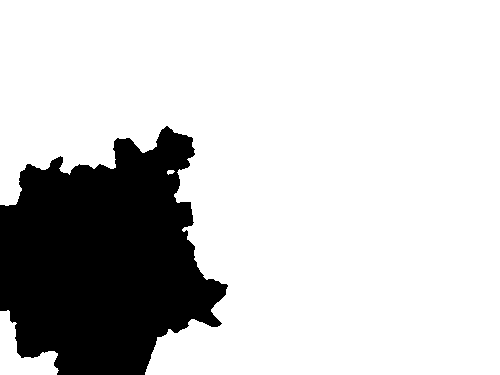}&
  \includegraphics[height=1.8cm]{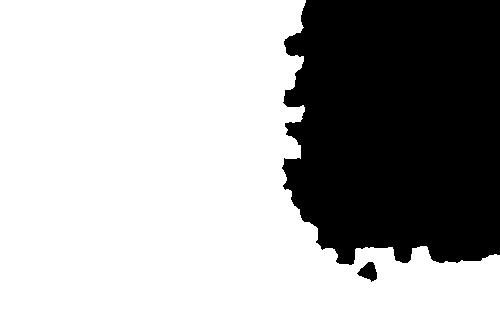}&
  \includegraphics[height=1.8cm]{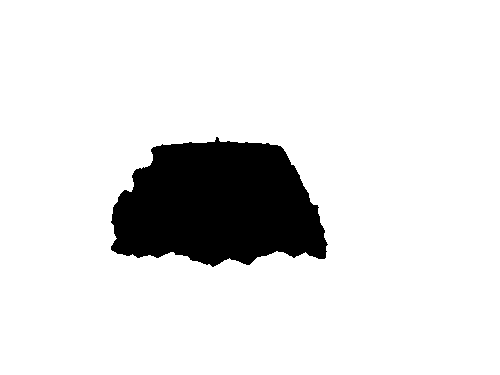}\\
  \begin{sideways}\parbox{15mm}{\centering\footnotesize QP~\cite{grady2006randomwalk,sinop2007seeded}}\end{sideways} &
  \includegraphics[height=1.8cm]{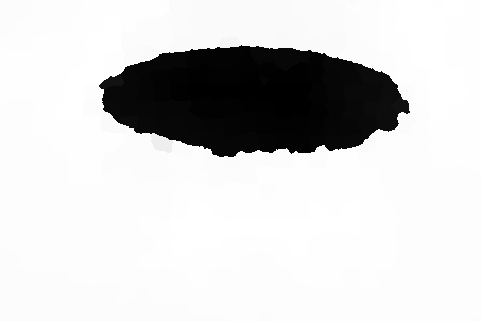}&
  \includegraphics[height=1.8cm]{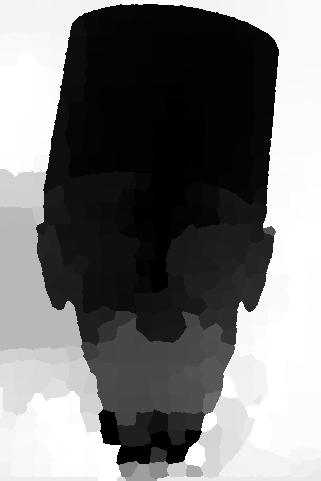}&
  \includegraphics[height=1.8cm]{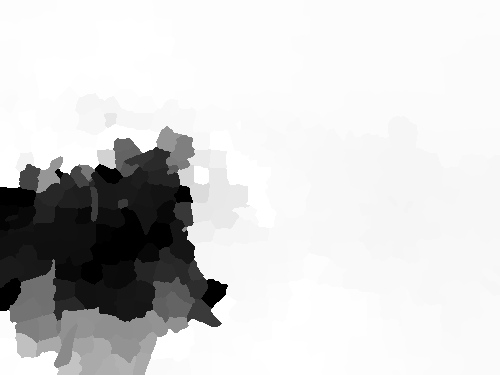}&
  \includegraphics[height=1.8cm]{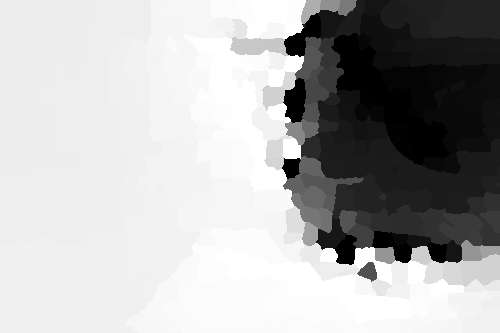}&
  \includegraphics[height=1.8cm]{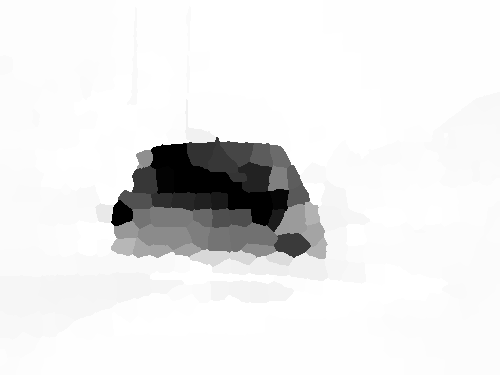}\\
  \begin{sideways}\parbox{15mm}{\centering\footnotesize Our method}\end{sideways} &
  \includegraphics[height=1.8cm]{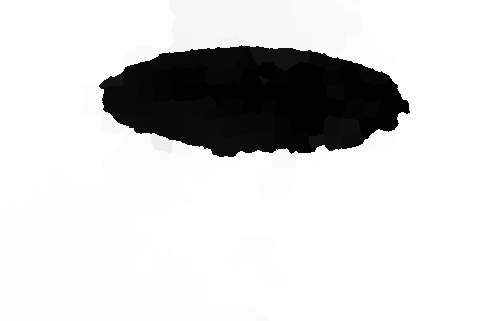}&
  \includegraphics[height=1.8cm]{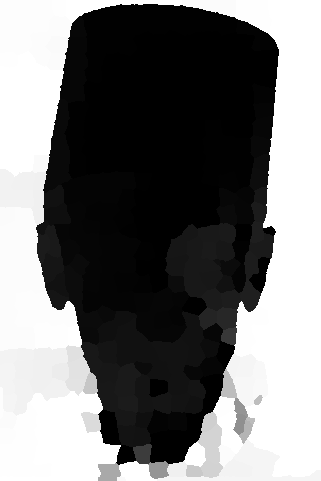}&
  \includegraphics[height=1.8cm]{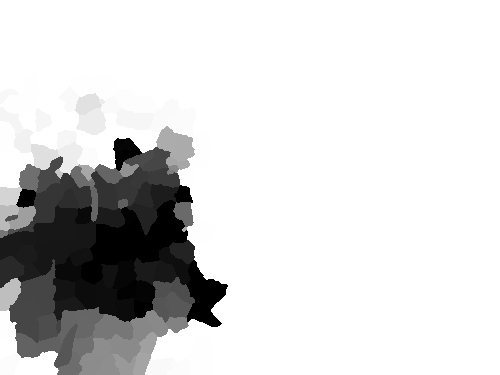}&
  \includegraphics[height=1.8cm]{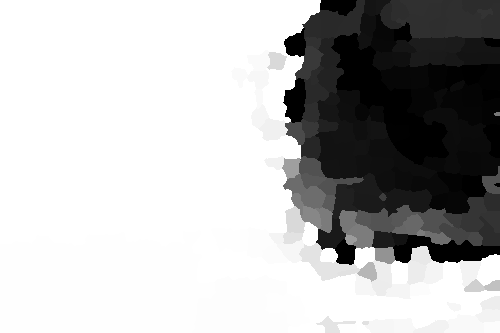}&
  \includegraphics[height=1.8cm]{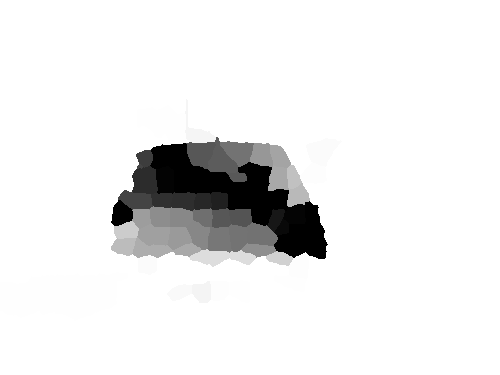}\\
    \end{tabular}
  \end{center} \caption{Continuous labels before thresholding from LP, QP and our method on example inputs in Fig.~\ref{FIG:OxfordSeg1}}\label{FIG:OxfordSeg_labels1}
\end{figure}

\begin{figure}
    \begin{center}
\begin{tabular}
{
@{\hspace{0mm}}c@{\hspace{0.5mm}}c@{\hspace{0.5mm}}c@{\hspace{0.5mm}}c@{\hspace{0.5mm}}c@{\hspace{0.5mm}}c @{\hspace{0.5mm}}c
@{\hspace{1mm}}c@{\hspace{1mm}}c@{\hspace{1mm}}c@{\hspace{1mm}}c@{\hspace{1mm}}c @{\hspace{1mm}}c
}
  \begin{sideways}\parbox{15mm}{\centering\footnotesize LP~\cite{LiHongDong10LPSeg}}\end{sideways} &
  \includegraphics[height=2cm]{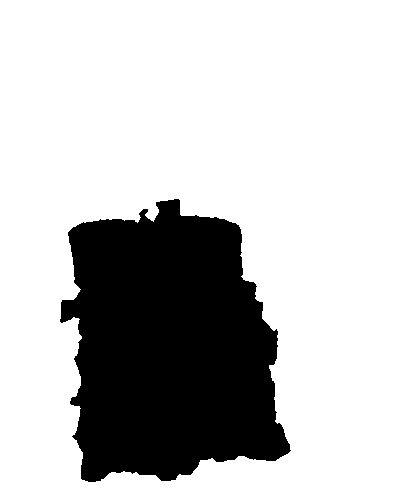}&
  \includegraphics[height=2cm]{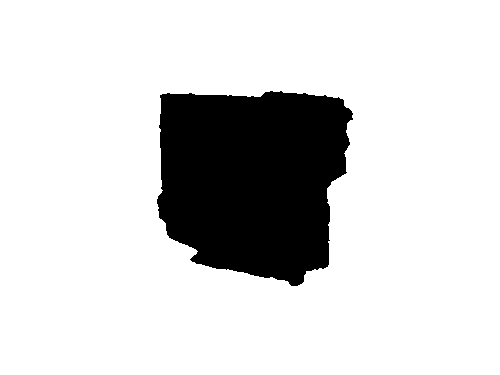}&
  \includegraphics[height=2cm]{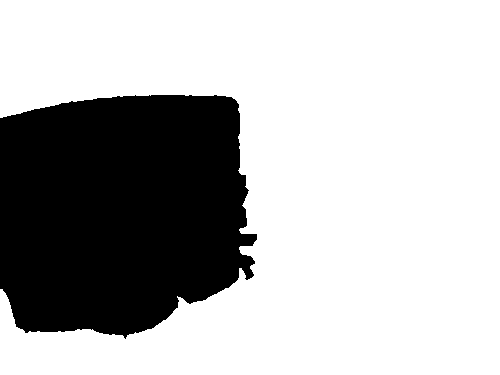}&
  \includegraphics[height=2cm]{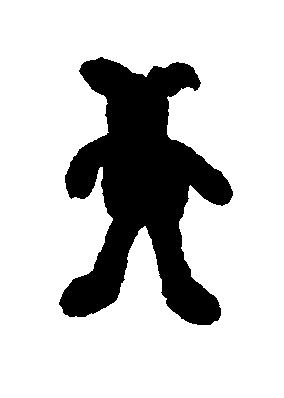}&
  \includegraphics[height=2cm]{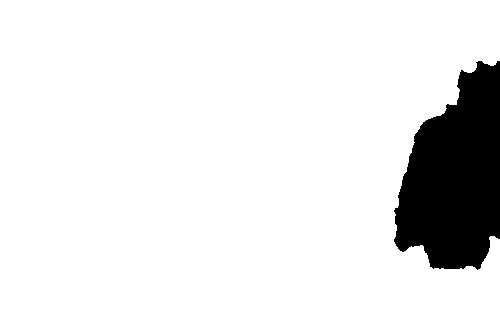}\\
  \begin{sideways}\parbox{15mm}{\centering\footnotesize QP~\cite{grady2006randomwalk,sinop2007seeded}}\end{sideways} &
  \includegraphics[height=2cm]{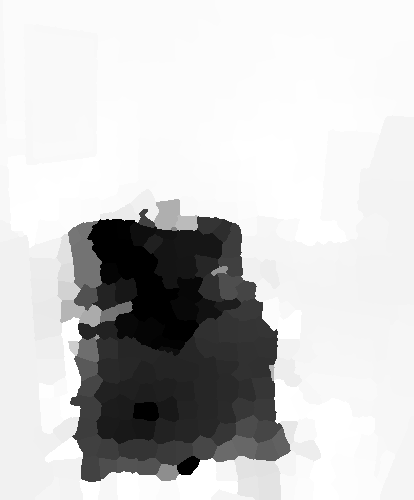}&
  \includegraphics[height=2cm]{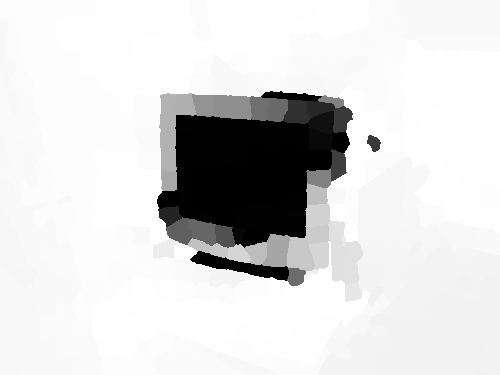}&
  \includegraphics[height=2cm]{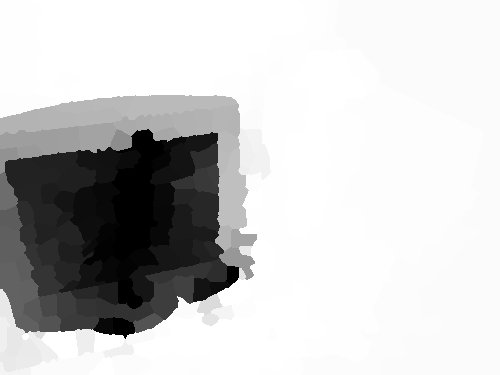}&
  \includegraphics[height=2cm]{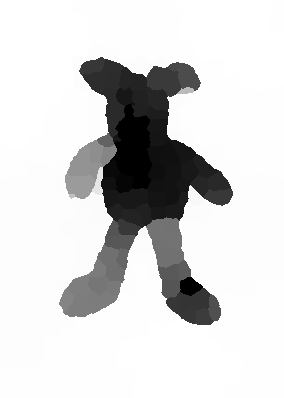}&
  \includegraphics[height=2cm]{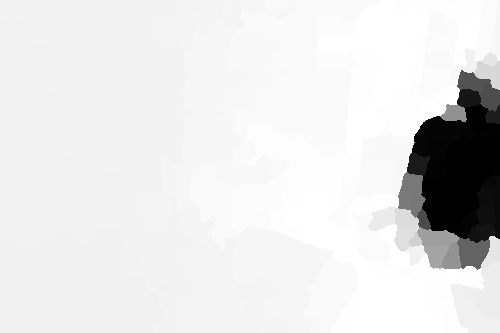}\\
  \begin{sideways}\parbox{15mm}{\centering\footnotesize Our method}\end{sideways} &
  \includegraphics[height=2cm]{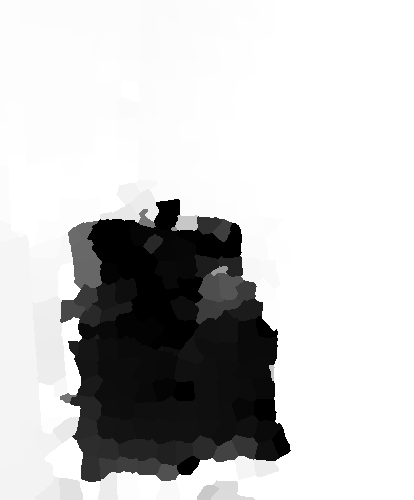}&
  \includegraphics[height=2cm]{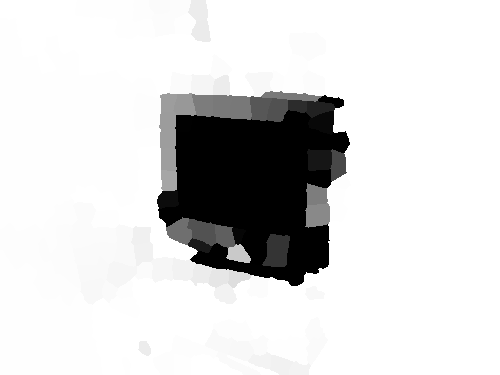}&
  \includegraphics[height=2cm]{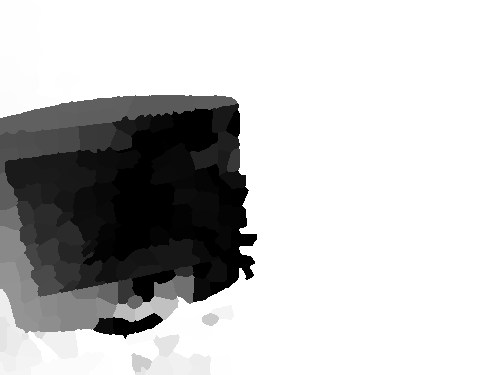}&
  \includegraphics[height=2cm]{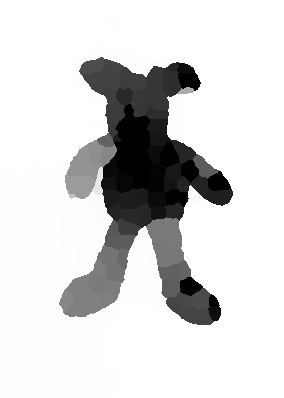}&
  \includegraphics[height=2cm]{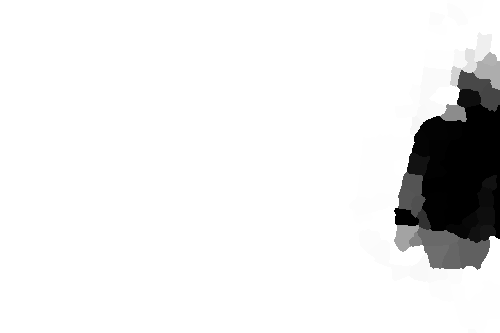}\\
    \end{tabular}
  \end{center} \caption{Continuous labels before thresholding from LP, QP and our method on example inputs in Fig.~\ref{FIG:OxfordSeg2}}\label{FIG:OxfordSeg_labels2}
\end{figure}

\begin{figure}
    \begin{center}
\begin{tabular}
{
@{\hspace{0mm}}c@{\hspace{0.5mm}}c@{\hspace{0.5mm}}c@{\hspace{0.5mm}}c@{\hspace{0.5mm}}c@{\hspace{0.5mm}}c @{\hspace{0.5mm}}c
@{\hspace{1mm}}c@{\hspace{1mm}}c@{\hspace{1mm}}c@{\hspace{1mm}}c@{\hspace{1mm}}c @{\hspace{1mm}}c
}
  \begin{sideways}\parbox{15mm}{\centering\footnotesize LP~\cite{LiHongDong10LPSeg}}\end{sideways} &
  \includegraphics[height=1.8cm]{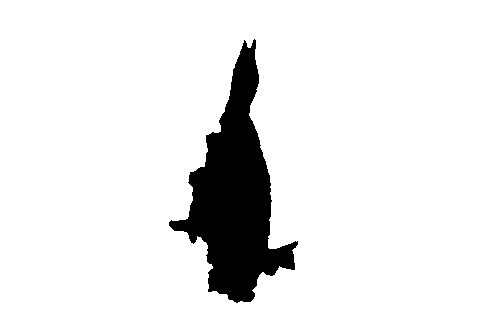}&
  \includegraphics[height=1.8cm]{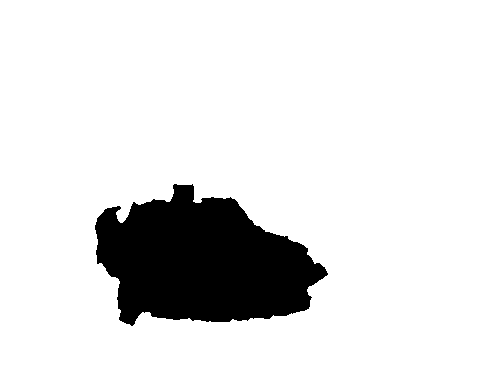}&
  \includegraphics[height=1.8cm]{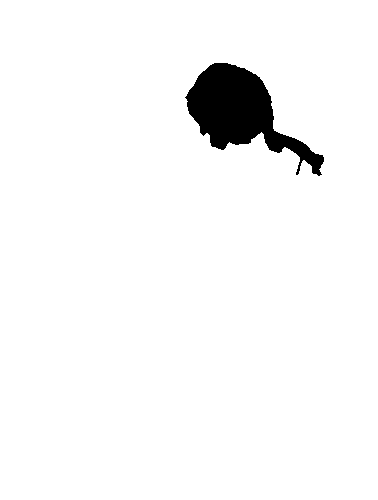}&
  \includegraphics[height=1.8cm]{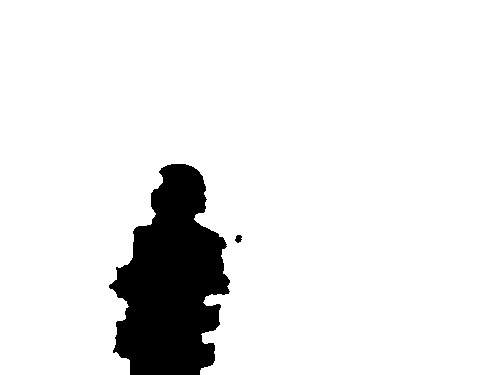}&
  \includegraphics[height=1.8cm]{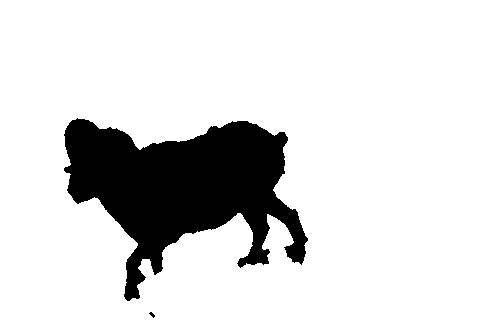}\\
  \begin{sideways}\parbox{15mm}{\centering\footnotesize QP~\cite{grady2006randomwalk,sinop2007seeded}}\end{sideways} &
  \includegraphics[height=1.8cm]{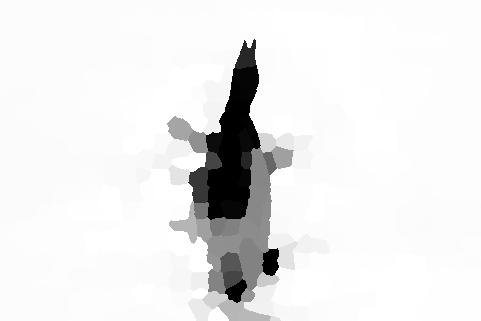}&
  \includegraphics[height=1.8cm]{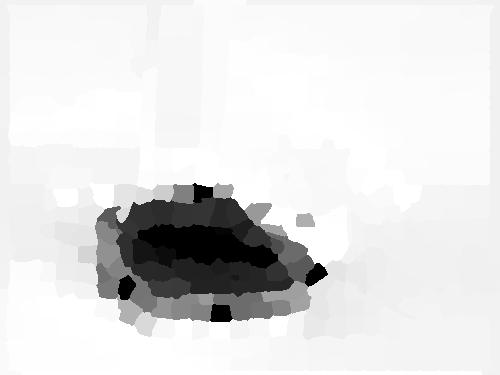}&
  \includegraphics[height=1.8cm]{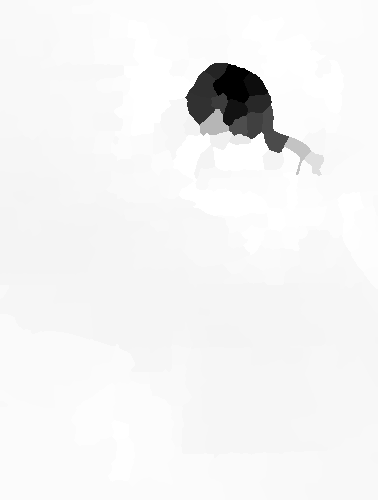}&
  \includegraphics[height=1.8cm]{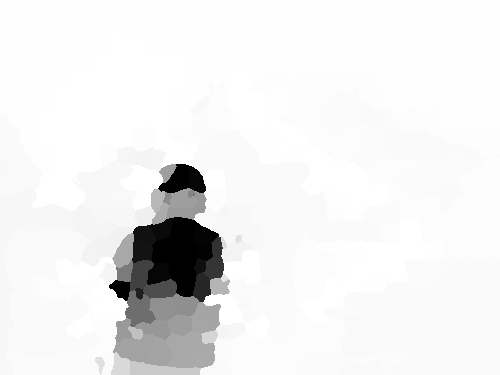}&
  \includegraphics[height=1.8cm]{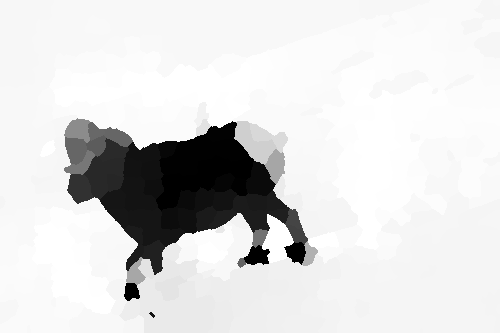}\\
  \begin{sideways}\parbox{15mm}{\centering\footnotesize Our method}\end{sideways} &
  \includegraphics[height=1.8cm]{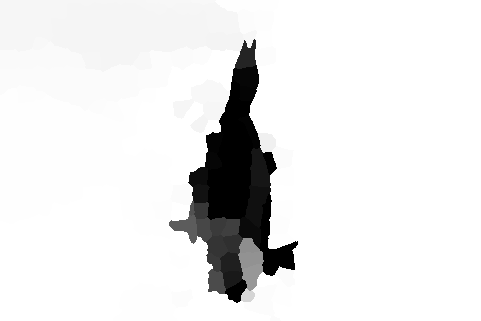}&
  \includegraphics[height=1.8cm]{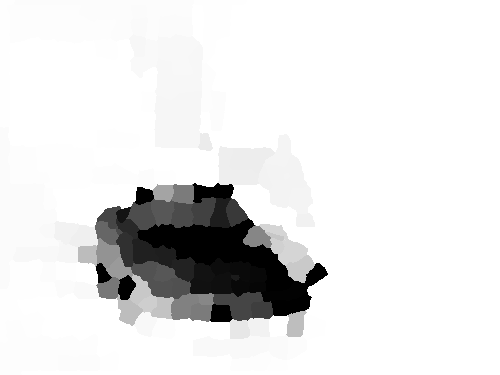}&
  \includegraphics[height=1.8cm]{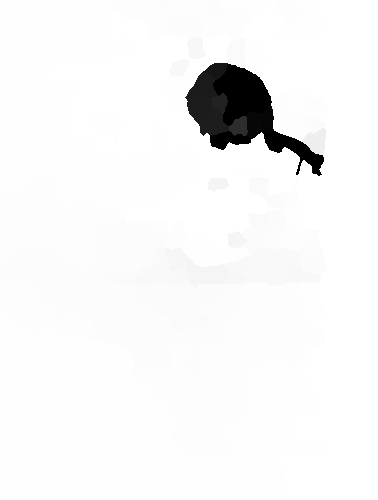}&
  \includegraphics[height=1.8cm]{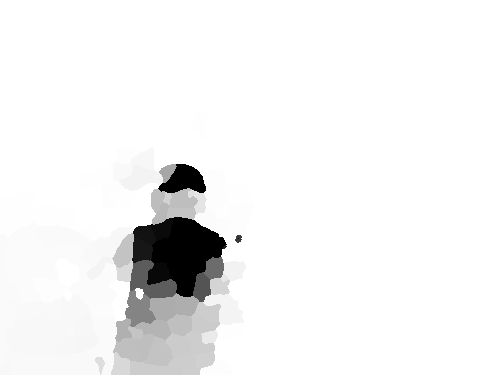}&
  \includegraphics[height=1.8cm]{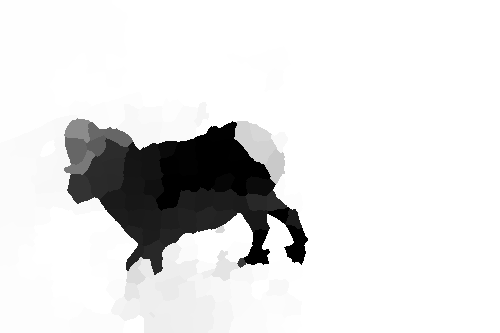}\\
    \end{tabular}
  \end{center} \caption{Continuous labels before thresholding from LP, QP and our method on example inputs in Fig.~\ref{FIG:OxfordSeg3}}\label{FIG:OxfordSeg_labels3}
\end{figure}

We present some additional experimental results in Figs.~\ref{FIG:OxfordSeg1}-\ref{FIG:OxfordSeg3}. The corresponding continuous solutions before thresholding are presented in Figs.~ \ref{FIG:OxfordSeg_labels1}-\ref{FIG:OxfordSeg_labels3}.

\end{document}